%% file: main_arxiv.tex
\documentclass[letterpaper, 10 pt, conference]{ieeeconf}  

\IEEEoverridecommandlockouts                              

\usepackage{adjustbox}
\usepackage[font=small,belowskip=-1.5pt]{caption} 
\usepackage{tikz}
\usepackage[colorlinks=true]{hyperref}

\usepackage{amssymb}
\usepackage{pifont}
\input{vcl-shortcuts.tex}
\usepackage{alphalph}
\usepackage{bbm}

\newcommand{\rt}[1]{\textcolor{black}{#1}}
\newcommand{\ours}{\rt{GA-Nav}}

\title{\LARGE \bf
GA-Nav: Efficient Terrain Segmentation for Robot Navigation in Unstructured Outdoor Environments

}

\author{Tianrui Guan, Divya Kothandaraman, Rohan Chandra,
Adarsh Jagan Sathyamoorthy, Kasun Weerakoon, \\
and Dinesh Manocha \\
\small{(Code and Videos at \url{https://gamma.umd.edu/offroad})}
}

\begin{document}

\maketitle
\thispagestyle{empty}
\pagestyle{empty}

\begin{abstract}


We present an efficient learning-based method for identifying safe and navigable regions in off-road terrains and unstructured environments from RGB images. Our approach classifies terrains based on their navigability levels using coarse-grained semantic segmentation. We propose \ours{}, a novel group-wise attention mechanism to distinguish between navigability levels of different terrains, which can improve different backbone designs.
Our group-wise attention loss enables the network to explicitly focus on the different groups' features with low spatial resolution for efficient inference while maintaining a high level of accuracy compared to other SOTA methods. 
We show through extensive evaluations on the RUGD and RELLIS-3D datasets that our learning algorithm improves visual perception accuracy in off-road terrains for navigation. We compare our approach with prior work on these datasets and achieve an improvement over the state-of-the-art mIoU by 
2.25-39.05\% on RUGD and 5.17-19.06\% on RELLIS-3D. 
In addition, we deploy \ours{} on a Clearpath Jackal and a Husky robot for real-world navigation demonstrations. 
Our approach improves the performance of the navigation algorithm in terms of success  rate by 10\%  and  results in smoother trajectories while maintaining the best surface selection capabilities above 80\%. Further, our method reduces the false positive rate of forbidden regions by 37.79\%.
Code, videos, and a full technical report are available at \href{https://gamma.umd.edu/offroad/}{gamma.umd.edu/offroad}.

\end{abstract}

\input{sections/introduction}

\input{sections/related_work_full}
\input{sections/overview_full}
\input{sections/experiments_full}
\input{sections/conclusions}

\section*{APPENDIX}

\input{sections/dataset}

\input{QualitativeResults/qualitative2}

\input{QualitativeResults/qualitative3}





{\small
\bibliographystyle{IEEEtran}
\bibliography{citation}
}

\end{document}

%% file: vcl-shortcuts.tex
\usepackage{times}
\usepackage{epsfig}
\usepackage{graphicx}
\usepackage{float}
\usepackage{wrapfig}
\usepackage{amsmath,amssymb}
\usepackage{bm,xspace}
\usepackage{comment}
\usepackage{verbatim}
\usepackage{multirow}
\usepackage{balance}
\usepackage{url}
\usepackage{booktabs}
\usepackage{etoolbox,siunitx}
\usepackage{calc}
\usepackage{pifont,hologo}
\usepackage{graphicx}
\usepackage{xcolor}

\usepackage{multicol}
\usepackage{algorithm2e}
\usepackage{makecell}
\usepackage{arydshln}
\usepackage{multirow}
\usepackage{subcaption}

\usepackage{thmtools}
\usepackage{thm-restate}

\newcommand{\ra}[1]{\renewcommand{\arraystretch}{#1}}

\usepackage[super]{nth}
\usepackage{nicefrac}
\robustify\bfseries

\def\tt{\mathbf{t}}

\DeclareMathSymbol{@}{\mathord}{letters}{"3B}

\def\latex/{\LaTeX}
\def\bibtex/{\hologo{BibTeX}}


%% file: sections/introduction.tex
\section{Introduction}
Recent developments in autonomous driving and mobile robots have significantly increased the interest in outdoor navigation~\cite{cityscape,kittiseg}. In many applications such as mining, disaster relief \cite{murphy2014disaster}, agricultural robotics \cite{r2018research}, or environmental surveying, the robot must navigate in uneven terrains or scenarios that lack a clear structure or well-identified navigation features.

A key issue in developing autonomous navigation capabilities in off-road environments is finding safe and navigable regions that can be used by a mobile robot. For instance, terrains like concrete or asphalt are smooth and highly navigable, while rocks or gravel are typically bumpy and may not be navigable. Our goal in this work is to learn to visually differentiate various terrains in unstructured outdoor environments and classify them based on their navigability.

An important aspect of classifying navigable regions in unstructured environments is using visual perception capabilities such as semantic segmentation~\cite{zhao2017pyramid, yuan2020objectcontextual, segformer, segmenter}. 
Semantic segmentation is a pixel-level task that assigns a label for every pixel in the image. Prior works in segmentation~\cite{zhao2017pyramid, yuan2020objectcontextual} have been limited to structured environments~\cite{cityscape,kittiseg} and may not be well-suited, in general, for aiding robot navigation in off-road terrains for several reasons.

Firstly, many off-road scenes require differentiating terrain classes that have highly similar appearances (\textit{e.g.}, water and a puddle) with overlapping boundaries. In some cases, certain terrain classes could be occupying a small portion of the image, which require highly accurate segmentation. In terms of existing methods~\cite{zhao2017pyramid, yuan2020objectcontextual, segformer, segmenter}, such scenarios lead to similar feature embeddings for multiple classes that can ultimately result in wrong classifications. Furthermore, for navigation, misclassifications of dangerous or forbidden regions for the robot could lead to disastrous consequences.

\begin{figure}[t]
    \centering
    \includegraphics[width = \columnwidth]{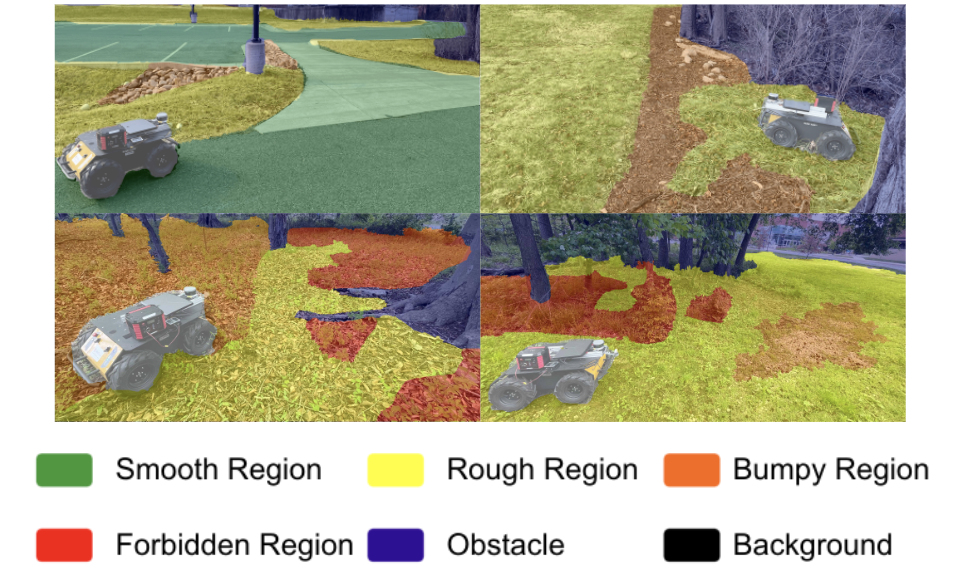}
    \vspace{-10pt}
    \caption{We highlight the performance of our algorithm on unstructured outdoor terrains. 
    We classify the navigable regions (shown in different colors) using our segmentation algorithm that uses RGB images as the input in real time.
    The top images correspond to different outdoor terrains where the robots are navigating in an autonomous manner. The bottom images show the terrain segmentation results.
    Our model achieves an improvement of $2.25-39.05\%$ in terms of mIoU over prior segmentation algorithms on complex outdoor terrains and improves navigation in the real world by $10\%$ in terms of success rate.}
    \vspace{-7mm}
    \label{fig:cover}
\end{figure}

Secondly, perception methods that aid off-road robot navigation must manage their computational efficiency without compromising on their segmentation accuracy. Thirdly, it has been shown that in unstructured environments, navigation systems may not require fine-grained semantic segmentation \cite{outdoor,large_nav1}. For example, it is sufficient to recognize trees and poles as obstacles for collision avoidance rather than segmenting them individually as different types of objects. A coarser approach that classifies various types of objects based on their navigation characteristics is more appropriate for aiding navigation, which is the focus of our method.

\noindent{\bf Main Results:}
We present \ours{}, a novel and efficient learning-based approach for identifying the navigability of different terrains from RGB images or videos. Our approach is designed to identify different \textit{terrain groups} in off-road and unstructured outdoor terrains. We present a novel architecture that finds a good balance between segmentation accuracy and computational overhead and outperforms existing transformer-based segmentation methods. Unlike previous methods~\cite{dpt, maskformer, segformer,segmenter}, which only focus on fine-grained classification, our method implicitly learns how to classify and cluster different classes simultaneously according to annotations, which leads to improved accuracy among different terrains. We show that our method has an advantage over existing methods on coarse-grained segmentation task used for navigation. 
To demonstrate \ours{}'s benefits for real-time robot navigation, we integrate it with a learning-based navigation approach and evaluate the performance on Clearpath Jackal and Husky robot in unstructured outdoor environments. 

The key contributions of our work include:

\begin{enumerate}
    \item We propose a novel architecture which fuses a multi-scale feature extractor with a transformer architecture. Our group-wise segmentation head can fuse visual features from different scales and explicitly focus on different terrain types, which leads to better accuracy on different surfaces of varying areas.
    Our approach is compatible with different feature extractors and can improve performance on various SOTA multi-scale backbone designs. We outperform prior methods for semantic segmentation and navigable region classifications in terms of mIoU by $2.25-39.05\%$ on RUGD and $5.17-19.06\%$ on RELLIS-3D.
    
    \item We introduce a group-wise attention loss for fast and accurate inference. We show that with our group-wise attention (GA) loss, we can improve the inference complexity without much performance degradation. Our proposed method results in better performance than existing transformer-based and other segmentation methods with one of the lowest run-times and complexity requirements.
    
    \item We integrate \ours{} with TERP\cite{terp}, an outdoor navigation planner to highlight the benefits of our segmentation approach for robot navigation in complex outdoor environments. We show that our modified navigation method outperforms other navigation methods by 10\% in terms of success rate and, 2-47\% in terms of selecting the surface with the best navigability and a decrease of 4.6-13.9\% in trajectory roughness. Further, \ours{} reduces the false positive rate of forbidden regions by 37.79\%.

\end{enumerate}

%% file: sections/related_work_full.tex
\section{Related work}

\subsection{Semantic Segmentation}
Semantic segmentation is an important task in computer vision that involves assigning a label to each pixel in an image. The problem has been widely studied in the literature~\cite{hao2020brief}.
Many deep learning architectures have been proposed, including OCRNet~\cite{yuan2020objectcontextual}, PSPNet \cite{zhao2017pyramid}, etc. 
Most recently, there have been many transformer-based methods~\cite{setr, dpt} that demonstrate better performance than other CNN-based methods but are more computationally expensive. Therefore, there has been increased discussion of more efficient design based on transformers for fast inference time and lower computational cost, including \cite{segformer,segmenter}. 
While most of these architectures work well on structured datasets like CityScapes, they do not work well for off-road datasets due to ill-defined boundaries and confusing class features. Segmentation methods like Global Convolution Networks \cite{peng2017large} are designed to deal with complex boundaries and do not scale well in terms of performance due to convoluted and overlapping boundaries that are generally not found in structured datasets.

\subsection{Navigation in Outdoor Scenes}
Prior work in uneven terrain navigation includes techniques for everything from mobile robots~\cite{outdoor, small_nav1, kahn2020badgr, robot_nav}
to large vehicles~\cite{large_nav1, off_road_trav1, pointcloud_hazard}. In these algorithms, learning the navigation characteristics of the terrain and its traversability is a crucial step. At a broad level, there are three types~\cite{dissertation1} of approaches that are used to determine the terrain features: 1) proprioceptive-based methods, 2) geometric methods, and 3) appearance-based methods. The proprioceptive-based methods~\cite{vibrate1,vibrate2} 
use frequency domain vibration information gathered by the robot sensors to classify the terrains using machine learning techniques. However, these methods require the robot to navigate through the region to collect data and assume that a mobile robot can navigate over the entire terrain. The geometric-based methods~\cite{pointcloud_hazard, off_road_trav1, roughness2}
generally use Lidar and stereo cameras to gather 3D point cloud and depth information of the environment. This information can be used to detect the elevation, slope, and roughness of the environment as well as obstacles in the surrounding area. Nevertheless, the accuracy of the detection outcome is usually governed by the range of the sensors. Appearance-based methods~\cite{sift_based} usually extract road features with SIFT or SURF features and use MLP to classify a set of terrain classes. 
Our goal is to focus on the visual perceptive, leveraging the performance and efficiency for robot navigation. Our method computes the navigable regions from RGB images in unstructured environments and outputs a safe trajectory in this condition.

\begin{figure*}[t]
    \centering
    \includegraphics[width=\textwidth]{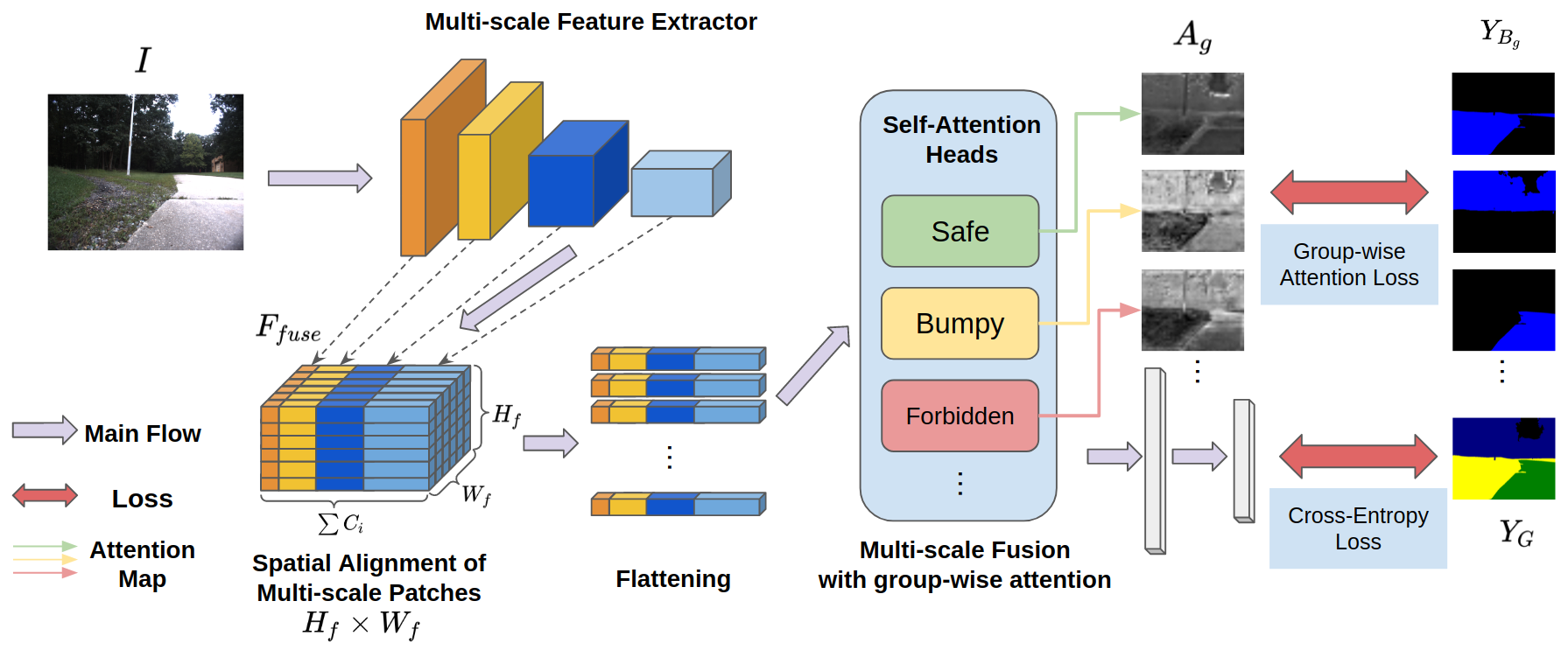}
    \caption{\textbf{Architecture of our proposed network \ours:} 
    We introduce a novel group-wise attention head for multi-scale fusion. After we convert multi-scale features into several groups of features, the group-wise attention loss acts between the multi-head attention outputs $A_g$ corresponding to group $g$ and the corresponding group-wise binary ground truth $Y_{B_g}$ to explicitly guide the network towards accurate predictions of different groups. 
    }
    \label{fig:overview}
    \vspace{-15pt}
\end{figure*}


\subsection{Off-road Datasets}
Recent developments in semantic segmentation have achieved high accuracy on object datasets like PASCAL VOC~\cite{pascal}, COCO~\cite{coco}, and ADE20K~\cite{ade20k2}, as well as driving datasets like Cityscape~\cite{cityscape}
and KITTI~\cite{kittiseg}. However, there has not been much work on recognition or segmentation in unstructured off-road scenes, which is important for navigation. Perception in an unstructured environment is more challenging since many object classes (e.g., puddles or asphalt) lack clear boundaries. RUGD~\cite{RUGD2019IROS} and RELLIS-3D~\cite{rellis} are two recent datasets available for off-road semantic segmentation. The RUGD dataset consists of various scenes like trails, creeks, parks, and villages with fine-grained semantic segmentation annotations. The RELLIS-3D dataset is derived from RUGD and includes unique terrains like puddles. In addition, RELLIS-3D includes Lidar data and 3D Lidar annotations. We use these datasets to evaluate the performance of our algorithm.



%% file: sections/overview_full.tex
\section{\ours{} For Terrain Classification}

We present \ours, an approach for classifying the navigability of different terrains in off-road environments via coarse-grained segmentation.
Most of the existing segmentation methods fail in unstructured environments due to highly similar visual features and unclear boundary of the objects.
Use of multi-head self-attention is popular in neural network design due to its superior performance over prior state-of-the-art methods in segmentation~\cite{setr, dpt}, image classification~\cite{pham2021meta}, and object detection~\cite{coco, srinivas2021bottleneck, m3detr}, but the quadratic nature of its computational complexity has been an issue for real-time use like navigation.

To solve performance degradation in unstructured environment and computation limitation, we design a simple and lightweight architecture, which uses self-attention at the multi-scale fusion step and generated attention maps to calculate loss for better performance and low level of computational overhead.



The rest of this section is structured as follows. We begin by stating our problem in Section~\ref{def}. We then discuss the various components of our approach, namely the backbone architecture (Section~\ref{backbone}), \ours{} segmentation head (Section~\ref{group_wise}), and an attention-based loss function (Section~\ref{loss}). Finally, we clarify the difference between \ours{} and other attention-based methods and briefly explain its integration with planning and navigation modules for navigation demonstration.


\subsection{Problem Definition}
\label{def}
The input consists of an RGB image $I \in \mathbb{R}^{3\times H \times W}$ and the corresponding ground-truth semantic segmentation labels $Y\in \mathbb{Z}^{H\times W}$ denoting the category to which each pixel belongs among $G$ different groups. We use new coarse-grain labels $Y_G \in \mathbb{Z}^{H\times W}$ based on sematic labels provided by the dataset. For each group, we also compute the binary mask $Y_{B_g} \in \{0, 1\}^{H\times W}$ for $g = 1, ..., G$. Our goal is to perform coarse-grain semantic segmentation on terrain images $I$ and compute probability maps $P\in\mathbb{R}^{G\times H \times W}$. Each entry in $P$ is a probability distribution of a pixel in that entry location belonging to one of the $G$ groups.  



\subsection{Backbone Design}
\label{backbone}

Vision Transformers (ViTs)~\cite{srinivas2021bottleneck, 16x16} are a type of deep neural network (DNN) for feature extraction and are designed as an alternative to the widely used ResNet~\cite{resnet} architecture via Multi-Head Self-Attention (MHSA). MHSA, or self-attention applied on image patches, is the main component of ViTs, contributing to much of their success over ResNets in computer vision-based tasks. Our approach also uses a transformer-based backbone architecture and leverages the MHSA module to extract and fuse multi-scale features. Our architecture can adapt to any backbone design with multi-scale features; in particular, we use Mixed Transformer (MiT)~\cite{segformer} with some modifications as our backbone. In addition, to show the advantages of our design, we also show two alternative backbones with our design, ResNet50~\cite{resnet} and Bottleneck Transformer (BoT)~\cite{srinivas2021bottleneck}.

Given an RGB image $I \in \mathbb{R}^{3\times H \times W}$, we first divide it into several patches as input to the transformer encoders. Unlike ViT and MiT, which take non-overlapping patches of shape 16 by 16 and 4 by 4, respectively, we choose patches of size 7 by 7 with a stride of 4 in our design. After passing the patches to the first transformer encoder blocks, we continue to take 2 by 2 patches on the output features. Therefore, after the initial encoders, each following block of transformer encoders reduces the spatial resolution further by a factor of 2 on both height and width. After each transformer encoder block, we obtain multi-scale features $F_i$ with spatial resolution $H_i \times W_i = \frac{H}{2^{i+1}} \times \frac{W}{2^{i+1}}$, where the feature channel $C_i$ = \{32, 64, 160, 256\} and i = \{1, 2, 3, 4\}. We utilize all those multi-scale features in our segmentation head.

\subsection{GANav Segmentation Head}
\label{group_wise}

The RGB image $I$ is passed through the feature extractor to obtain multi-scale feature maps $F_i \in \mathbb{R}^{C_i\times H_i \times W_i}$ after $i$-th encoder block. 

\noindent\textbf{Spatial Alignment and Resolution Reduction:}
To prepare for our fusion, we need to combine features at the same spatial locality. We use bi-linear interpolation to resize $F_i$ into shape ($C_i\times H_f \times W_f$), where $H_f$ and $W_f$ are equal to $H_i$ and $W_i$ with any chosen $i$ for all scales. The final feature before fusion, $F_{fuse}$, is concatenated over channel dimension with shape ($\sum{C_i}\times H_f \times W_f$)). 

To reduce the computational complexity, our efficient design prefers to align patch features from different scales to one scale with a smaller spatial resolution as the network bottleneck (i.e., choose a smaller $H_f$ and $W_f$). As we expect, as we reduce $H_f$ and $W_f$, the spatial resolution of $F_{fuse}$ reduces so the complexity goes down with some performance degradation. In Section~\ref{loss}, we also introduce the group-wise attention loss to not only improve the overall performance but also mitigate the performance drop from choosing a smaller spatial resolution. 

\noindent\textbf{Multi-scale Fusion with Group-wise Attention:}
Given an input feature vector $A_{in} \in \mathbb{R}^{N\times C_{in}}$, the output self-attention map $A_{out}\in \mathbb{R}^{N\times C_{out}}$ is computed as follows:
\begin{equation}
     A_{out} = softmax(k(A_{in})^T \cdot q(A_{in}))^T \cdot v(A_{in}),
     \label{eq: attention}
\end{equation}
where $k(A_{in}),\ q(A_{in}),\ v(A_{in})$ represent the key, query, and value feature maps, respectively, and $k$, $q$, $v$ are linear projections, as in the self-attention \cite{fu2019dual} literature. 

This feature map $F_{fuse}$ is reshaped (and transposed) into $F_{flat} \in \mathbb{R}^{ H_fW_f\times \sum C_{i}}$, which is passed through the MHSA block with $G$ attention heads. The MHSA component fuses flatten multi-scale feature $F_{flat}$ to produce the new feature maps $F_{out} \in \mathbb{R}^{C_{out}\times H_f \times W_f}$ and generates $G$ attention maps (one for each group), $A_1, A_2, \ldots, A_G \in[0, 1]^{H_f W_f\times H_f W_f}$. The final output $P\in\mathbb{R}^{G\times H \times W}$ is obtained through a standard procedure of a segmentation network by a series of $1\times1$ convolutions and up-sampling from $F_{out}$.

As demonstrated in Fig.~\ref{fig:overview}, in our detection head, we use MHSA to fuse multi-scale features and generate different attention maps for each group. We use those attention maps as an additional branch in the detection head and train the detection head to resemble the group distribution by explicitly guiding each attention map towards a corresponding category using a binary cross-entropy loss. Our network skips the attention map processing branch during inference time for efficiency.

Attention maps can also capture the relevancy between two pixel locations. Intuitively, $A_{g, [x, y]}$ represents the amount of attention that the pixel at the $x^\textrm{th}$ position must pay to the pixel at the $y^\textrm{th}$ position. Given an attention map $A_g$, its main diagonal represents the relevance of each location with respect to the attention head $h_g$. We train the attention network to learn feature maps that are most relevant to its corresponding group through group-wise attention loss.

We use group-wise attention as a fusion method for multi-scale features, which has been shown to be quite powerful in combining different modalities~\cite{model_fusion}, different scales and representations~\cite{m3detr}, and temporal information~\cite{tmp_fusion} for various tasks. We will show that our novel design of group-wise attention for multi-scale fusion can outperform most existing transformer-based methods with outstanding efficiency on coarse-grain segmentation tasks in challenging off-road environments.

\subsection{Group-wise Attention Loss}
\label{loss}

Whenever we choose to reduce the resolution during the spatial alignment step, performance degradation is unavoidable. Based on the structure of our group-wise attention head, we propose group-wise attention loss to boost the performance and mitigate such effects. 

For each attention head $h_i$, we have its corresponding attention map $A_i \in [{0,1}]^{L\times L}$, where $L = H_f\times W_f$. We take the main diagonal of $A_i$ and reshape it to $B_i \in [{0,1}]^{H\times W}$ using bi-linear image resizing. Each pixel in $B_i$ represents the self-attention score with respect to $h_i$.
 
To guide each attention-map in the multi-head self-attention module, we apply a binary cross-entropy loss function: 
\begin{equation}
     \mathcal{L}_{\textrm{GA}}^{g} = - \sum_{h,w} y_{\textrm{G}} \log(B_g),
     \label{eq: loss_cA}
\end{equation}
where $y_G \in Y_G$. This equation calculates loss between the predictions of the self-attention output and the corresponding group's binary ground truth with respect to the group. 

The purpose of group-wise attention loss is not to use attention maps to accurately predict the group's distribution; instead, as an intermediate layer, it aims to provide guidance regarding which region an attention head should focus on for further classification.

As in traditional segmentation models, we optimize our network with a multi-class cross-entropy loss and an auxiliary loss:

\noindent\textbf{Cross-Entropy Loss:} This is the standard semantic segmentation cross-entropy loss, defined as follows:
\begin{equation}
     \mathcal{L}_\textrm{CE} = - \sum_{h,w} \sum_{g\in G
     } y_\textrm{GT}\log(P_g
     ),
     \label{eq: loss_ce}
\end{equation}
where $h,w$ represent the dimensions of the image, $G$ represents the set of groups, $P_g$ denotes the output probability map corresponding to group $g$, and $y_{GT}$ corresponds to the ground-truth annotations.












\noindent\textbf{Auxiliary Loss via Deep Supervision:} Deep supervision was first proposed in \cite{deepsuper} and has been widely used in the task of segmentation \cite{zhao2017pyramid, yuan2020objectcontextual, 2018psanet}. Adding an existing good-performing segmentation decoder head like FCN~\cite{long2015fully} in parallel to our GA head during training not only provides strong regularization but also reduces training time.

\subsection{ \ours{} v.s. Prior Methods}

In this section we highlight the difference between some transformer-based methods and the proposed method, \ours{}.

SETR~\cite{setr}, DPT~\cite{dpt}: The designs are significantly different, particularly regarding the computational overhead of those methods. In addition, their method uses positional embedding in the encoding stage while our method removes any positional embedding in the backbone.


Segformer~\cite{segformer}, Segmenter~\cite{segmenter}: Those methods, as well as \ours{}, also focus on efficient training and inference. Segformer does not have any parameters dedicated to different classes except for the final classification layer. In addition, the decoder consists of several MLP layers for simple and efficient design. Segmenter and \ours{} both have trainable parameters dedicated to specific classes/groups: Segmenter uses $G$ class embeddings that correspond to each semantic class and appends those tokens to the end of the entire image patch sequence for the decoder, while \ours{} splits feature channels into $G$ portions of equal size for different attention heads that belong to different groups. Finally, \ours{} uses a smaller spatial resolution in the decoder bottleneck than the other two methods to improve the run-time and compensated for the complexity brought by MHSA.

\subsection{\ours{}-based Segmentation for Robot Navigation}
We combine \ours{} with TERP~\cite{terp}, an outdoor navigation planner to highlight the benefits of our segmentation algorithm for robot navigation. TERP utilizes the robot's pose, goal information and a point cloud based elevation map to generate a cost map from an attention based perception module. Before interfacing with TERP, we convert \ours{}'s segmentation results into a segmentation cost map. This is performed based on the following steps: 1. Projecting \ours{}'s segmented output onto the ground plane in front of the robot using homography transformations, and 2. Assigning discrete cost values for the different groups. We assign zero cost for smooth regions and increase the non-zero costs for rough, bumpy, forbidden and obstacle regions. 

\begin{figure}[t]
    \centering
    \includegraphics[width = \columnwidth]{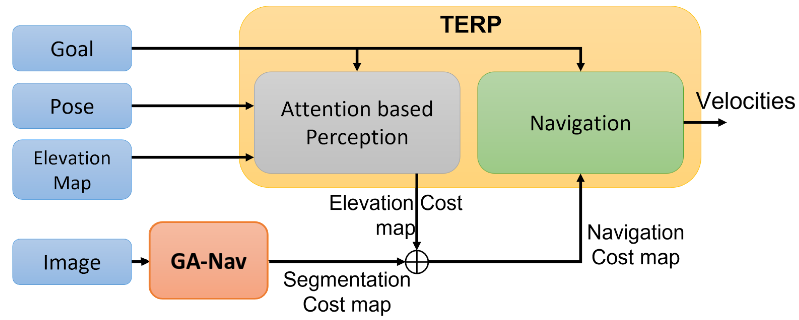}
    \caption{\textbf{\ours{}-Based Outdoor Navigation:} We integrate \ours{} with an elevation based planner TERP \cite{terp} to highlight benefits of our segmentation algorithm. Our architecture decouples semantic segmentation and elevation data for better perception and planning. The advantages of this formulation are highlighted in Section \ref{nav_comparisons}.}
    \label{fig:system-architecture}
\end{figure}

The computed segmentation cost map is element-wise added with the elevation cost map generated from TERP's perception module to obtain the final navigation cost map. This resulting cost map is fed into TERP's navigation module, which computes least-cost trajectories towards the robot's goal. The overall system architecture is highlighted in Fig. \ref{fig:system-architecture}.

%% file: sections/experiments_full.tex
\section{Results and Analysis}
\label{exp}

\subsection{Implementation Details}

We use two off-road terrain datasets, RUGD~\cite{RUGD2019IROS} and RELLIS-3D~\cite{rellis}, as benchmarks for all experiments and comparisons. Optimization is done using a stochastic gradient descent optimizer with a learning rate of 0.01 and a decay of 0.0005. We adopt the polynomial learning rate policy with a power of 0.9. We augment the data with horizontal random flip and random crop. In the ablations and comparisons, for fairness, we benchmark all our models at 240K iterations. For the RUGD dataset, we use a batch size of 8 and a crop size of $300\times375$ (the resolution of the original image is $688\times550$). On the other hand, for the RELLIS-3D dataset, due to the high resolution of the image, we train models with a batch size of 2 and a crop size $375\times600$ (the resolution of the original image is $1920\times1600$). Our implementation is based on MMSeg~\cite{mmseg2020} and will be released.

We benchmark our models on the standard segmentation metrics: Intersection over Union (IoU), mean IoU (mIoU), mean pixel accuracy (mAcc), and average pixel accuracy (aAcc). For an image $I$, let $P(x, y)$ be the predicted label at pixel location $(x, y)$, $G(x, y)$ be the ground truth label at $(x, y)$, $\mathbbm{1}(X)$ be the indicator function, and $B$ be a set of the class labels:

\noindent \textbf{Intersection over Union (IoU) for class i:}
$$IoU_{i} = \frac{\sum_{I}\sum_{x, y} \mathbbm{1}(P(x,y) = i \text{ and } G(x, y) = i) }{\sum_{I}\sum_{x, y} \mathbbm{1}(P(x,y) = i \text{ or } G(x, y) = i) }$$

\noindent \textbf{Mean IoU (mIoU):} 
$$mIoU = \frac{\sum_{i} mIoU_i}{\sum_{B} 1}$$

\noindent \textbf{Mean Pixel Accuracy (mAcc):} 
$$mAcc = \frac{\sum_{i\in B} (\sum_{I}\sum_{x, y, G(x, y) = i} \mathbbm{1}(P(x,y) = G(x, y)) )}{\sum_{B} 1}$$

\noindent \textbf{Averaged Pixel Accuracy (aAcc):} 
$$aAcc = \frac{\sum_{I}\sum_{x, y} \mathbbm{1}(P(x,y) = G(x, y)) }{\sum_{I}\sum_{x, y} 1 }$$

\begin{figure}[t]
    \centering
    \includegraphics[width = \columnwidth]{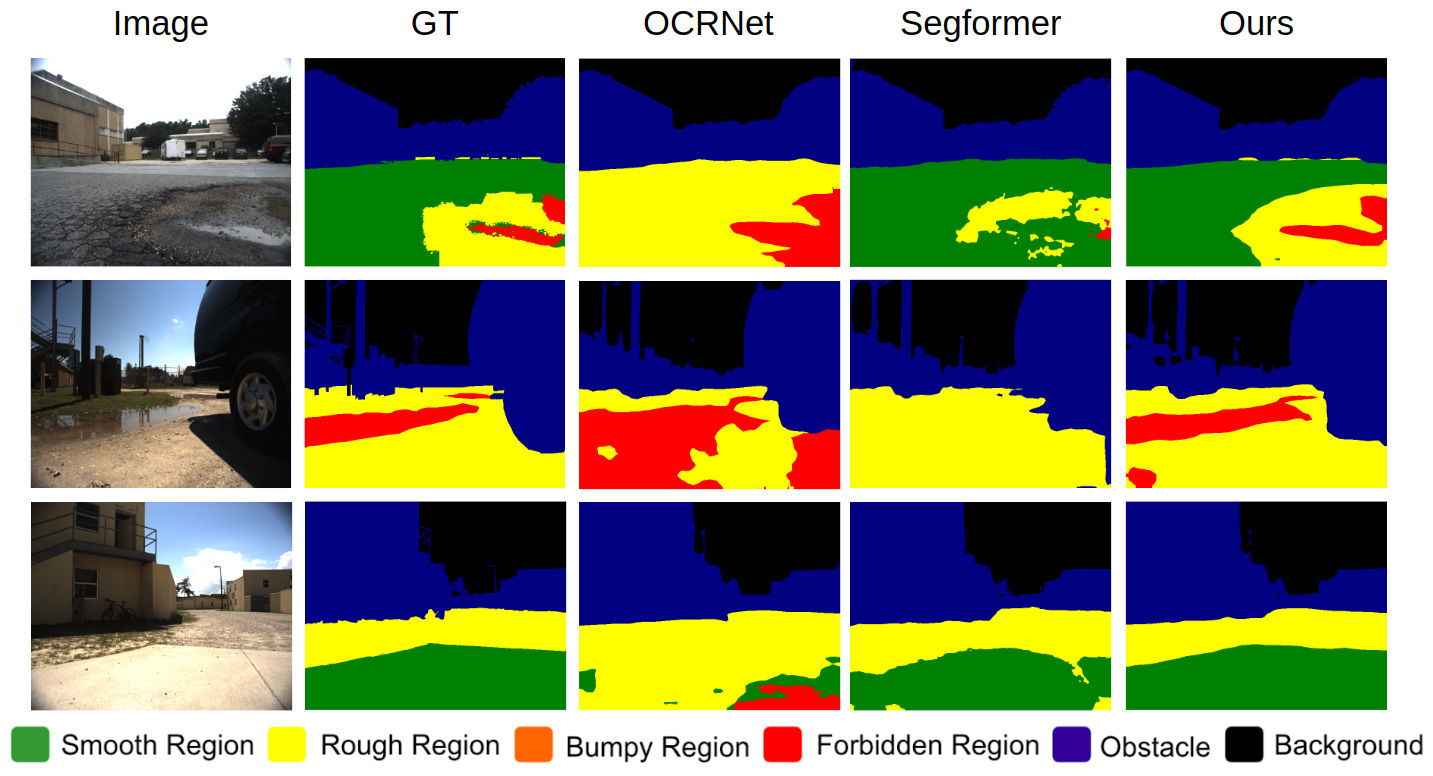}
    \caption{\textbf{Qualitative Results:} Each row highlights the original image, the ground truth (GT) label, the predictions of two SOTA methods, and the prediction of our method. We observe that our method can detect different navigable regions and is robust with respect to objects with similar colors.}
    \vspace{-3mm}
    \label{fig:visualisations}
\end{figure}

We evaluate our methods with 6 coarse-grain semantic classes: smooth (concrete, asphalt), rough (gravel, grass, dirt), bumpy (small rocks, rock-bed), forbidden (water, bushes), obstacle, and background. Some visualizations for method comparisons are provided in Fig.~\ref{fig:visualisations}.

\subsection{Ablation Studies}

\noindent\textbf{Effect of Multi-scale Fusion Design:} In Table~\ref{tab:ablation3}, we show the effect of our multi-scale fusion design with group-wise attention. We show an improvement of $14.76\%$ and $11.43\%$ in terms of mIoU on RUGD and RELLIS-3D, respectively.

\noindent\textbf{Group-wise Attention (GA) Design:} In Table~\ref{tab:ablation}, we show the effect of our segmentation head design with different backbones on RUGD and RELLIS-3D datasets. We observe that, regardless of the backbone choice, group-wise attention improves performance (in terms of mIoU) by $6.67-10.85\%$ and $4.18-10.31\%$ on RUGD and RELLIS-3D, respectively.

\input{Tables/ablation3}

\input{Tables/ablation_full}
\input{Tables/ablation2_full}

\input{Tables/runtime_full}

\noindent\textbf{Ablation on Spatial Reduction with GA Head:} In Table~\ref{tab:ablation2}, we show the relations between performance and resolution change during spatial alignment. We can see that without GA heads, as we change the resolution to 1/16 and 1/32, there are performance drops of $1.3\%$ and $4.0\%$ in terms of mIoU on RELLIS-3D. The performance drops do not appear for RUGD when we change the spatial reduction from 1/8 to 1/16; the main reason may be that the overall performance is low for RUGD before using GA heads, or the original resolution for RUGD is lower than RELLIS-3D, so the performance drop due to spatial information loss has not started to appear. We can still see the performance drop when reducing it to 1/32. After we add the GA component, the mIoU performance improves by $7.46-10.85\%$ and $2.85-5.85\%$ on RUGD and RELLIS-3D, respectively. In addition, we no longer see the performance drop on RELLIS-3D with the proposed GA heads. Use of GA heads allows us to reduce spatial resolution for efficient inference without compromising the final performance.




\subsection{Run-time Analysis and Computational Cost}

\begin{figure}[t]
    \vspace{-6mm}
    \centering
    \includegraphics[width = 1.05\columnwidth]{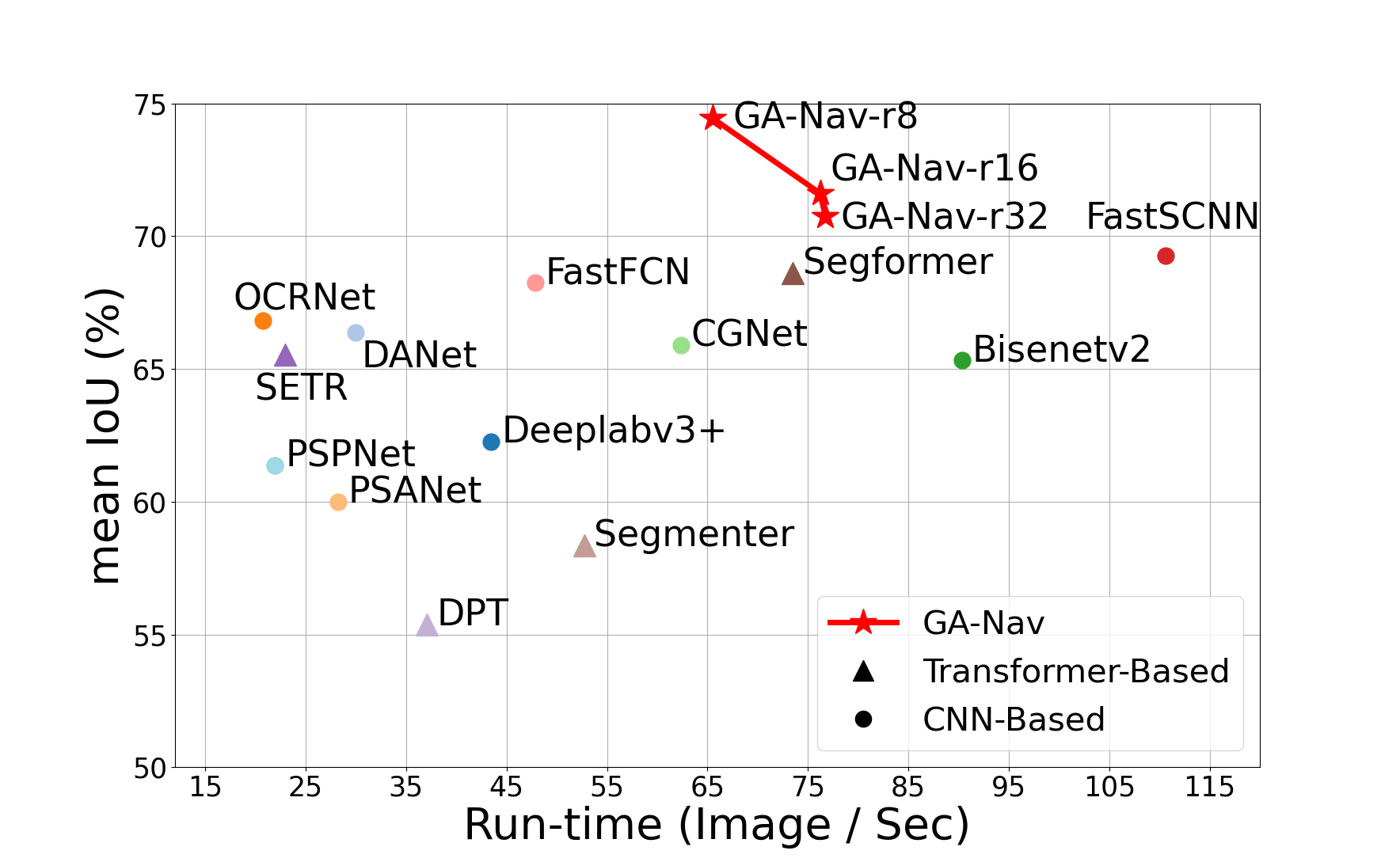}
    \caption{\textbf{Performance vs. Inference Time on RELLIS-3D:} The run-times are calculated based on the average time of 200 images.}
    \label{fig:runtime}
    \vspace{-3mm}

\end{figure}

In Fig.~\ref{fig:runtime}, we highlight the model efficiency of our method while maintaining SOTA performance. We can see that, even if we compromise the performance for inference rate in \ours{}-r16 and \ours{}-r32, our method still outperforms other methods in terms of mIoU.

In Table~\ref{tab:runtime}, we give more details on model parameter size, GFLOPS, cost during inference, and run-time. While our method doesn't always have the smallest model size and GLOPS, our method has a very low memory requirement. Our method also has the best inference time among listed transformer-based methods and most of the CNN-based methods, given that our method achieves state-of-the-art accuracy, as shown in Section~\ref{sota_comp}.

\subsection{Results and State-of-the-art Comparisons}
\label{sota_comp}
The results on different coarse-grain classes are presented in Table \ref{tab:comparison}. 
Our architecture improves the state-of-the-art mIoU by $1.66-44.35\%$ on RUGD. We can see that our method has the best overall mIoU scores. While \ours{} does not always give the best results for each separate class and the performance is sometimes close to the second best IoU score, our method has very stable performance for each class. \ours{} performs well on detecting different terrains, while other methods only have good performance on one of the terrain classes.
Similarly, on the RELLIS dataset, \ours\  demonstrates an improvement of $5.17-19.06\%$ in terms of mIoU.




\subsection{Outdoor Navigation}

We evaluate the benefits of our \ours{} segmentation algorithm for robot navigation in real-world outdoor scenarios using a Clearpath Jackal and a Husky robot mounted with an Intel RealSense camera. The camera has a field of view (FOV) of 70 degrees. We use an Alienware laptop with an Intel i9 CPU (2.90GHz x 12) and an Nvidia GeForce RTX 2080 GPU mounted on the robot to run \ours{}.
We also utilize a velodyne 3D Lidar for robot localization and elevation map generation. 
We test the navigation performance of our segmentation and elevation-based planner on a variety of terrains around a university campus, including concrete walkways, asphalt trails, dirt roads, and off-road terrains. 








The following metrics are used to compare the navigation performance of our method with the Dynamic Window Approach (DWA)~\cite{dwa}, TERP~\cite{terp}, Segformer~\cite{segformer}, and OCRNet~\cite{yuan2020objectcontextual} based planners.

\begin{itemize}
\item \textbf{Success Rate: } The number of successful attempts for the robot to reach its goal while avoiding relatively non-traversable regions and collisions over the total number of trials.

\item \textbf{Trajectory Roughness} - The cumulative vertical motion gradient experienced by the robot along a trajectory.

\item \textbf{Trajectory Selection:} The percentage of the robot's trajectory length along the most navigable surface out of the total trajectory length (i.e., the selection of smoother regions over others).

\item \textbf{False Positive of the Forbidden Region} -  The percent-age of the frames that include a misclassified forbidden region out of all the frames during navigation.

\end{itemize}

\input{Tables/comparison_full}

\begin{figure*}[t]
    \centering

    \begin{subfigure}[c]{0.95\textwidth}
    \includegraphics[width = \textwidth]{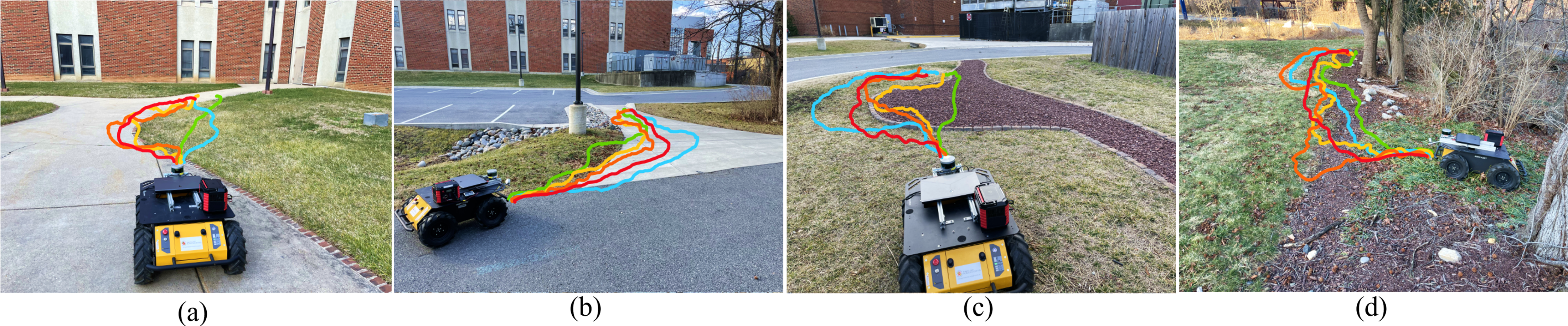}
    \end{subfigure}

    \caption{\textbf{Navigation Results:} Robot trajectories when navigating with Ours (red), Segformer (orange), OCRNet (yellow), TERP (blue), and DWA (green). (a) Walkway; (b) Trail; (c) Dirt road; (d) Off-road:. We observe that the trajectories generated by \ours{} maximize the navigation along the smoothest surface of the available terrains while maintaining the highest success rate.}
    \label{fig:navigation_reuslts}
    \vspace{-6mm}

\end{figure*}

\input{Tables/realworld_perf}
\subsection{Navigation Comparisons}
\label{nav_comparisons}

The navigation results for \ours{}, DWA~\cite{dwa}, TERP~\cite{terp}, Segformer~\cite{segformer}, and OCRNet~\cite{yuan2020objectcontextual} based planners are shown in Table \ref{tab:comparison_table}. We observe that all three segmentation-based planners perform well in terms of success rate under Walkway and Trail scenarios due to the clear margins between the different surfaces. Even though Segformer~\cite{segformer} and OCRNet~\cite{yuan2020objectcontextual} display reasonable performances under scenarios similar to trained datasets, we observe significant performance degradation in terms of trajectory selection during off-road conditions.  However, \ours{} displays significant improvement in terms of all three metrics under cluttered and unseen scenarios in Fig. \ref{fig:navigation_reuslts} (c) and \ref{fig:navigation_reuslts} (d). Specifically, accurate segmentation from \ours{} increases the utilization of the most navigable surface during navigation. This behavior leads to better trajectory selection percentages than the other methods, as shown in the Table \ref{tab:comparison_table}. 

Our baseline planner and TERP incorporate terrain elevation data to perform the navigation task. Hence, such methods cannot differentiate between the surface properties in multi-terrain scenarios with similar elevation conditions. This leads to trajectory generation along less navigable surfaces, in contrast to segmentation-based methods. Therefore, we observe a noticeable degradation in the trajectory smoothness. However, the usage of segmentation data for the baseline planner significantly enhances the navigable trajectory selection capabilities of the algorithm. 

%% file: Tables/ablation3.tex
\begin{table}[t]
\centering
\resizebox{\columnwidth}{!}{%
\begin{tabular}{ccccc}
  \toprule
  \textbf{Dataset} & \textbf{Multi-scale} & \textbf{Fusion method}   & \textbf{mIoU} $\uparrow$ & \textbf{mAcc} $\uparrow$ \\
  \midrule
   \multirow{2}[3]{*}{RUGD} & \ding{55} & -  & 74.32 & 80.41  \\ 
  &\checkmark& Linear & 77.64 & 86.78  \\ 
  &\checkmark& GW-Attn. & 89.08  & 93.55  \\ 
  \midrule
   \multirow{2}[3]{*}{RELLIS-3D} & \ding{55}  & -  & 63.01 & 68.7 \\ 
    &\checkmark&  Linear  & 68.91 & 81.5 \\ 
    &\checkmark&  GW-Attn. & 74.44 & 86.49 \\ 
  \bottomrule
\end{tabular}
}
\caption{\textbf{Ablation studies on multi-scale fusion on RUGD and RELLIS-3D val sets:} 
The ``Multi-scale" column shows whether we use a single or all scale features from the feature extractor, and the ``Fusion method" column shows either we use a simple linear layer or the proposed segmentation head for the fusion.
}

\label{tab:ablation3}

\vspace{-4mm}
\end{table}

%% file: Tables/ablation_full.tex
\begin{table}[t]
\centering
\resizebox{\columnwidth}{!}{%
\begin{tabular}{cccccc}
  \toprule
  \textbf{Dataset} & \textbf{Backbone} & \textbf{GW-Attn.}   & \textbf{mIoU} $\uparrow$ & \textbf{mAcc} $\uparrow$ & \textbf{aAcc} $\uparrow$ \\
  \midrule
   \multirow{5}[6]{*}{RUGD} & \multirow{1}[3]{*}{ResNet50\cite{resnet}} & \ding{55} & 71.55 & 87.67  & 91.63 \\ 
  && \checkmark & 78.22  & 85.09 & 92.91 \\ 
         \cmidrule{2-6}
&  \multirow{1}[3]{*}{BoT~\cite{srinivas2021bottleneck}}   &  \ding{55} & 74.32 & 80.41 & 93.53 \\ 
  && \checkmark & 83.24 & 89.84 & 94.26 \\ 
           \cmidrule{2-6}
&  \multirow{1}[3]{*}{MiT~\cite{segformer}}   & \ding{55}  & 77.64 & 86.78 & 93.47  \\ 
  && \checkmark & 89.08  & 93.55 & 95.66 \\ 
  \midrule
  \multirow{5}[6]{*}{RELLIS-3D} & \multirow{1}[3]{*}{ResNet50\cite{resnet}}  & \ding{55} & 61.82  & 71.55 & 81.72 \\ 
    &&    \checkmark & 63.31 & 71.57 & 83.14\\ 
             \cmidrule{2-6}
      &  \multirow{1}[3]{*}{BoT~\cite{srinivas2021bottleneck}} & \ding{55} & 63.01 & 68.7 & 92.2 \\ 
    &&  \checkmark & 70.63 & 81.32 & 90.23\\ 
             \cmidrule{2-6}
      &  \multirow{1}[3]{*}{MiT~\cite{segformer}}  & \ding{55} & 68.91 & 81.5 & 85.57 \\ 
    &&    \checkmark & 74.44 & 86.49 & 91.69\\ 
  \bottomrule
\end{tabular}
}

\caption{\textbf{Ablation studies on GA heads with different backbones on RUGD and RELLIS-3D val sets:} We observe that the group-wise attention mechanism is successful in improving mIoU by $7.26-11.44\%$ and $4.18-10.31\%$ on the RUGD and RELLIS datasets, respectively, regardless of the backbone choice.
}

\label{tab:ablation}

\vspace{-3mm}
\end{table}

  
  
    
    



%% file: Tables/ablation2_full.tex
\begin{table}[t]
\centering
\resizebox{\columnwidth}{!}{%
\begin{tabular}{cccccc}
  \toprule
  \textbf{Dataset} & \textbf{Spatial Reduction} & \textbf{GW-Attn.}   & \textbf{mIoU} $\uparrow$ & \textbf{mAcc} $\uparrow$ & \textbf{aAcc} $\uparrow$ \\
  \midrule
   \multirow{5}[6]{*}{RUGD} & \multirow{1}[3]{*}{\large{$\frac{1}{8} \times \frac{1}{8}$}} & \ding{55} & 77.64 & 86.78  & 93.47 \\ 
  && \checkmark & 89.08 & 93.55 & 95.66 \\ 
         \cmidrule{2-6}
&  \multirow{1}[3]{*}{\large{$\frac{1}{16} \times \frac{1}{16}$}}   & \ding{55}  & 79.61 & 87.76 & 92.85 \\ 
  && \checkmark & 87.07 & 91.74 & 94.99 \\ 
           \cmidrule{2-6}
&  \multirow{1}[3]{*}{\large{$\frac{1}{32} \times \frac{1}{32}$}}   & \ding{55}  & 76.81 & 85.14 & 92.24  \\ 
  && \checkmark & 84.9  & 90.36 & 94.24 \\ 
  \midrule
   \multirow{5}[6]{*}{RELLIS-3D} & \multirow{1}[3]{*}{\large{$\frac{1}{8} \times \frac{1}{8}$}}  & \ding{55} & 68.91  & 81.5 & 85.57 \\ 
    &&    \checkmark & 71.76 & 81.62 & 90.31\\ 
             \cmidrule{2-6}
       &  \multirow{1}[3]{*}{\large{$\frac{1}{16} \times \frac{1}{16}$}} & \ding{55} & 67.61 & 79.95 & 84.03 \\ 
    &&  \checkmark & 71.62 & 84.51 & 90.26\\ 
             \cmidrule{2-6}
       &  \multirow{1}[3]{*}{\large{$\frac{1}{32} \times \frac{1}{32}$}}  & \ding{55} & 64.91 & 75.56 & 83.51 \\ 
    &&    \checkmark & 70.76 & 84.4 & 90.44\\ 
  \bottomrule
\end{tabular}
}
\caption{\textbf{Ablation studies on GA heads with different spatial reduction on RUGD and RELLIS-3D val sets:} We show that higher spatial reduction can decrease the perception accuracy and the group-wise attention can mitigate the performance drop due to resolution reduction. ``$\frac{1}{N}\times \frac{1}{N}$" in the second column means the bottleneck spatial resolution is reduced by a factor of N on height and width compared to the input resolution.}

\label{tab:ablation2}

\vspace{-6mm}
\end{table}

%% file: Tables/runtime_full.tex
\begin{table}[t]
\centering
\resizebox{\columnwidth}{!}{%
\begin{tabular}{c|cccc}
  \toprule
 \multirow{2}[0]{*}{\textbf{Method}}  & \textbf{Params $\downarrow$} & \multirow{2}[0]{*}{\textbf{GFLOPS $\downarrow$}} & \textbf{Inf Mem $\downarrow$} & \textbf{Run-time $\uparrow$} \\
 & \textbf{(M)} & & \textbf{(MiB)} &\textbf{(img/s)}\\
  \midrule
  
PSPNet \cite{zhao2017pyramid}  & 48.97 & 258.9 & 1635 & 21.97 \\
DeepLabv3+ \cite{deeplabv3+}  & 43.58 & 256.08 & 1443 & 43.48 \\
DANet \cite{fu2019dual}  & 49.82 & 288.81 & 1425 & 29.97 \\
OCRNet \cite{yuan2020objectcontextual} & 36.51 & 221.47 & 1407 & 20.76\\
PSANet \cite{2018psanet}  & 59.13 & 289.52 & 1629 & 28.26 \\
BiseNetv2~\cite{bisenetv2}  & 14.78 & 17.88 & 1137 & 90.33\\
CGNet~\cite{cgnet}  & \underline{0.493} & 5.05 & 1087 & 62.35\\
FastSCNN~\cite{fastscnn}  & 1.45 & \underline{1.35} & \underline{1081} & \underline{110.63}\\
FastFCN~\cite{fastfcn}  & 68.7 & 189.61 & 1707 & 47.83\\
\midrule
*SETR~\cite{setr}  & 309.17 & 312.21 & 2295 & 22.96\\
*DPT~\cite{dpt}  & 109.67 & 255.49 & 1633 & 37.03 \\
*Segformer~\cite{segformer}  & \underline{3.72} & 9.29 & 1143 & \underline{73.49}\\
*Segmenter~\cite{segmenter}  & 6.69 & \underline{6.57} & \underline{1109} & 52.72\\
\midrule
\textbf{*\ours{}-r8}  & 6.94 & 18.69 & 1415 & 65.53\\
\textbf{*\ours{}-r16} & 6.57 & 7.7 & \underline{1121} & 76.20\\
\textbf{*\ours{}-r32}  & \underline{6.21} & \underline{4.54} & 1131 & \underline{76.74}\\
  
  \bottomrule
\end{tabular}
}
\caption{\textbf{Model Complexity:} We show the model parameter size, GLOPS, GPU memory cost during inference, and run-time. For each measurement, we underline one best number among CNN-based methods, one among transformer-based methods, and one among variations of \ours{}.
}

\label{tab:runtime}

\vspace{-2mm}
\end{table}

%% file: Tables/comparison_full.tex
\begin{table*}[t]
\centering
\vspace{7pt}
\ra{1.05}
\Large
\begin{adjustbox}{max width=\textwidth}
\begin{tabular}{l|l|cccccc|cc}
  \toprule[2pt]
  \textbf{Dataset} &
  \textbf{Methods (IoU)} & \textbf{Smooth Region} & \textbf{Rough Region} & \textbf{Bumpy Region} &  \textbf{Forbidden Region} & \textbf{Obstacle} & \textbf{Background} & \textbf{mIoU} $\uparrow$ & \textbf{aAcc} $\uparrow$\\
\hline
     \multirow{13}[6]{*}{RUGD} & PSPNet \cite{zhao2017pyramid} & 48.62 & 88.92 & 69.45 & 29.07 & 87.98 & 78.29 & 67.06 & 92.85 \\
     & DeepLabv3+ \cite{deeplabv3+} & 5.86 & 84.99 & 50.40 & 25.04 & 87.50 & \underline{\textbf{81.47}} & 55.88 & 91.51\\
     & DANet \cite{fu2019dual} & 2.26 & 81.47 & 8.69 & 15.00 & 82.54 & 74.86 & 44.14 & 88.81 \\
     & OCRNet \cite{yuan2020objectcontextual} & 66.29 & 89.47 & 76.15 & 59.14 & 88.77 & 79.17 & 76.50 & 93.46 \\
     & PSANet \cite{2018psanet} & 34.92 & 87.70 & 35.64 & 8.66 & 86.95 & 78.97 & 55.47 & 92.13 \\
     
     & BiseNetv2~\cite{bisenetv2} & 24.27 & 89.99 & \underline{\textbf{89.99}} & 83.31 & 90.93 & 75.29 & 75.1 & 93.4  \\
     & CGNet~\cite{cgnet} & 40.84 & 90.39 & 85.67 & 76.21 & 89.75 & 74.48 & 76.22 & 93.29  \\
     & FastSCNN~\cite{fastscnn} & 83.03 & 92.82 & 87.69 & 81.05 & 90.94 & 75.11 & 85.11 & 94.77 \\
     & FastFCN~\cite{fastfcn}  & 26.27 & 89.85 & 85.95 & \underline{84.13} & \underline{91.23} & 75.63 & 75.51 & 93.46 \\
     & *SETR~\cite{setr} & 89.77 & 92.46 & 84.58 & 70.33 & 89.55 & 70.47 & 82.86 & 94.09 \\
     & *DPT~\cite{dpt} & 1.04 & 81.23 & 22.98 & 25.84 & 89.18 & 74.5 & 49.13 & 88.77 \\
     & *Segformer~\cite{segformer} & \underline{93.26} & \underline{93.16} & 87.56 & 77.31 & 91.2 & 78.5 & \underline{86.83} & \underline{95.17}  \\
     & *Segmenter~\cite{segmenter} & 90.39 & 91.17 & 83.96 & 65.43 & 87.8 & 68.17 & 81.15 & 93.22  \\
            \cmidrule{2-10}
     & \textbf{*\ours{}-r8} & \underline{\textbf{95.15}} & \underline{\textbf{94.45}} & \underline{89.83} & \underline{\textbf{86.25}} & \underline{\textbf{91.95}} & \underline{76.86} & \underline{\textbf{89.08}} & \underline{\textbf{95.66}}  \\
    & \textbf{*\ours{}-r16} &  93.68 & 93.95 & 88.73 & 83.86 & 90.86 & 71.36 & 87.07 & 94.99 \\
    & \textbf{*\ours{}-r32} & 92.76 & 93.28 & 87.44 & 79.9 & 89.55 & 66.46 & 84.9 & 94.24 \\

  \midrule
     \multirow{13}[6]{*}{RELLIS-3D} & PSPNet \cite{zhao2017pyramid} & 69.21 & 80.99 & 8.89 & 53.7 & 60.7 & 94.67 & 61.36 & 86.01 \\
     & DeepLabv3+ \cite{deeplabv3+} & 65.76 & 79.84 & 19.72 & 47.52 & 64.88 & 95.92 & 62.27 & 85.84\\
     & DANet \cite{fu2019dual} & 72.93 & \underline{85.18} & 13.10 & \underline{60.60} & 70.53 & \underline{95.95} & 66.38 & \underline{89.11}\\
     & OCRNet \cite{yuan2020objectcontextual} & \underline{74.67} & 83.04 & 27.76 & 60.44 & 62.35 & 92.58 & 66.81 & 86.95\\
     & PSANet \cite{2018psanet} & 64.06 & 75.29 & 17.08 & 47.45 & 61.74 & 94.31 & 59.99 & 83.17 \\
     
     & BiseNetv2~\cite{bisenetv2} & 65.56 & 73.24 & 39.35 & 48.17 & \underline{71.91} & 93.78 & 65.33 & 83.03    \\
     & CGNet~\cite{cgnet} & 62.84 & 74.17 & 49.57 & 45.41 & 68.88 & 94.53 & 65.9 & 82.7 \\
     & FastSCNN~\cite{fastscnn} & 67.06 & 77.6 & \underline{\textbf{56.49}} & 49.76 & 70.31 & 94.43 & \underline{69.27} & 84.51 \\
     & FastFCN~\cite{fastfcn} & 70.51 & 79.15 & 49.72 & 51.37 & 63.9 & 94.82 & 68.24 & 84.1 \\            
     & *SETR~\cite{setr} & 65.37 & 78.64 & 40.89 & 52.59 & 63.8 & 91.87 & 65.53 & 83.59 \\
     & *DPT~\cite{dpt} & 5.42 & 76.65 & 47.13 & 54.87 & 62.74 & 85.5 & 55.38 & 81.61 \\
     & *Segformer~\cite{segformer} & 60.28 & 79.78 & 53.35 & 53.78 & 70.15 & 94.37 & 68.62 & 85.37\\
     & *Segmenter~\cite{segmenter} & 51.67 & 78.4 & 19.38 & 42.61 & 66.04 & 92.05 & 58.36 & 82.16\\
            \cmidrule{2-10}
     & \textbf{*\ours{}-r8} & \underline{\textbf{78.5}} & \underline{\textbf{88.25}} & \underline{37.28} & \underline{\textbf{72.34}} & \underline{\textbf{74.75}} & \underline{
\textbf{96.07}} & \underline{\textbf{74.44}} & \underline{\textbf{91.69}}  \\
    & \textbf{*\ours{}-r16} &  78.37 & 85.58 & 28.63 & 67.55 & 73.82 & 95.73 & 71.62 & 90.26\\
    & \textbf{*\ours{}-r32} & 76.73 & 86.86 & 23.49 & 71.58 & 71.24 & 94.65 & 70.76 & 90.44 \\

  \bottomrule[2 pt]
\end{tabular}

\end{adjustbox}
\caption{\textbf{Comparison with state-of-the-art methods on RUGD and RELLIS-3D:} We compare the performance of our method with transformer-based methods (with *) as well as other methods in terms of IoU and aAcc. \ours{} improves the state-of-the-art mIoU by $2.25-44.94\%$ on RUGD and $5.17-19.06\%$ on RELLIS-3D. Further, we show that our grouping method can improve the accuracy of classes like smooth region, bumpy region, and forbidden regions by large margins, thus improving safe navigation. We bold the best number and underline one best result from our method and one from other methods. The suffix of our method shows different designs of our network. For example, ``r8" means we reduce height and width of the spatial resolution by a factor of 8 at the spatial alignment step.}
\label{tab:comparison}
\end{table*}

%% file: Tables/realworld_perf.tex
\begin{table}[t]
\resizebox{\columnwidth}{!}{%
\begin{tabular}{c c c c c c} 
\toprule
\textbf{Metrics} & \textbf{Method} &  \textbf{Walkway} & \textbf{Trail} & \textbf{Dirt road} & \textbf{Off-road} \\ 
[0.5ex] 
\hline
\multirow{5}{*}{\rotatebox[origin=c]{0}{\makecell{\textbf{Success Rate (\%) $\uparrow$ }}}} 
 & DWA \cite{dwa} & 67 & 62 & 64 & 46   \\
 & TERP \cite{terp} & 79 & 73 & 79 & 62 \\
  & OCRNet \cite{yuan2020objectcontextual} & 75 & 69 & 70 & 62 \\
 & Segformer \cite{segformer} & 77 & 73 & \textbf{80} & 58 \\
 & \ours{} (ours) & \textbf{86} & \textbf{77} & 79 & \textbf{71} \\
\hline



\multirow{5}{*}{\rotatebox[origin=c]{0}{\makecell{\textbf{Traj. Roughness $\downarrow$}}}}  
 & DWA \cite{dwa}  & 0.186 & 0.254 & 0.318 & 0.952   \\
 & TERP \cite{terp} & 0.115 & 0.172 & 0.272 & 0.843 \\
 & OCRNet \cite{yuan2020objectcontextual} & 0.109 & 0.483 & 0.234 & 1.182 \\
  & Segformer \cite{segformer} & 0.124 & 0.233 & 0.346 & 1.038 \\
 & \ours{} (ours) & \textbf{0.098} & \textbf{0.164} & \textbf{0.211} & \textbf{0.725} \\

\hline

\multirow{5}{*}{\rotatebox[origin=c]{0}{\makecell{\textbf{Traj. Selection (\%) $\uparrow$}}}} 
 & DWA \cite{dwa}  & 05 & 48 & 23 & 14   \\
 & TERP \cite{terp} & 03 & 94 & \textbf{96} & 32 \\
 & OCRNet \cite{yuan2020objectcontextual} & 51 & 58 & 17 & 08 \\
 & Segformer \cite{segformer} & 46 & 62 & 57 & 72 \\
& \ours{} (ours) & \textbf{98} & \textbf{96} & 94 & \textbf{81} \\

\hline

\multirow{3}{*}{\rotatebox[origin=c]{0}{\makecell{\textbf{False Positive of} \\\textbf{Forbidden Region (\%) $\downarrow$}}}} 

 & OCRNet \cite{yuan2020objectcontextual} & 39.13 & 42.38 & 52.14 & 56.14 \\
 & Segformer \cite{segformer} & 30.84 & 35.64 & 46.57 & 51.26 \\
& \ours{} (ours) & \textbf{0.88} & \textbf{1.23} & \textbf{3.17} & \textbf{7.85} \\

\bottomrule

\end{tabular}
}
\caption{\small{\textbf{Navigation Comparisons:} We compare the performance of our method with state-of-the-art algorithms on four different scenarios, \rt{each with at least ten trials}. \ours{} consistently outperforms other methods in terms of success rate, trajectory roughness, and false positive rate of forbidden regions even in challenging \rt{unstructured} environments. We observe an \rt{average} increase of \rt{$10\%$} in terms of success rate, \rt{$43.5\%$} in terms of the best navigable trajectory selection and \rt{an average} decrease of \rt{$10.82\%$} in trajectory roughness. Further, \ours{} reduces the false positive rate of forbidden regions by $37.79\%$.}
}
\label{tab:comparison_table}
\vspace{-6mm}
\end{table}

%% file: sections/conclusions.tex
\section{Conclusions, Limitations, and Future Work}

We present a learning-based method for classifying various terrain types in off-road environments. We propose a novel and efficient segmentation network that can provide robust predictions and distinguish regions with different semantic meanings using group-wise attention. We demonstrate improvement in both accuracy and efficiency over SOTA methods on complex, unstructured terrain datasets and show that our method can be used in real-world scenarios and result in higher success rate for navigation.

Our approach has some limitations. For example, our current work focuses on perception and integrates with a learning-based planner implementation for real-world scenarios. As a result, our current performance analysis is mostly based on perception metrics and not takes into account many navigation metrics. In addition, we need to make sure that the data is labeled correctly and consistently. For example, trees can either be obstacles or background objects. An annotated dataset might not label them separately.

As part of our future work, we need to consider how to evaluate the accuracy of our method with more navigation metrics. We want to explore how this learning method can be applied and customized to other planning and navigation schemes and extend our approach to multiple sensor inputs for designing a more robust navigation scheme. 

\noindent\textbf{Acknowledgement.} This research was supported by Army Cooperative Agreement No. W911NF2120076 and ARO grant W911NF2110026.

%% file: sections/dataset.tex

\section{Unstructured Environments: Levels of Navigability}
\label{groupings}


There are certain characteristics, \textit{e.g.}, texture, color, temperature, etc., that highlight the differences between various terrains. For robot navigation, it is important to identify different terrains based on the navigability of that terrain, which is indicated by its texture~\cite{outdoor,off_road_class1,off_road_trav1}. Furthermore, there may be terrains with similar textures and navigability that similarly affect navigation capabilities but may pose challenges to visual perception systems such as semantic segmentation (due to the long-tailed distribution problem). In such cases, it is advantageous to group terrains with similar navigability into a single category. Given a dataset with $C$ different semantic classes, we regroup them into $G$ classes using the following criteria.

Terrain navigability varies for different robot systems. For instance, large autonomous vehicles may be better able to navigate on dirt roads than smaller mobile robots that would find such dirt roads difficult to traverse. Therefore, our approach is generally designed to identify different groupings of terrain categories and is not restricted to a fixed set of groupings. In Table~\ref{tab:grouping}, we show an example of a grouping based on terrain texture and other characteristics. Below, we describe in detail the characteristics of different terrains based on their texture and navigability.

\input{Tables/grouping}




\begin{itemize}
    \item \textit{Concrete, Asphalt (Smooth)}:  Concrete and asphalt are smooth terrains commonly found in urban roads. These terrains correspond to the ``flat'' category in the grouping provided by the CityScapes~\cite{cityscape} autonomous driving dataset. These terrains are navigable in outdoor environments~\cite{outdoor}, and most mobile robots should be able to navigate in these terrains. 
    
    \item \textit{Gravel, Grass, Dirt, Sand (Rough)}: These terrains have been termed rough on account of the increased friction encountered while traversing them~\cite{off_road_class1, off_road_class2}.
    Most existing perception modules in off-road environments~\cite{off_road_trav1, pointcloud_hazard} either do not consider these factors or may conservatively avoid such regions~\cite{outdoor, robot_nav} resulting in a sub-optimal solution, particularly in off-road navigation environments.
    To handle a large class of navigation methods,
    we want to be able to explicitly distinguish between smooth surfaces and rough surfaces. For example, in the presence of a smooth navigable region, a planning algorithm could prioritize it over a rough navigable region to reduce energy loss. 
    
    \item \textit{Small rocks, Rock-bed (Bumpy)}: Autonomous vehicles or large robots may be able to navigate through rocks, while smaller robots with weaker off-road capabilities may face issues in such terrains~\cite{outdoor, off_road_trav1, pointcloud_hazard, lidar_obstacle}.
    We provide a safe and flexible option where the planning scheme can be customized according to different scenes and different hardware characteristics of the robot. Specifically, this level of grouping can be ignored for off-road robots or vehicles by re-allocating these terrains to a different navigability group such as rough (for larger robots) or forbidden (for smaller robots). Our approach is general and handles dynamic groupings.
    
    \item \textit{Water, Bushes, etc. (Forbidden Regions)}: These are regions that the agent must avoid to prevent damage to the robot hardware.
    
    \item \textit{Obstacles}: Detecting obstacles such as trees, poles, etc. is critical for safe navigation. There has been a lot of work \cite{lidar_obstacle, pointcloud_hazard, off_road_trav1} on obstacle and hazardous terrain detection. 
    
    \item \textit{Background}: We use this buffer group to include background and non-navigable classes, including void, sky, and sign, that are not commensurate with any of the earlier definitions. 

\end{itemize}

%% file: Tables/grouping.tex
\begin{table}[b]
    \centering
    \vspace{5pt}
    \caption{\textbf{Texture-based Terrain Classification:} We show an example of a grouping of terrain classes based on their texture and navigability. Our approach is general and designed for dynamic groupings in which different terrains may be re-classified into different hierarchy levels depending on the type of robot.} 
    \begin{tabular}{ccc}
    \toprule
    \multicolumn{2}{c}{Hierarchy level} & Classes \\
    \midrule 
     \multirow{2}[3]{*}{Navigable}  & Smooth   &  Concrete, Asphalt\\
       & Rough   &  Gravel, Grass, Dirt, Sand\\
       & Bumpy  & Rock, Rock Bed \\
    \midrule 
      \multicolumn{2}{c}{Forbidden}  & Water, Bushes, Tall Vegetation\\
    \midrule 
      \multicolumn{2}{c}{Obstacles} & Trees, Poles, Logs, etc. \\
    \midrule 
      \multicolumn{2}{c}{Background} & Void, Sky, Sign\\
       \bottomrule
    \end{tabular}

    \vspace{-10pt}
    \label{tab:grouping}
\end{table}

%% file: QualitativeResults/qualitative2.tex
\begin{figure*}[p]
    \centering
    \begin{subfigure}[b]{0.15\textwidth}
    \includegraphics[width = \textwidth]{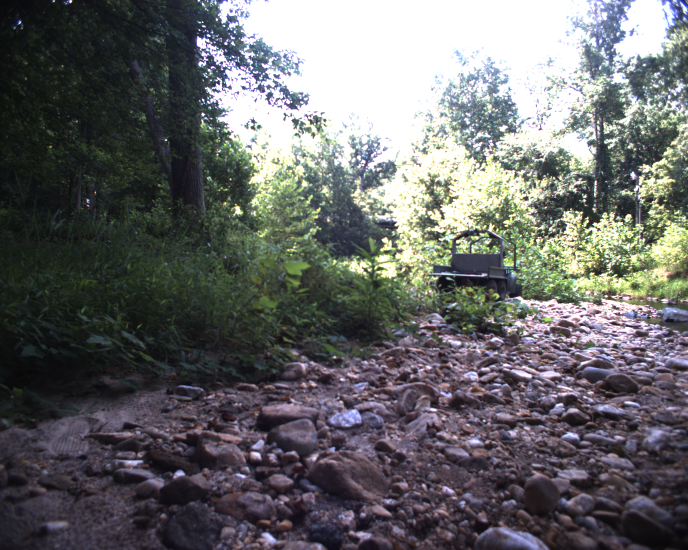}
    \caption{creek-1}
    \end{subfigure}
    \begin{subfigure}[b]{0.15\textwidth}
    \includegraphics[width = \textwidth]{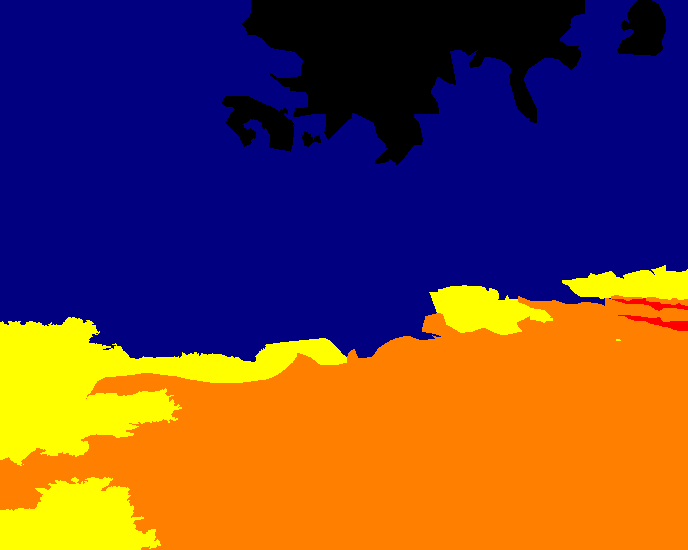}
    \caption{GT}
    \end{subfigure}
    \begin{subfigure}[b]{0.15\textwidth}
    \includegraphics[width = \textwidth]{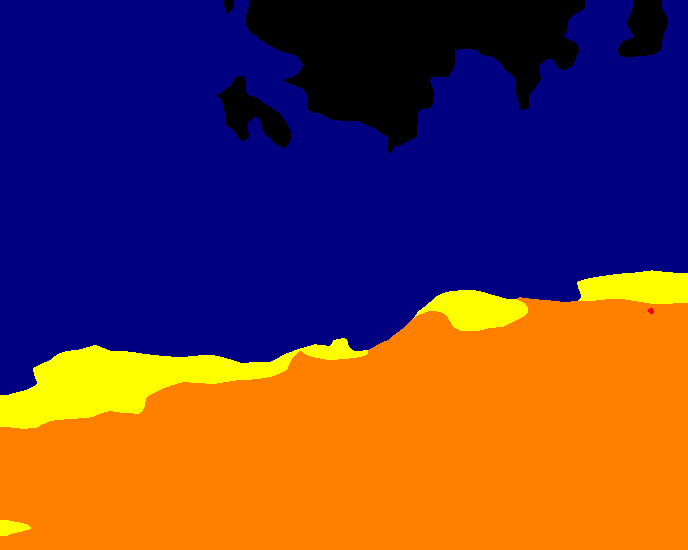}
    \caption{Ours}
    \end{subfigure}
    \begin{subfigure}[b]{0.15\textwidth}
    \includegraphics[width = \textwidth]{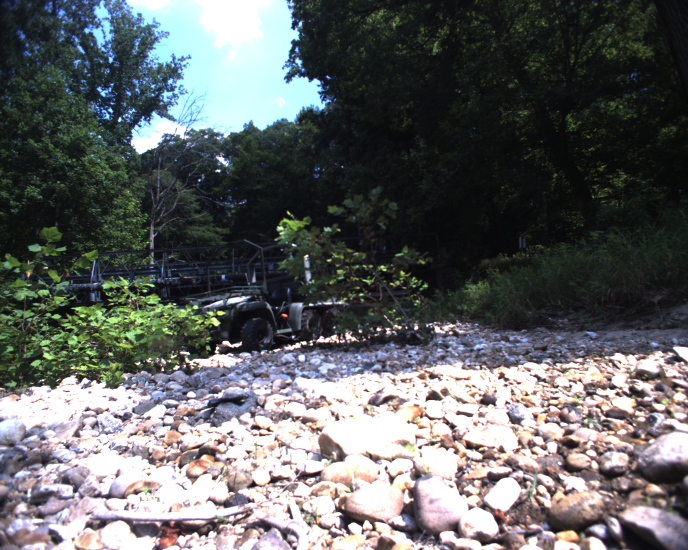}
    \caption{creek-2}
    \end{subfigure}
    \begin{subfigure}[b]{0.15\textwidth}
    \includegraphics[width = \textwidth]{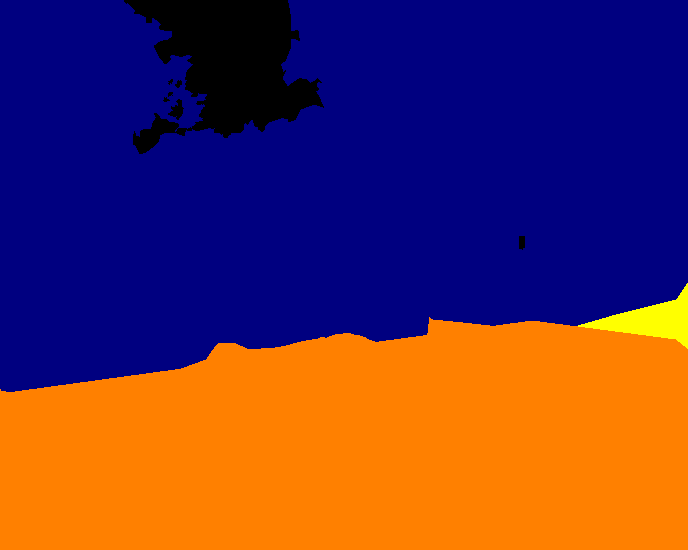}
    \caption{GT}
    \end{subfigure}
    \begin{subfigure}[b]{0.15\textwidth}
    \includegraphics[width = \textwidth]{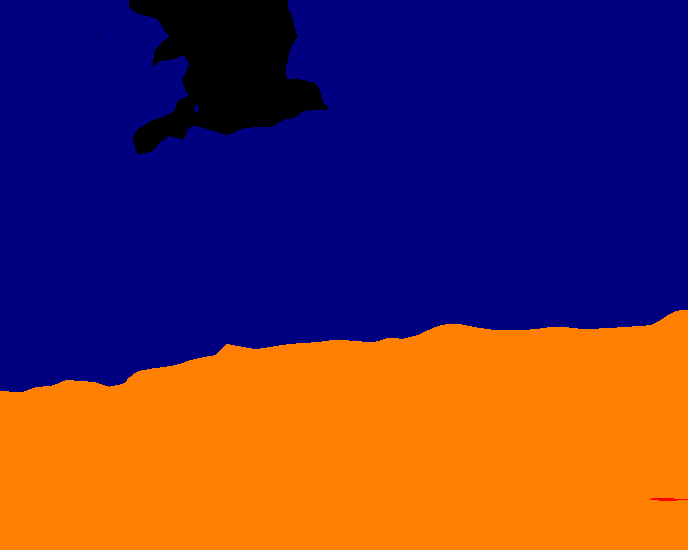}
    \caption{Ours}
    \end{subfigure}          
     \\
    \begin{subfigure}[b]{0.15\textwidth}
    \includegraphics[width = \textwidth]{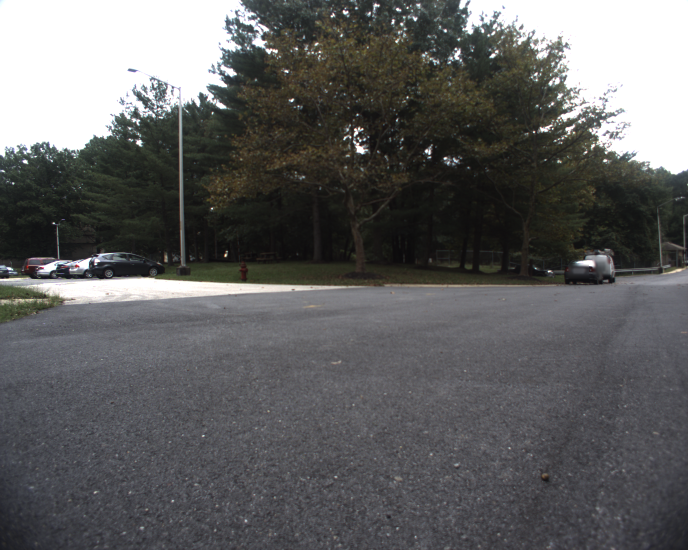}
    \caption{park-1}
    \end{subfigure}
    \begin{subfigure}[b]{0.15\textwidth}
    \includegraphics[width = \textwidth]{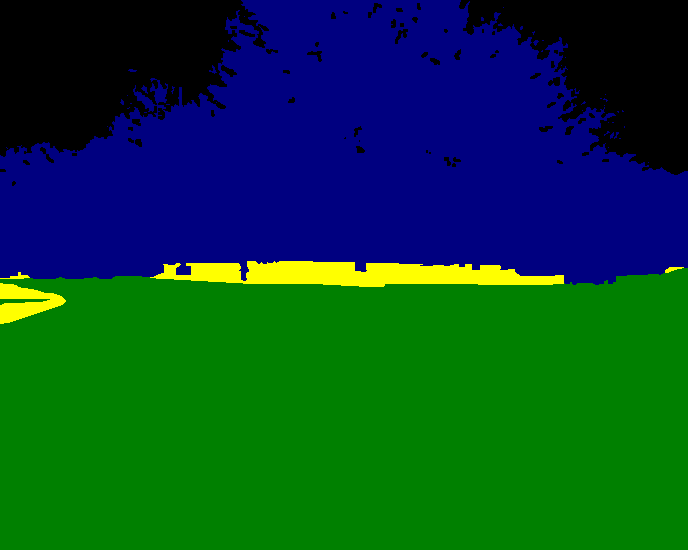}
    \caption{GT}
    \end{subfigure}
    \begin{subfigure}[b]{0.15\textwidth}
    \includegraphics[width = \textwidth]{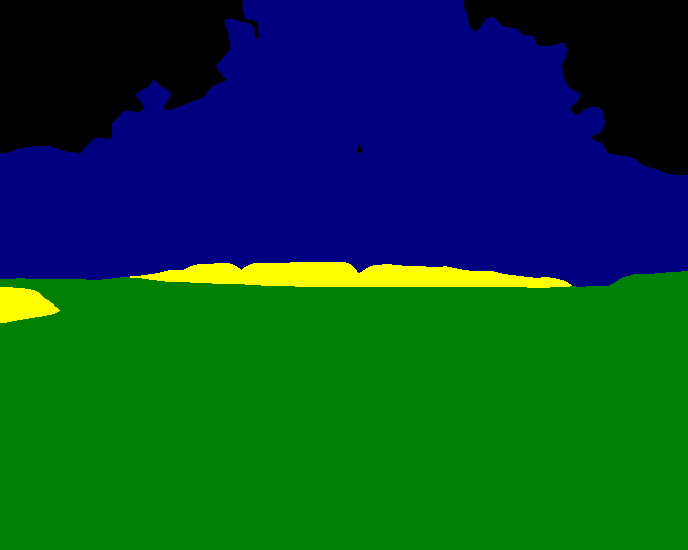}
    \caption{Ours}
    \end{subfigure}
    \begin{subfigure}[b]{0.15\textwidth}
    \includegraphics[width = \textwidth]{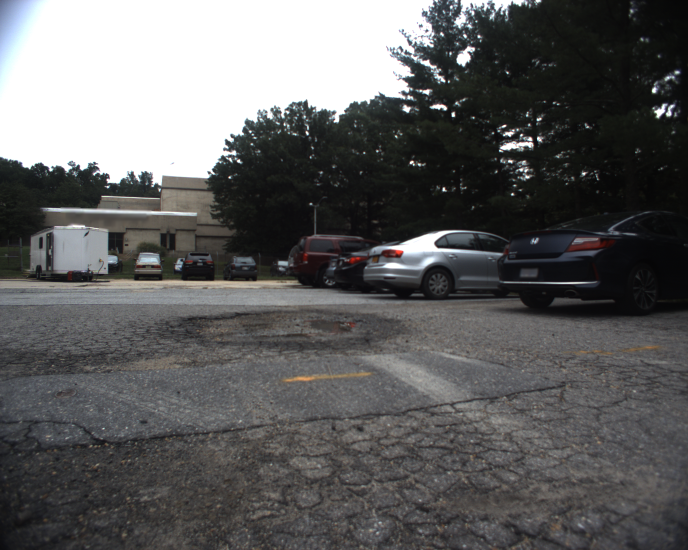}
    \caption{park-2}
    \end{subfigure}
    \begin{subfigure}[b]{0.15\textwidth}
    \includegraphics[width = \textwidth]{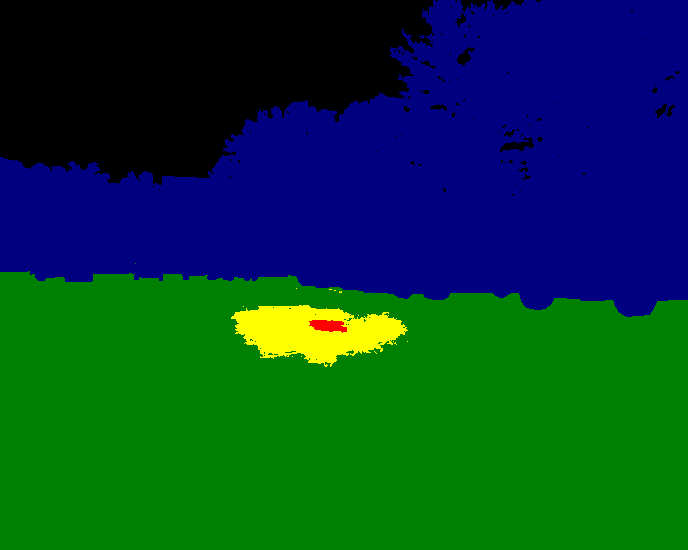}
    \caption{GT}
    \end{subfigure}
    \begin{subfigure}[b]{0.15\textwidth}
    \includegraphics[width = \textwidth]{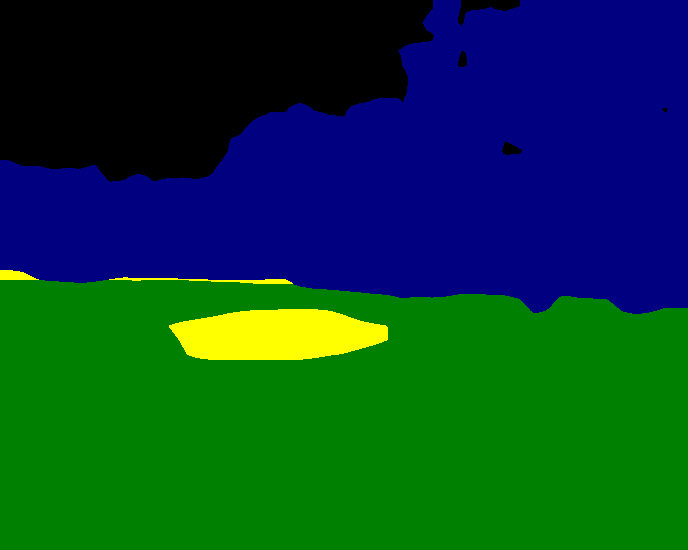}
    \caption{Ours}
    \end{subfigure}    
    \\
    \begin{subfigure}[b]{0.15\textwidth}
    \includegraphics[width = \textwidth]{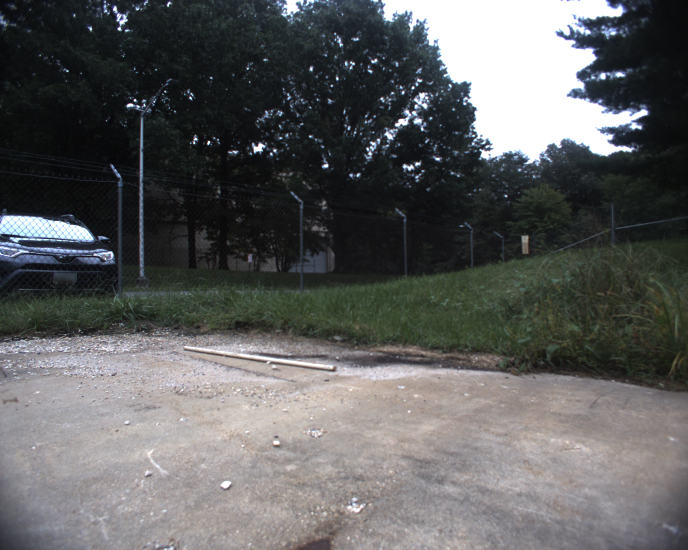}
    \caption{park-3}
    \end{subfigure}
    \begin{subfigure}[b]{0.15\textwidth}
    \includegraphics[width = \textwidth]{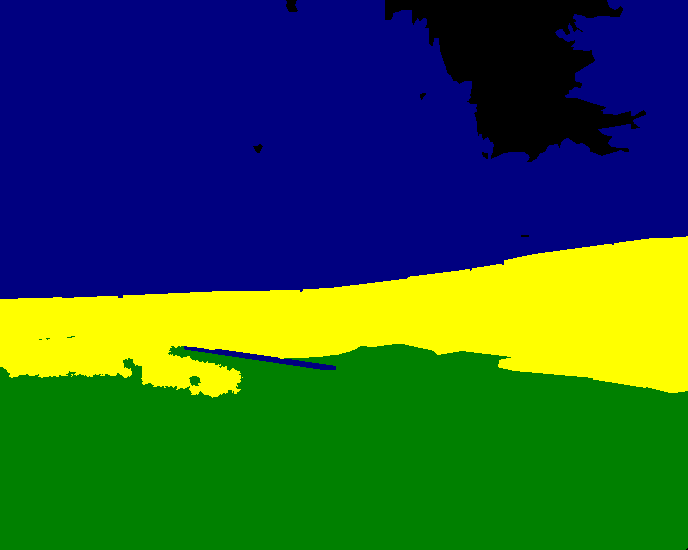}
    \caption{GT}
    \end{subfigure}
    \begin{subfigure}[b]{0.15\textwidth}
    \includegraphics[width = \textwidth]{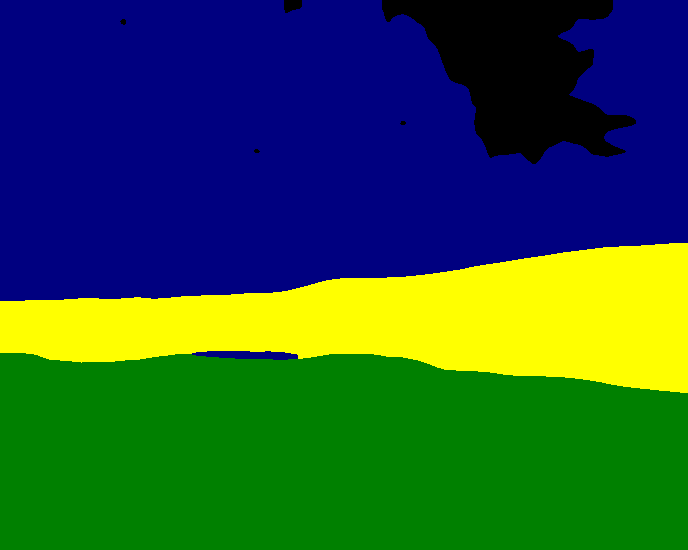}
    \caption{Ours}
    \end{subfigure}
    \begin{subfigure}[b]{0.15\textwidth}
    \includegraphics[width = \textwidth]{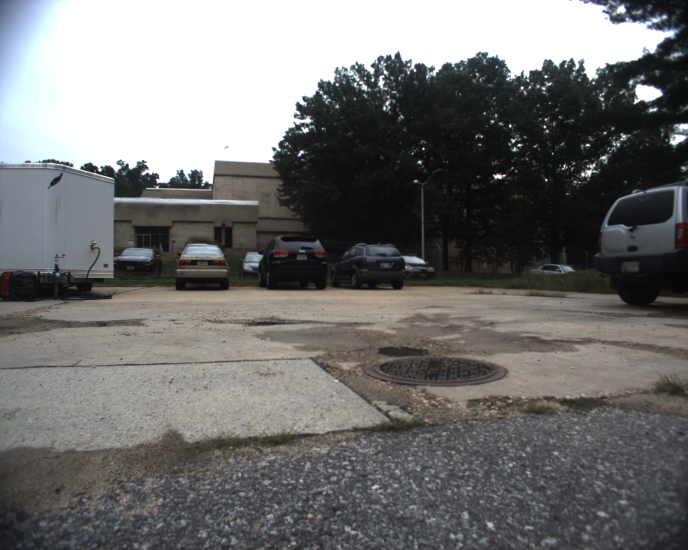}
    \caption{park-4}
    \end{subfigure}
    \begin{subfigure}[b]{0.15\textwidth}
    \includegraphics[width = \textwidth]{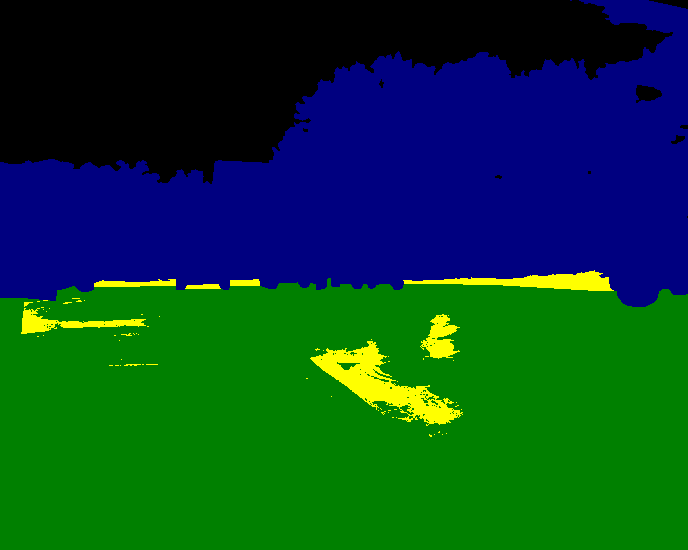}
    \caption{GT}
    \end{subfigure}
    \begin{subfigure}[b]{0.15\textwidth}
    \includegraphics[width = \textwidth]{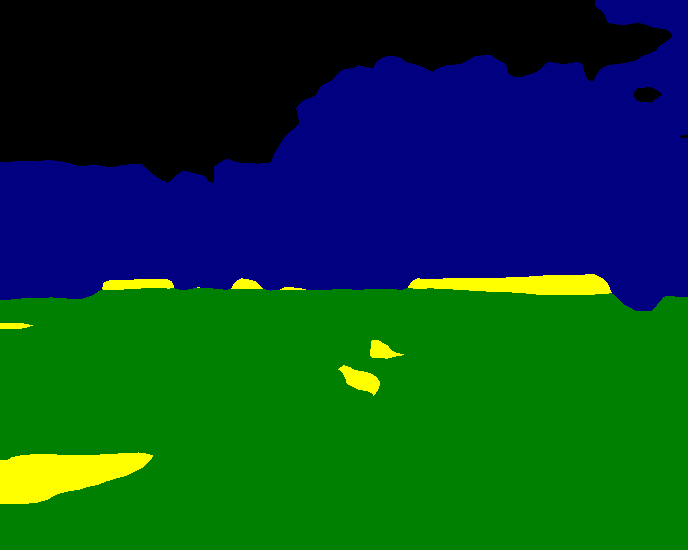}
    \caption{Ours}
    \end{subfigure}
    \\
    \begin{subfigure}[b]{0.15\textwidth}
    \includegraphics[width = \textwidth]{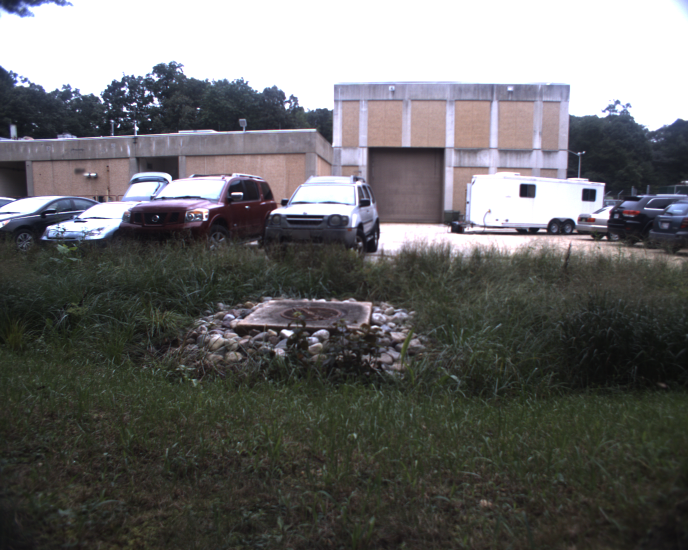}
    \caption{park-5}
    \end{subfigure}
    \begin{subfigure}[b]{0.15\textwidth}
    \includegraphics[width = \textwidth]{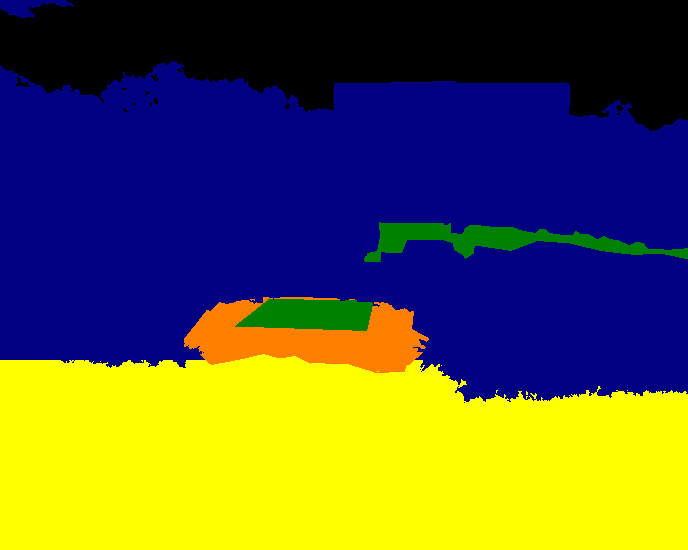}
    \caption{GT}
    \end{subfigure}
    \begin{subfigure}[b]{0.15\textwidth}
    \includegraphics[width = \textwidth]{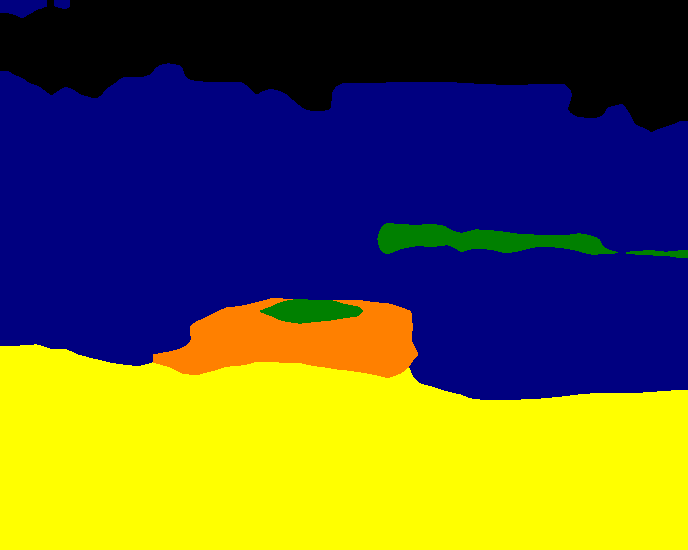}
    \caption{Ours}
    \end{subfigure}  
    \begin{subfigure}[b]{0.15\textwidth}
    \includegraphics[width = \textwidth]{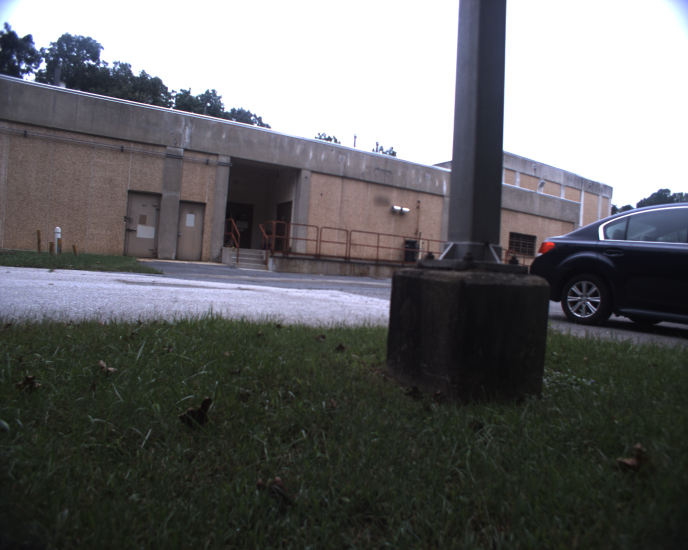}
    \caption{park-6}
    \end{subfigure}
    \begin{subfigure}[b]{0.15\textwidth}
    \includegraphics[width = \textwidth]{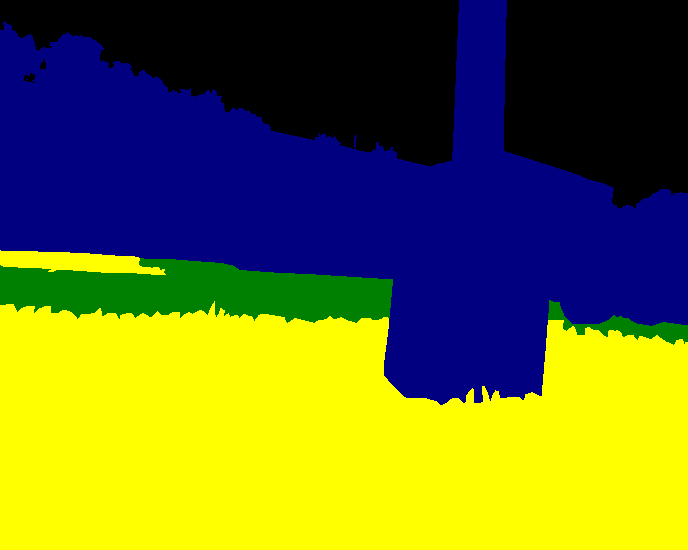}
    \caption{GT}
    \end{subfigure}
    \begin{subfigure}[b]{0.15\textwidth}
    \includegraphics[width = \textwidth]{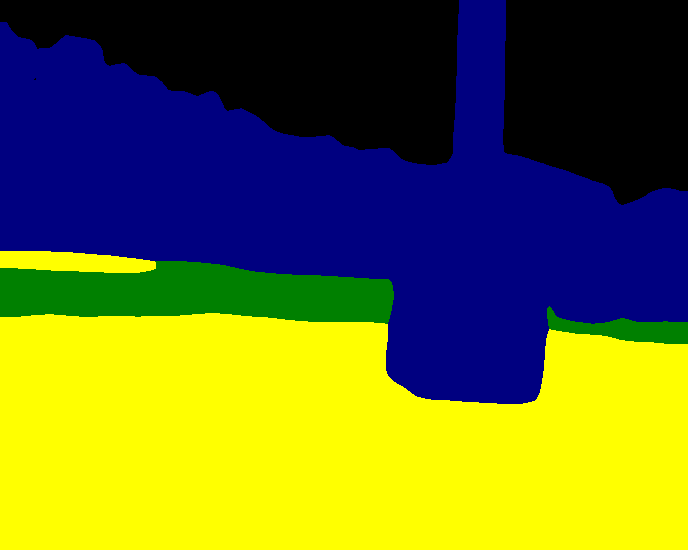}
    \caption{Ours}
    \end{subfigure}    
    \\
    \begin{subfigure}[b]{0.15\textwidth}
    \includegraphics[width = \textwidth]{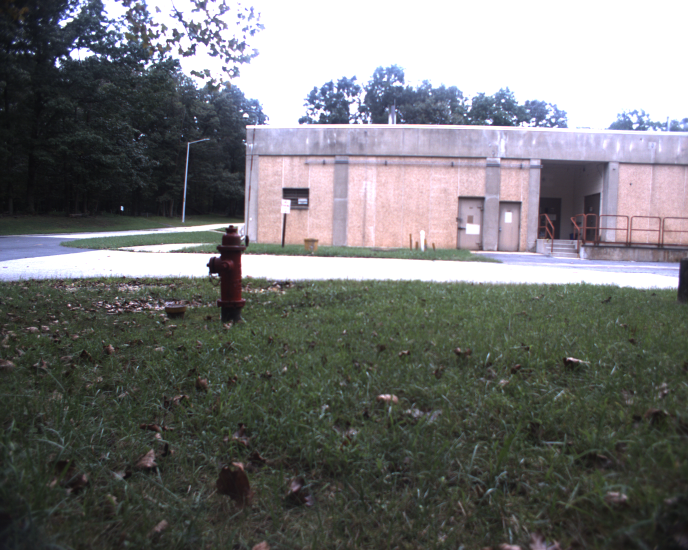}
    \caption{park-7}
    \end{subfigure}
    \begin{subfigure}[b]{0.15\textwidth}
    \includegraphics[width = \textwidth]{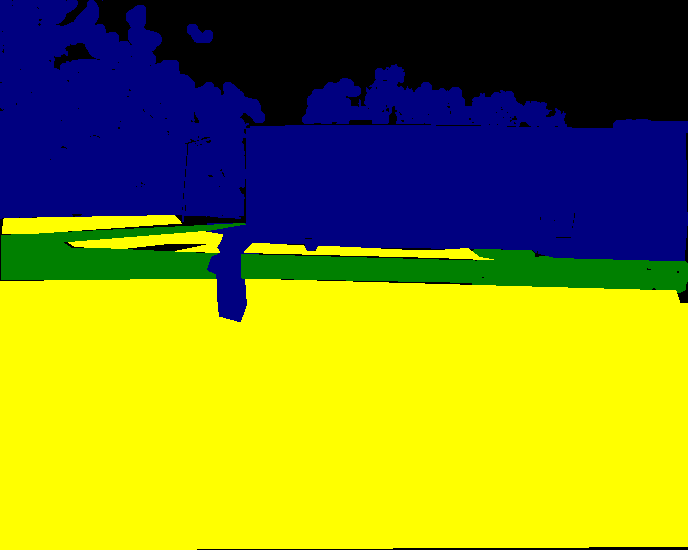}
    \caption{GT}
    \end{subfigure}
    \begin{subfigure}[b]{0.15\textwidth}
    \includegraphics[width = \textwidth]{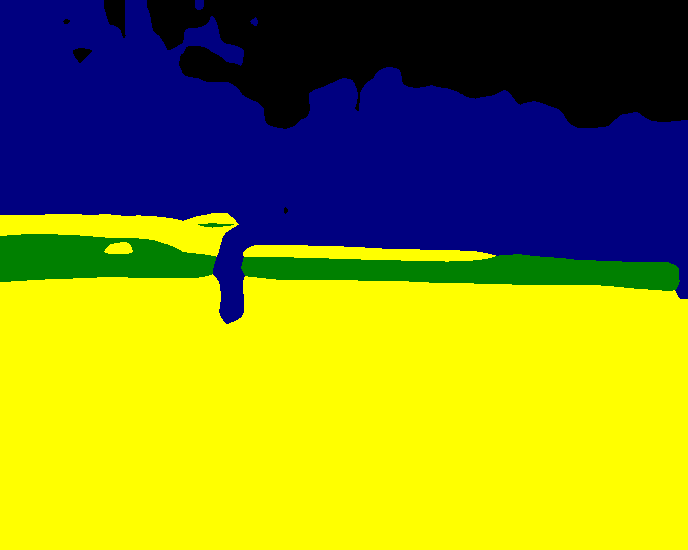}
    \caption{Ours}
    \end{subfigure}
    \begin{subfigure}[b]{0.15\textwidth}
    \includegraphics[width = \textwidth]{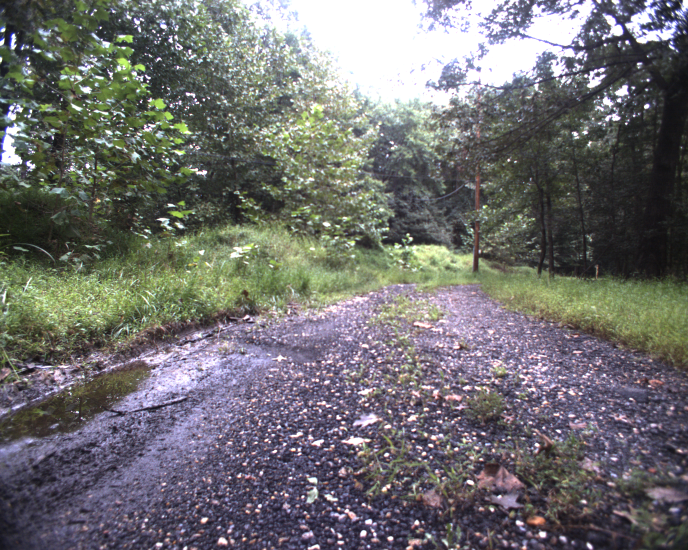}
    \caption{trail-1}
    \end{subfigure}
    \begin{subfigure}[b]{0.15\textwidth}
    \includegraphics[width = \textwidth]{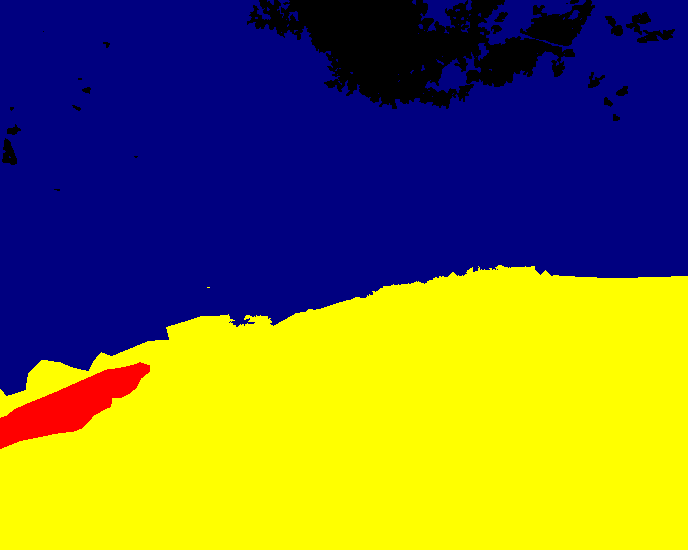}
    \caption{GT}
    \end{subfigure}
    \begin{subfigure}[b]{0.15\textwidth}
    \includegraphics[width = \textwidth]{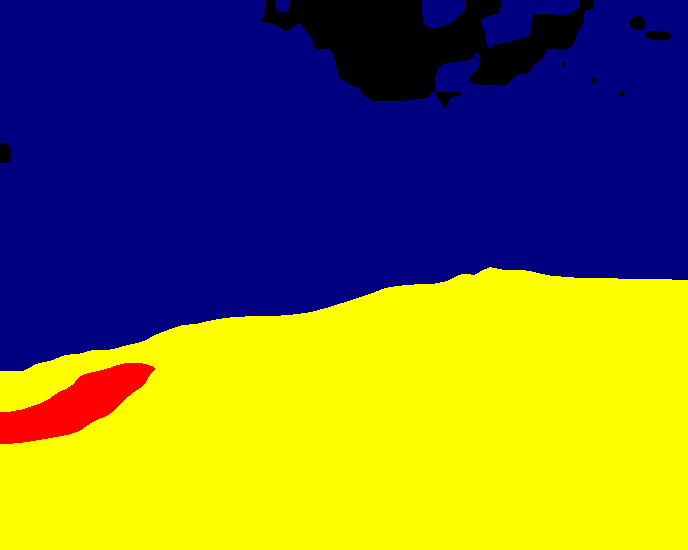}
    \caption{Ours}
    \end{subfigure}
    \\
    \begin{subfigure}[b]{0.15\textwidth}
    \includegraphics[width = \textwidth]{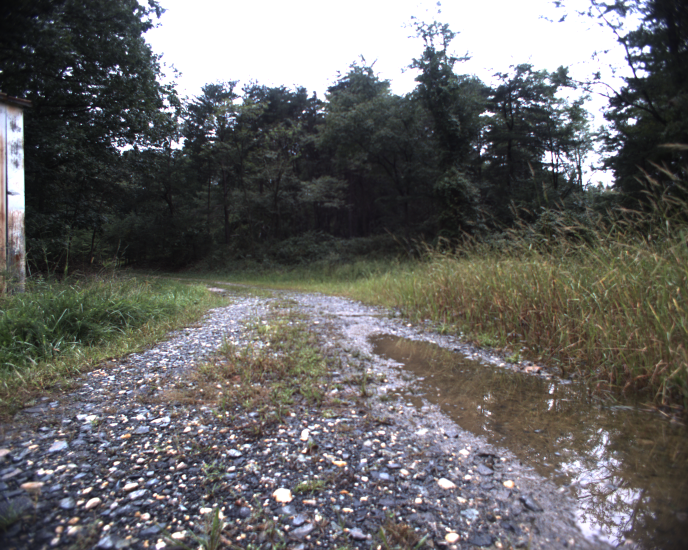}
    \caption{trail-2}
    \end{subfigure}
    \begin{subfigure}[b]{0.15\textwidth}
    \includegraphics[width = \textwidth]{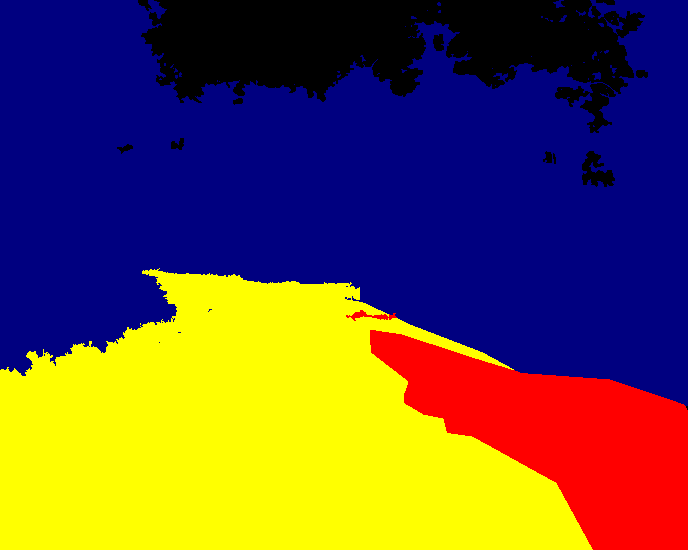}
    \caption{GT}
    \end{subfigure}
    \begin{subfigure}[b]{0.15\textwidth}
    \includegraphics[width = \textwidth]{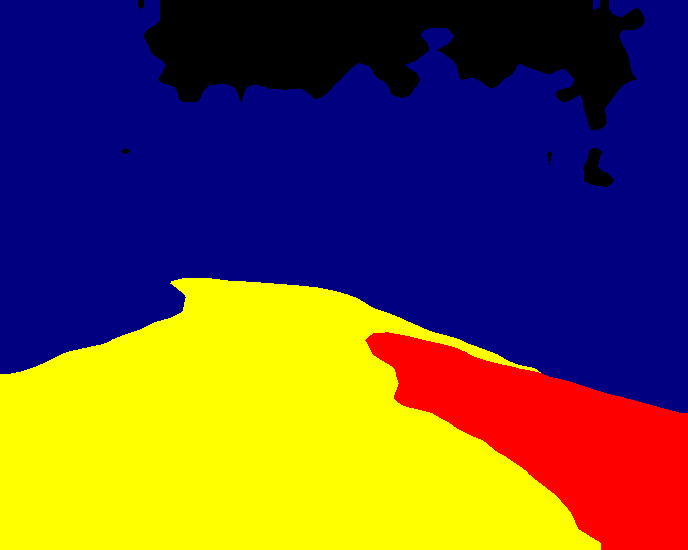}
    \caption{Ours}
    \end{subfigure}    
    \begin{subfigure}[b]{0.15\textwidth}
    \includegraphics[width = \textwidth]{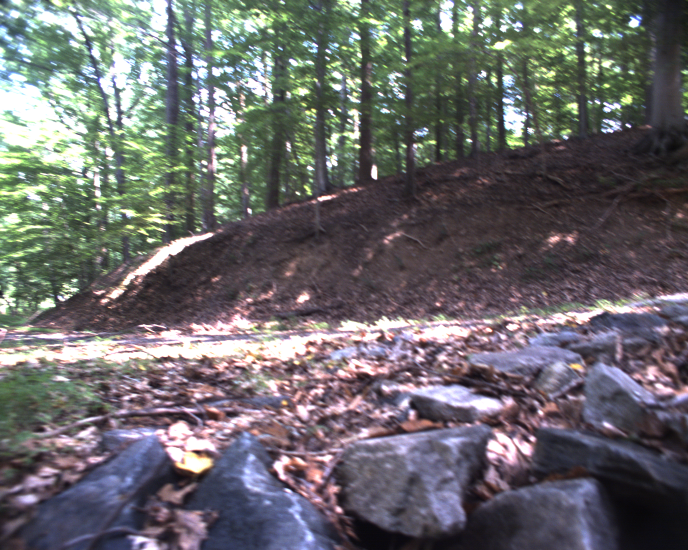}
    \caption{trail-3}
    \end{subfigure}
    \begin{subfigure}[b]{0.15\textwidth}
    \includegraphics[width = \textwidth]{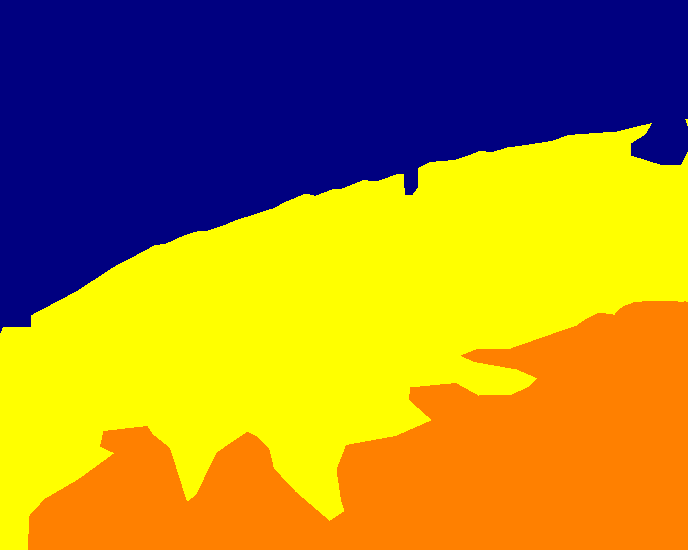}
    \caption{GT}
    \end{subfigure}
    \begin{subfigure}[b]{0.15\textwidth}
    \includegraphics[width = \textwidth]{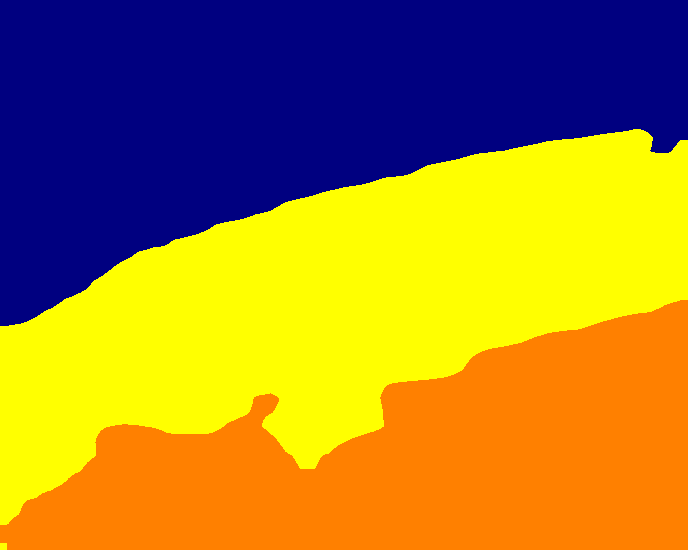}
    \caption{Ours}
    \end{subfigure}       
     \\
    \begin{subfigure}[b]{0.15\textwidth}
    \includegraphics[width = \textwidth]{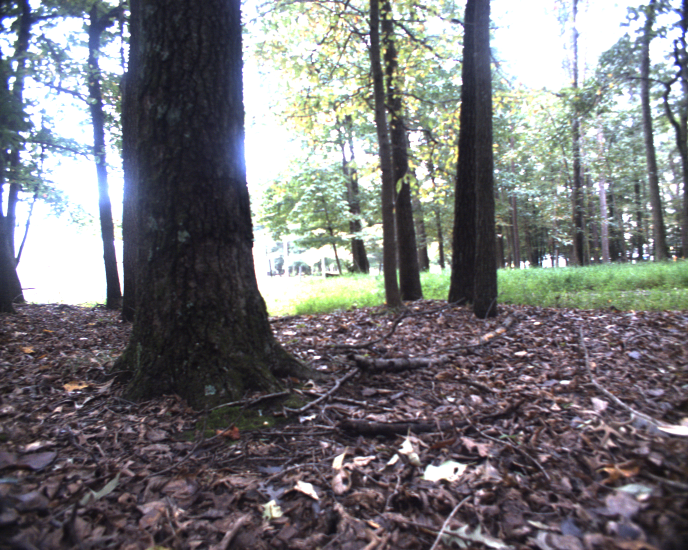}
    \caption{trail-4}
    \end{subfigure}
    \begin{subfigure}[b]{0.15\textwidth}
    \includegraphics[width = \textwidth]{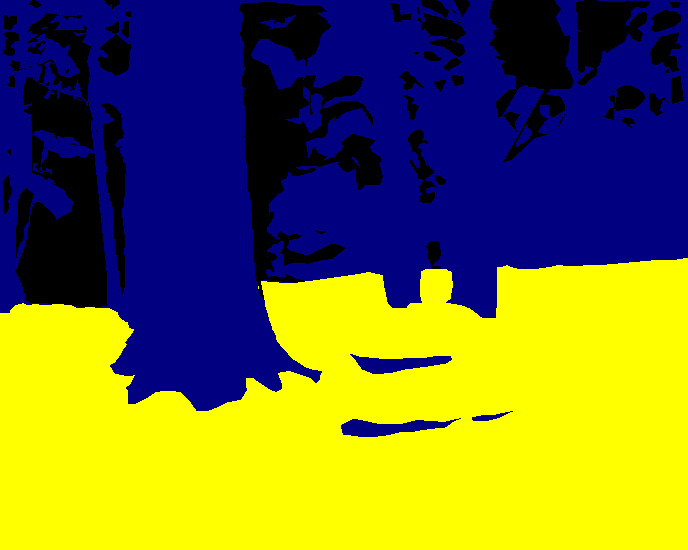}
    \caption{GT}
    \end{subfigure}
    \begin{subfigure}[b]{0.15\textwidth}
    \includegraphics[width = \textwidth]{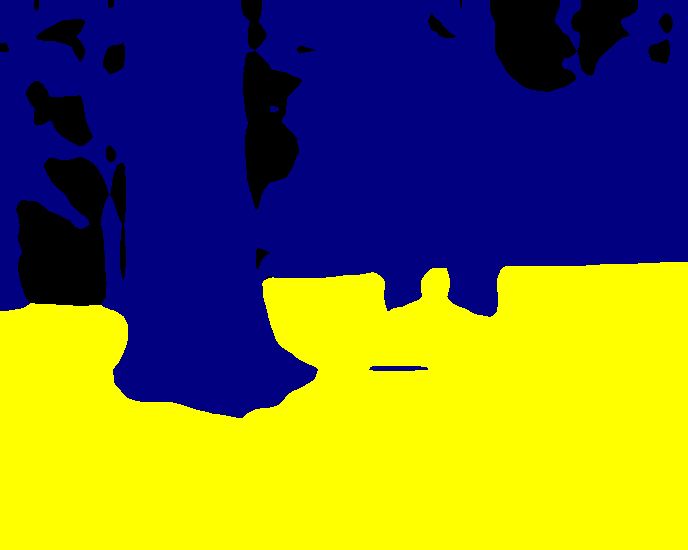}
    \caption{Ours}
    \end{subfigure}
    \begin{subfigure}[b]{0.15\textwidth}
    \includegraphics[width = \textwidth]{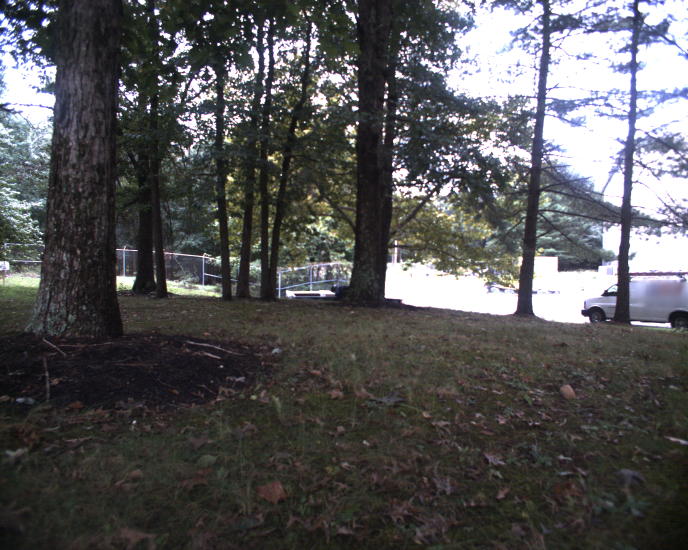}
    \caption{trail-5}
    \end{subfigure}
    \begin{subfigure}[b]{0.15\textwidth}
    \includegraphics[width = \textwidth]{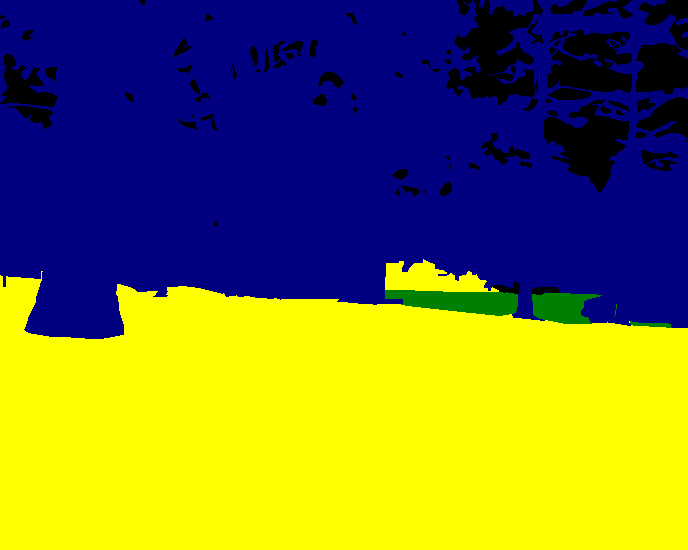}
    \caption{GT}
    \end{subfigure}
    \begin{subfigure}[b]{0.15\textwidth}
    \includegraphics[width = \textwidth]{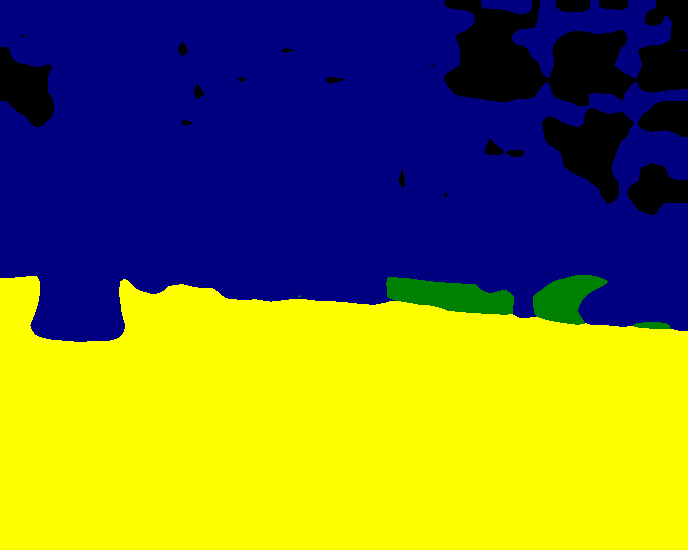}
    \caption{Ours}
    \end{subfigure}         
     \\
    \begin{subfigure}[b]{0.15\textwidth}
    \includegraphics[width = \textwidth]{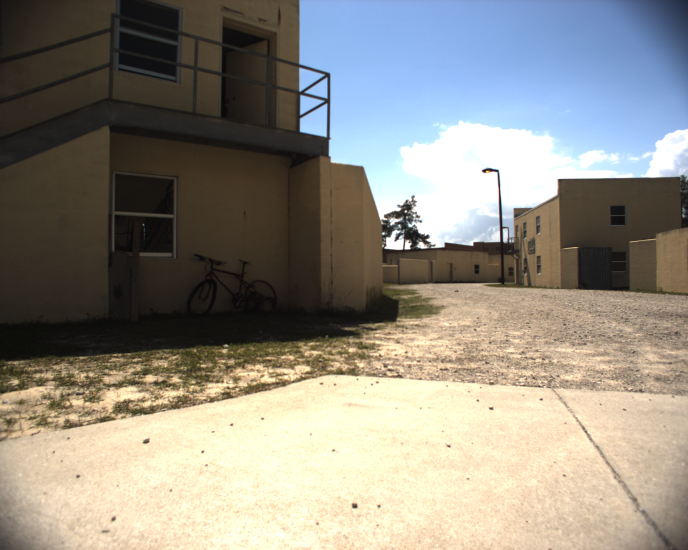}
    \caption{village-1}
    \end{subfigure}
    \begin{subfigure}[b]{0.15\textwidth}
    \includegraphics[width = \textwidth]{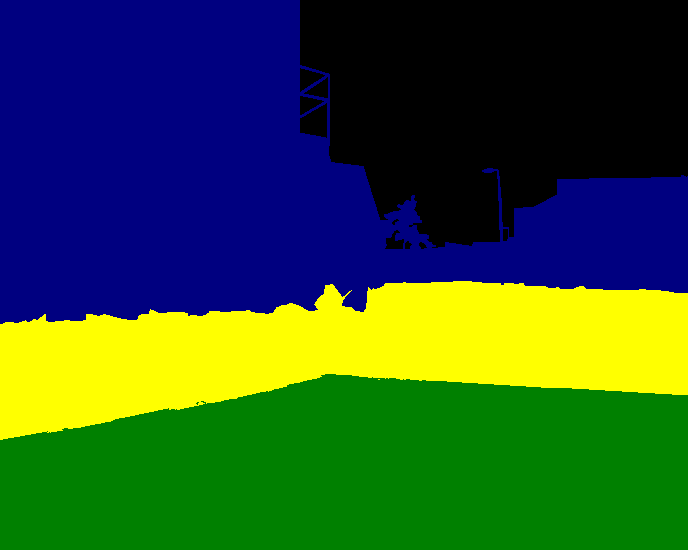}
    \caption{GT}
    \end{subfigure}
    \begin{subfigure}[b]{0.15\textwidth}
    \includegraphics[width = \textwidth]{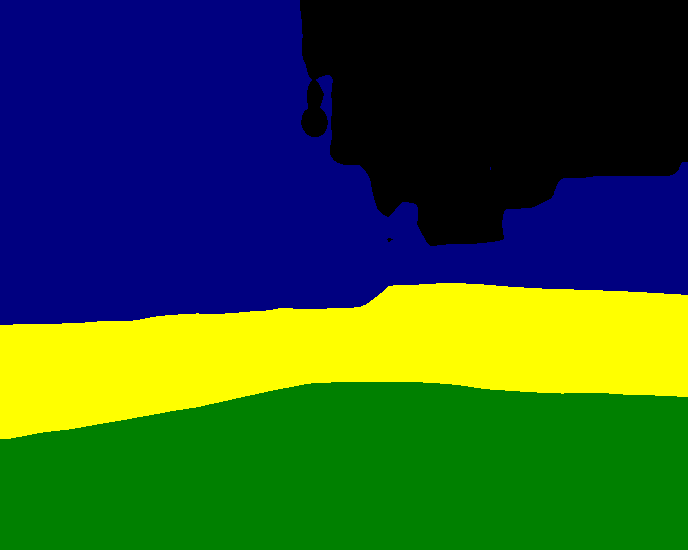}
    \caption{Ours}
    \end{subfigure}
    \begin{subfigure}[b]{0.15\textwidth}
    \includegraphics[width = \textwidth]{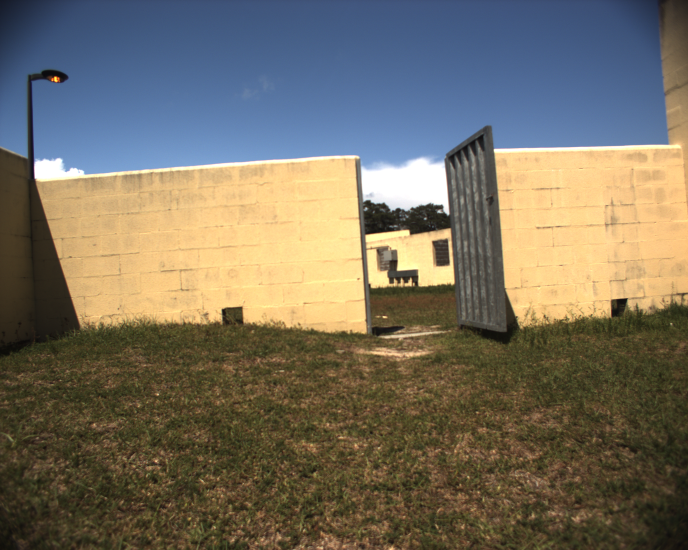}
    \caption{village-2}
    \end{subfigure}
    \begin{subfigure}[b]{0.15\textwidth}
    \includegraphics[width = \textwidth]{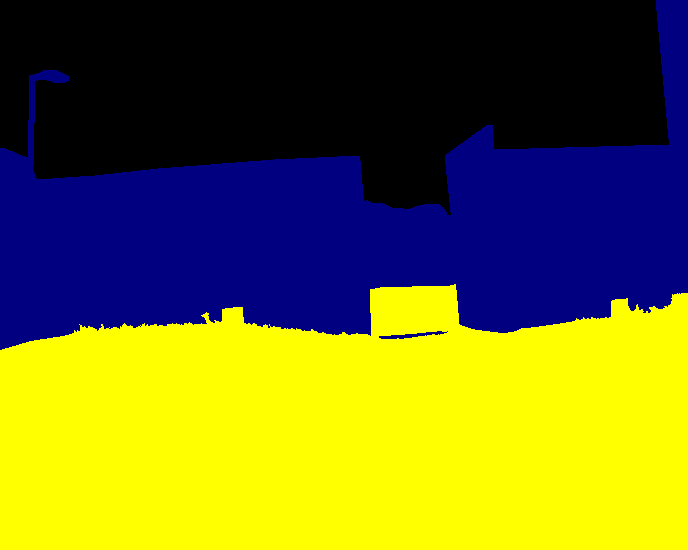}
    \caption{GT}
    \end{subfigure}
    \begin{subfigure}[b]{0.15\textwidth}
    \includegraphics[width = \textwidth]{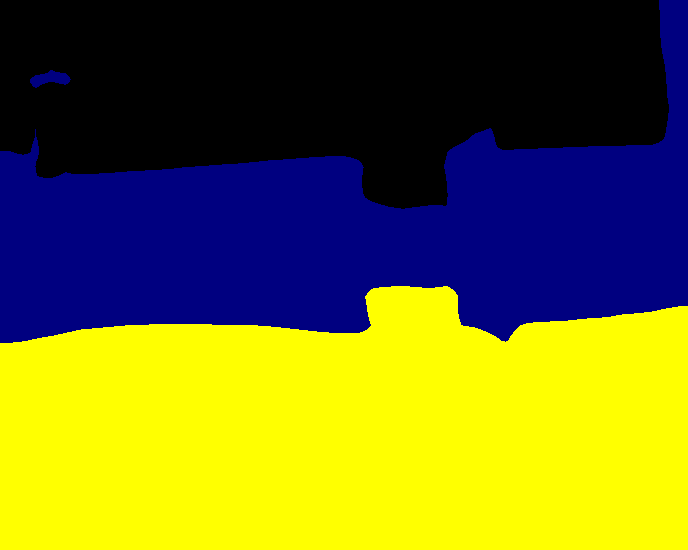}
    \caption{Ours}
    \end{subfigure}      
     \\
    \begin{subfigure}[b]{0.15\textwidth}
    \includegraphics[width = \textwidth]{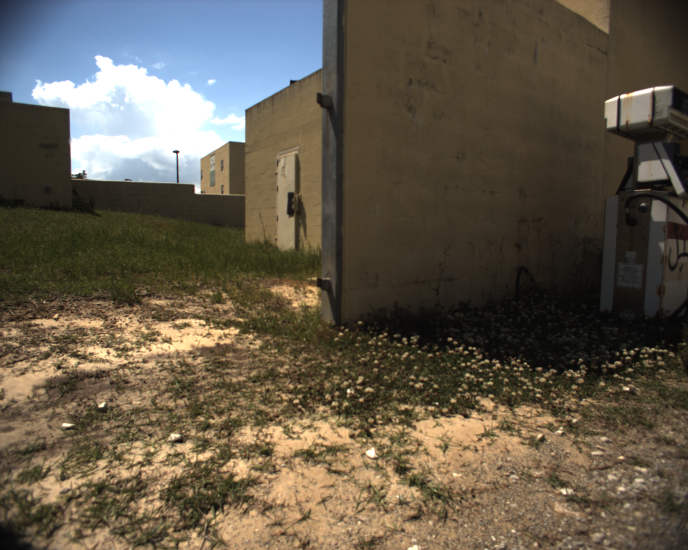}
    \caption{village-3}
    \end{subfigure}
    \begin{subfigure}[b]{0.15\textwidth}
    \includegraphics[width = \textwidth]{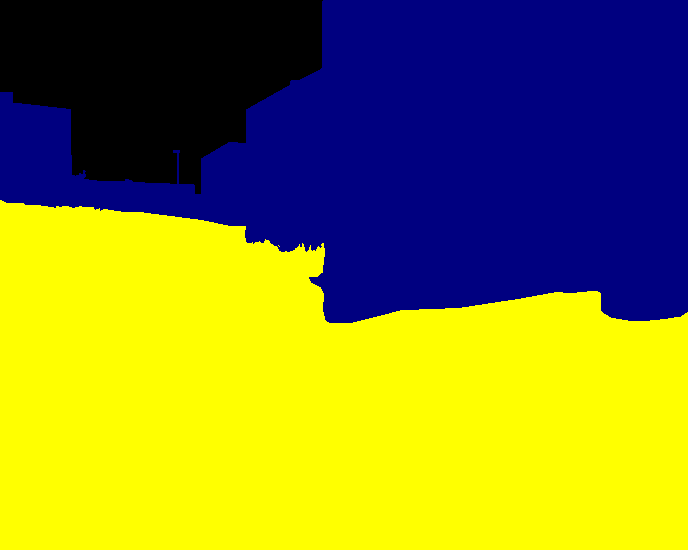}
    \caption{GT}
    \end{subfigure}
    \begin{subfigure}[b]{0.15\textwidth}
    \includegraphics[width = \textwidth]{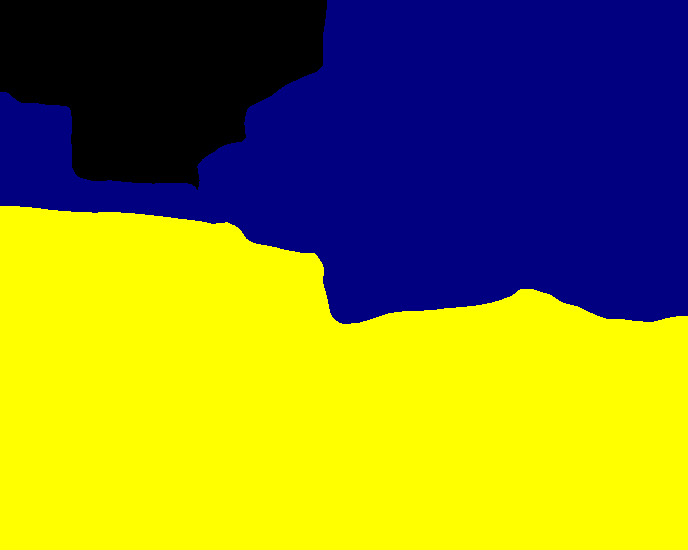}
    \caption{Ours}
    \end{subfigure}
    \begin{subfigure}[b]{0.15\textwidth}
    \includegraphics[width = \textwidth]{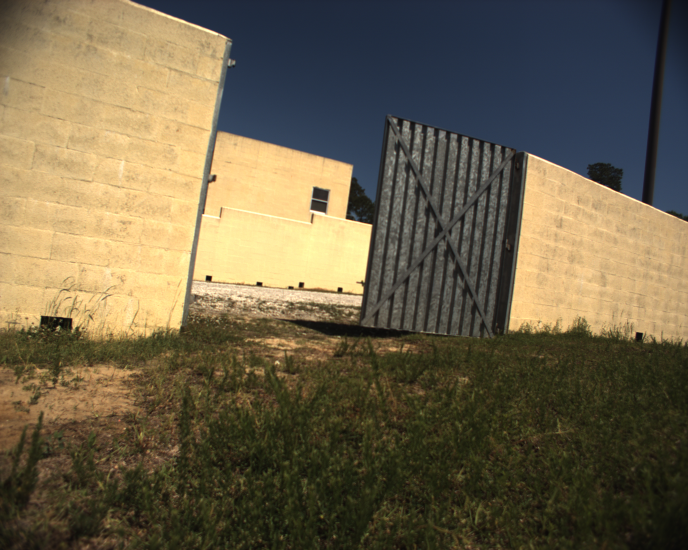}
    \caption{village-4}
    \end{subfigure}
    \begin{subfigure}[b]{0.15\textwidth}
    \includegraphics[width = \textwidth]{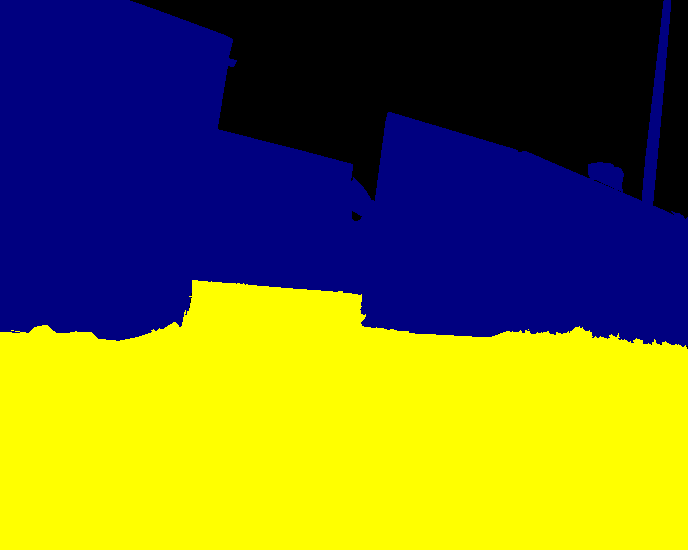}
    \caption{GT}
    \end{subfigure}
    \begin{subfigure}[b]{0.15\textwidth}
    \includegraphics[width = \textwidth]{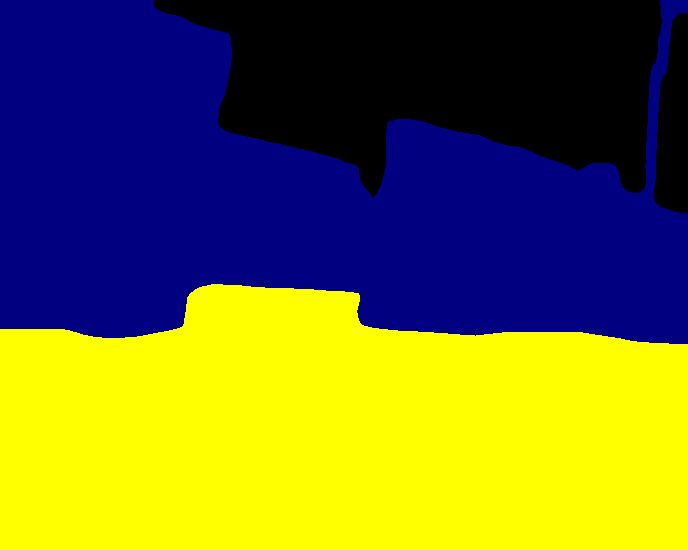}
    \caption{Ours}
    \end{subfigure}          
    \caption{\textbf{More Qualitative Results on RUGD}}
    \label{fig:visualisations2}
    
\end{figure*}

%% file: QualitativeResults/qualitative3.tex
\begin{figure*}[t]
    \centering
    \begin{subfigure}[b]{0.15\textwidth}
    \includegraphics[width = \textwidth]{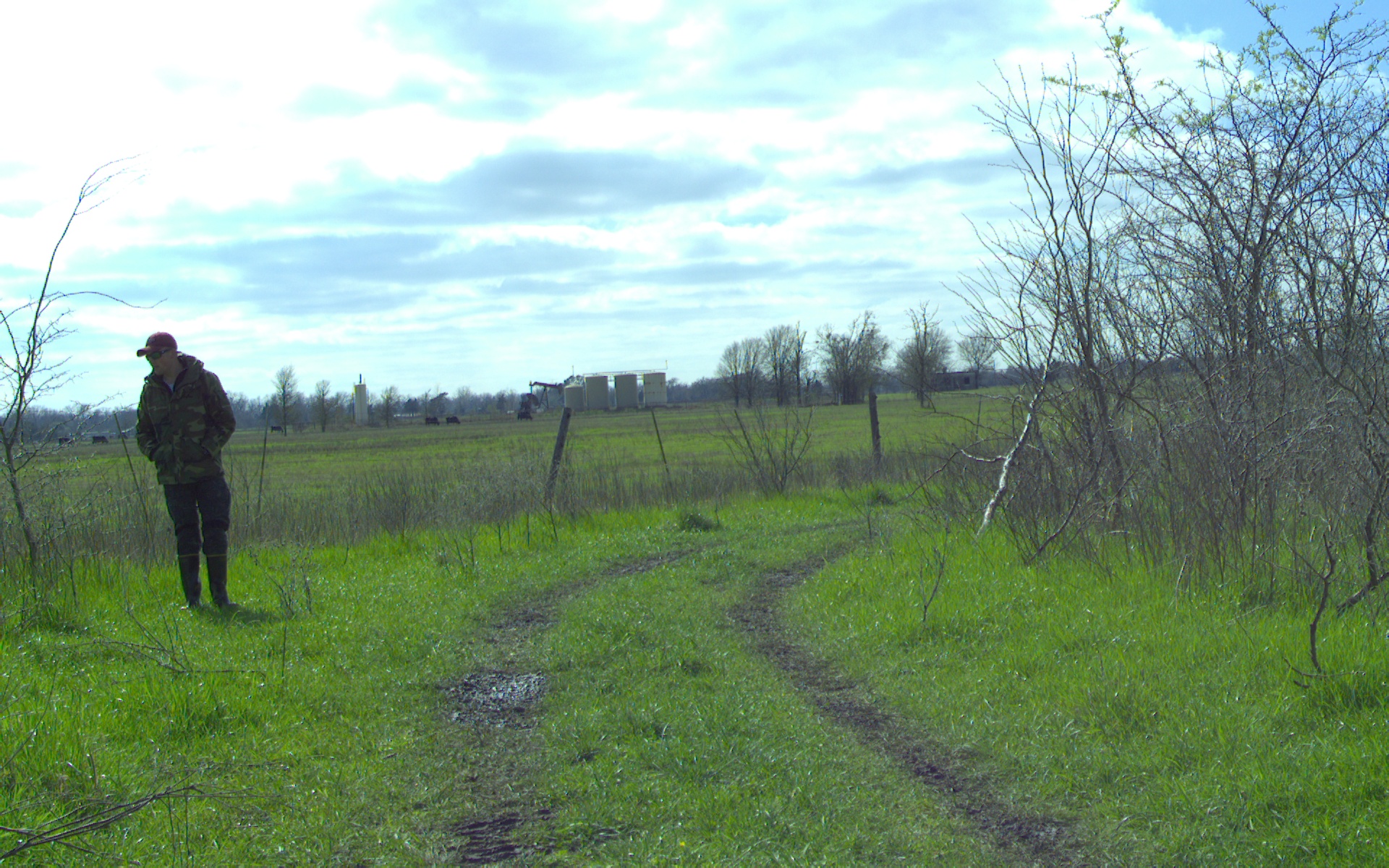}
    \caption{trail-1}
    \end{subfigure}
    \begin{subfigure}[b]{0.15\textwidth}
    \includegraphics[width = \textwidth]{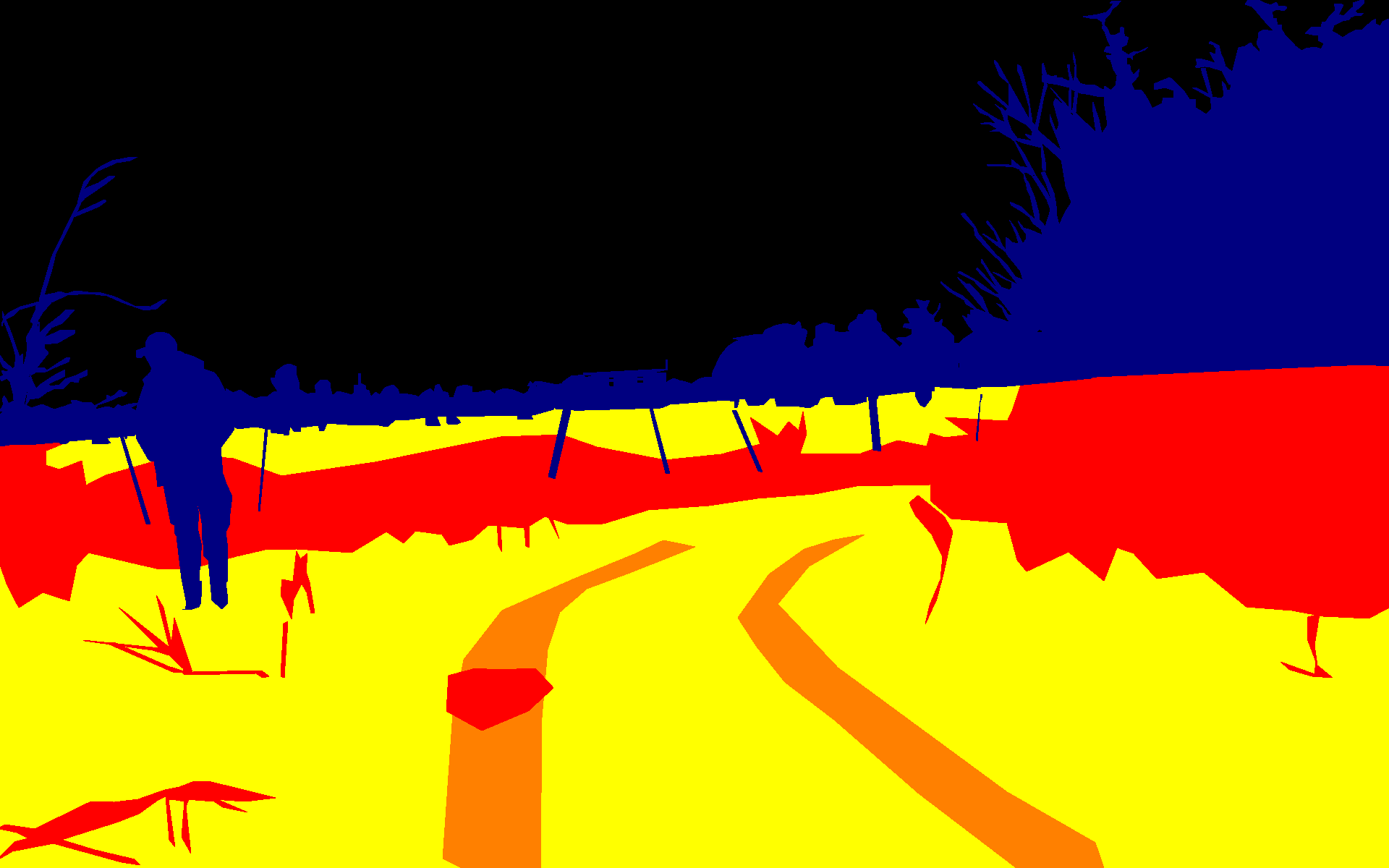}
    \caption{GT}
    \end{subfigure}
    \begin{subfigure}[b]{0.15\textwidth}
    \includegraphics[width = \textwidth]{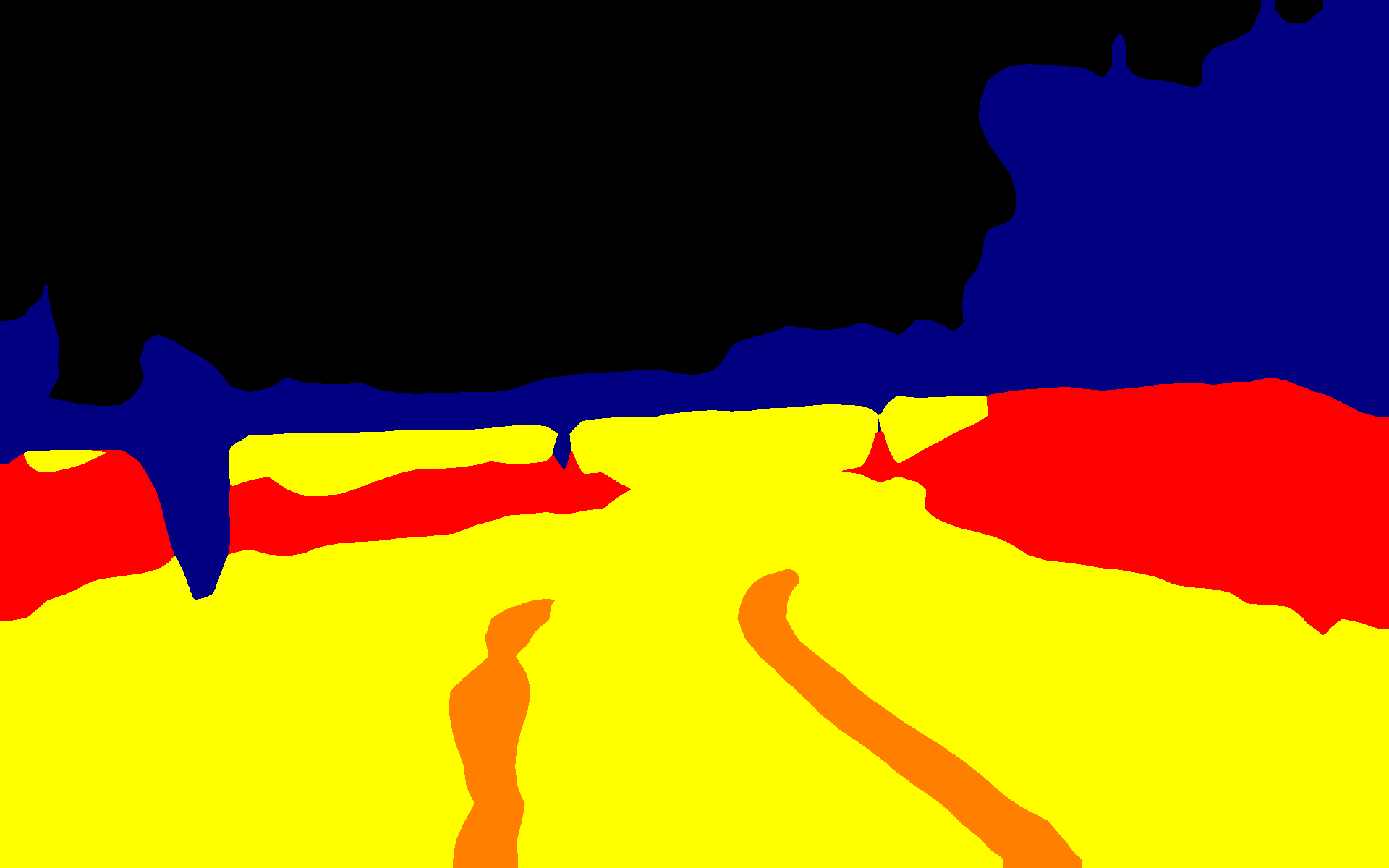}
    \caption{Ours}
    \end{subfigure}
    \begin{subfigure}[b]{0.15\textwidth}
    \includegraphics[width = \textwidth]{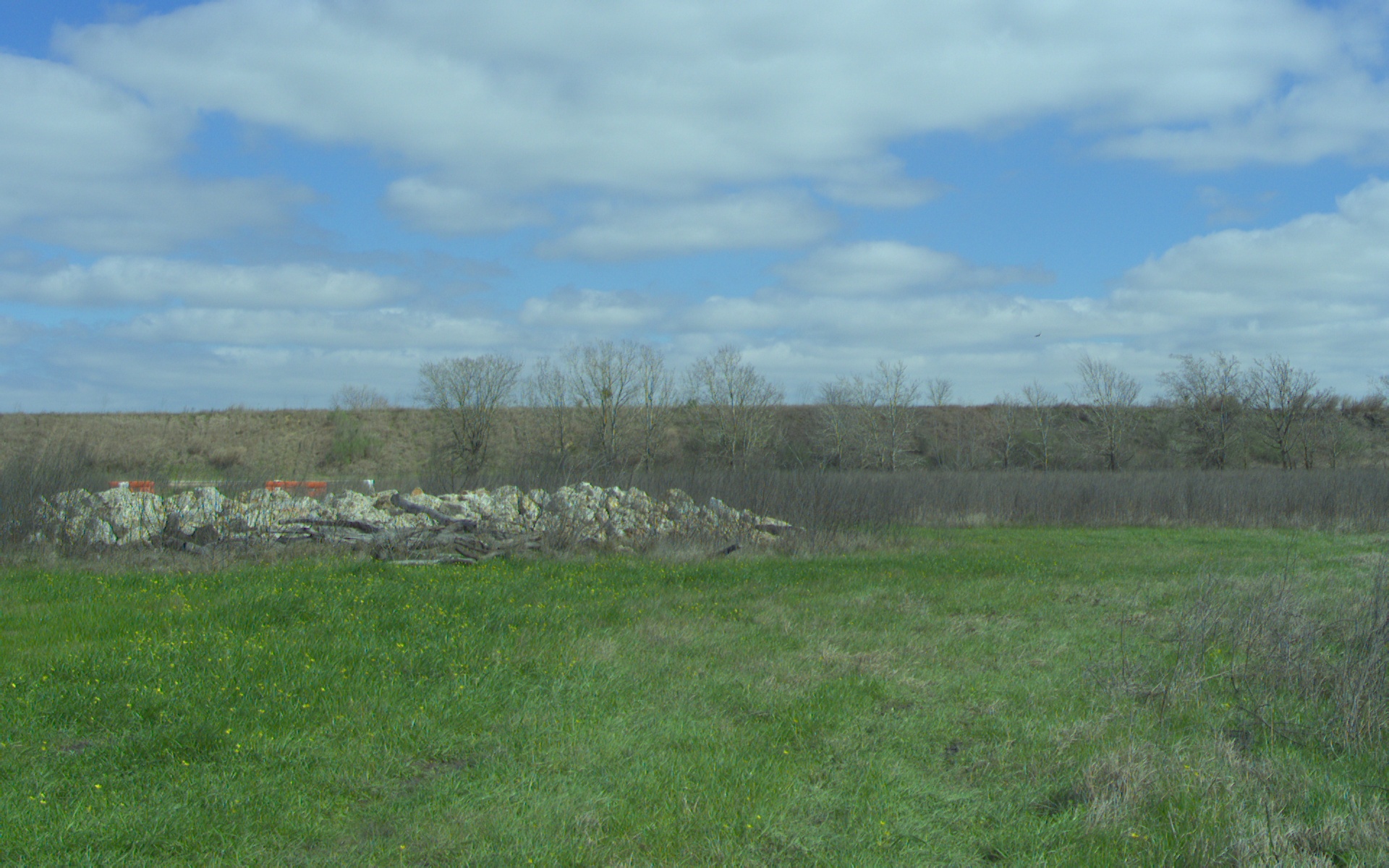}
    \caption{trail-2}
    \end{subfigure}
    \begin{subfigure}[b]{0.15\textwidth}
    \includegraphics[width = \textwidth]{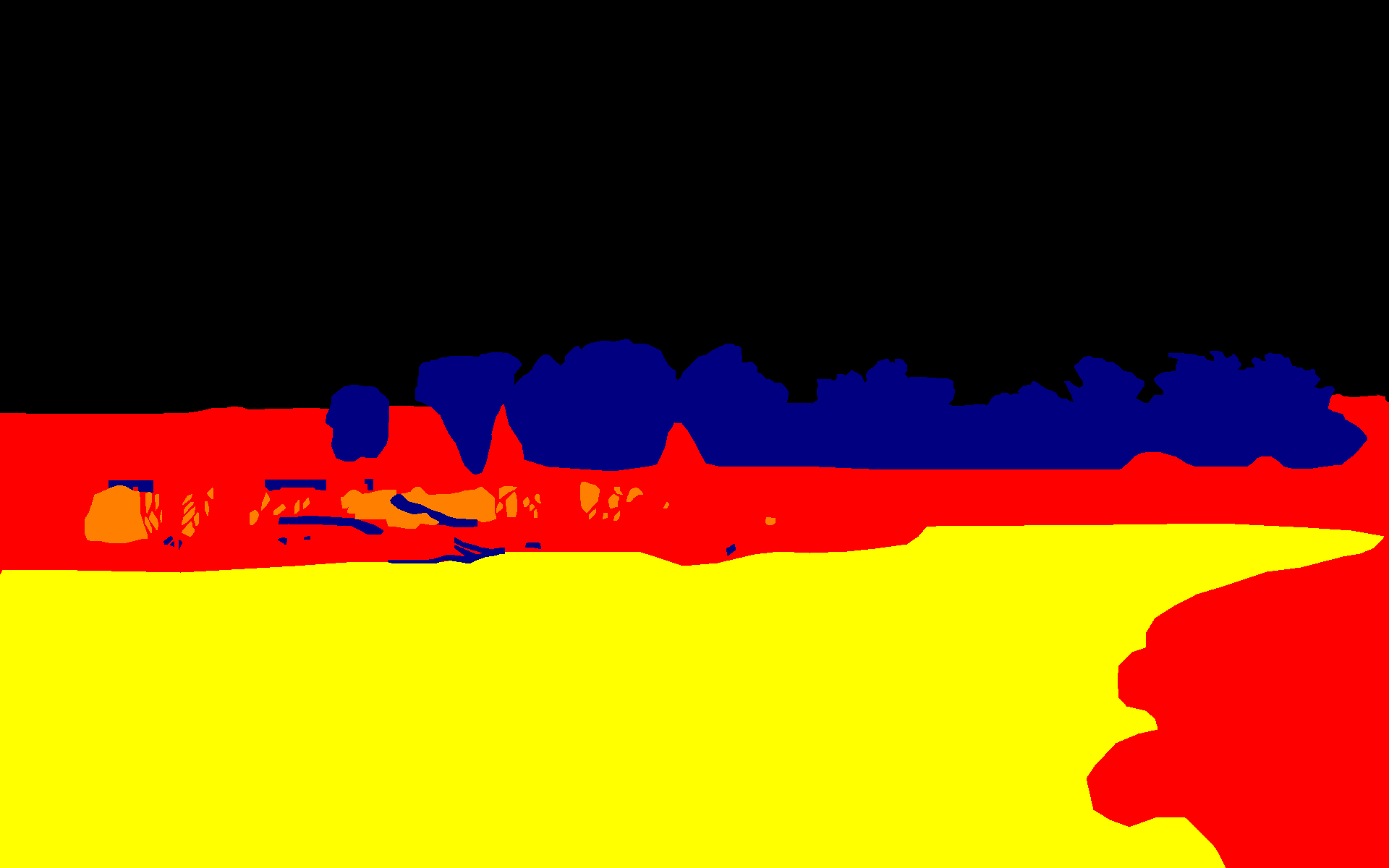}
    \caption{GT}
    \end{subfigure}
    \begin{subfigure}[b]{0.15\textwidth}
    \includegraphics[width = \textwidth]{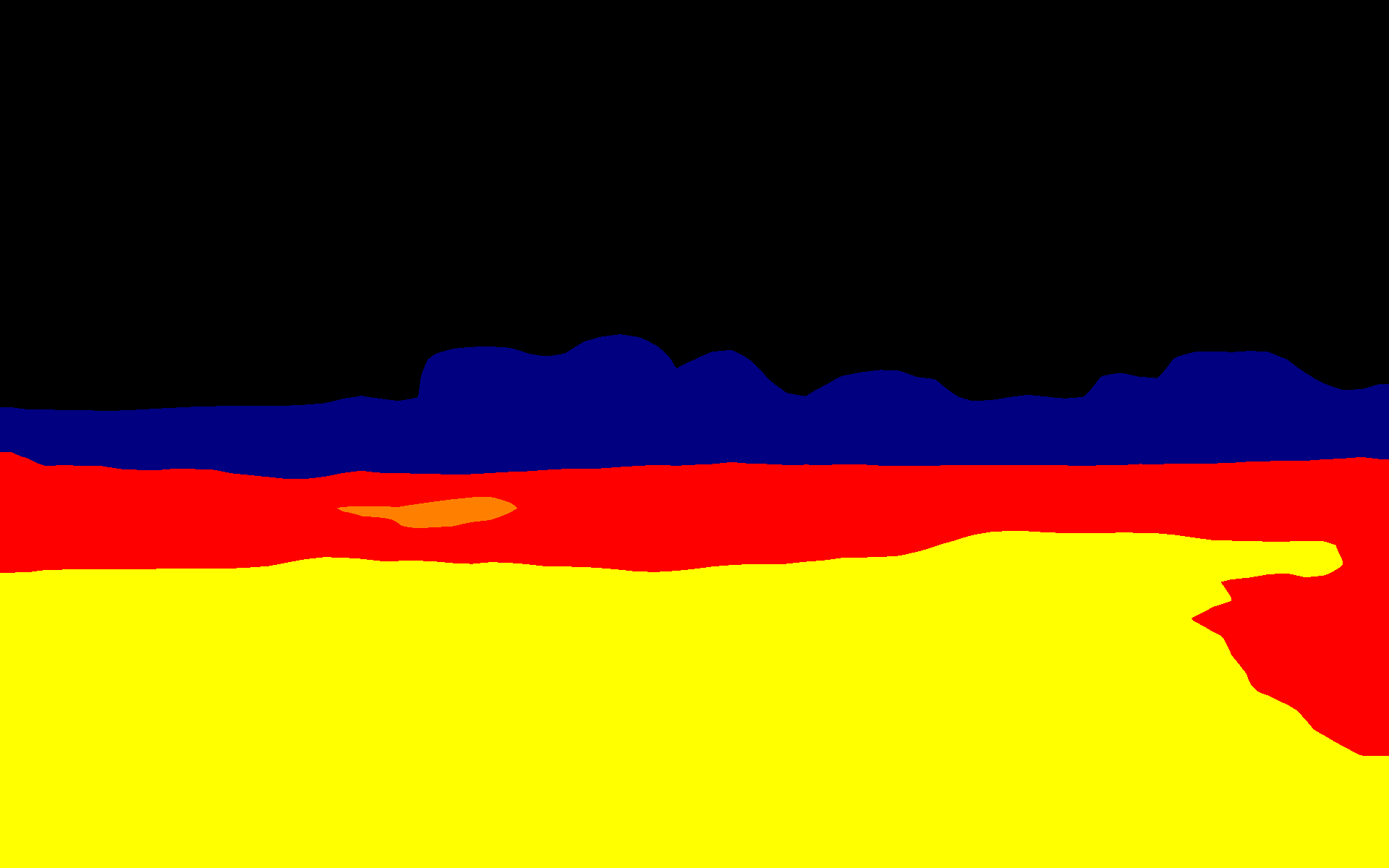}
    \caption{Ours}
    \end{subfigure}    
    \\
    \begin{subfigure}[b]{0.15\textwidth}
    \includegraphics[width = \textwidth]{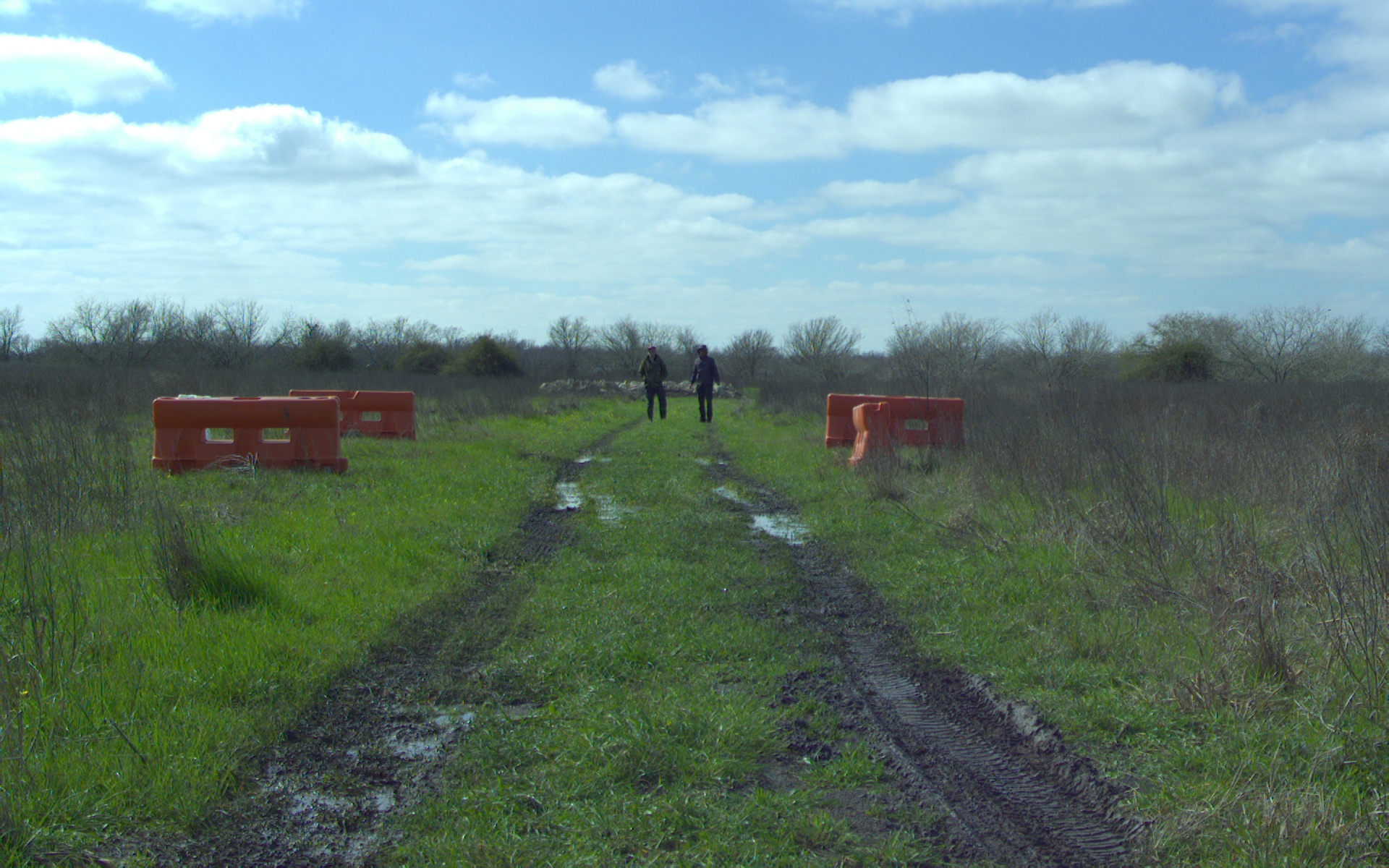}
    \caption{trail-3}
    \end{subfigure}
    \begin{subfigure}[b]{0.15\textwidth}
    \includegraphics[width = \textwidth]{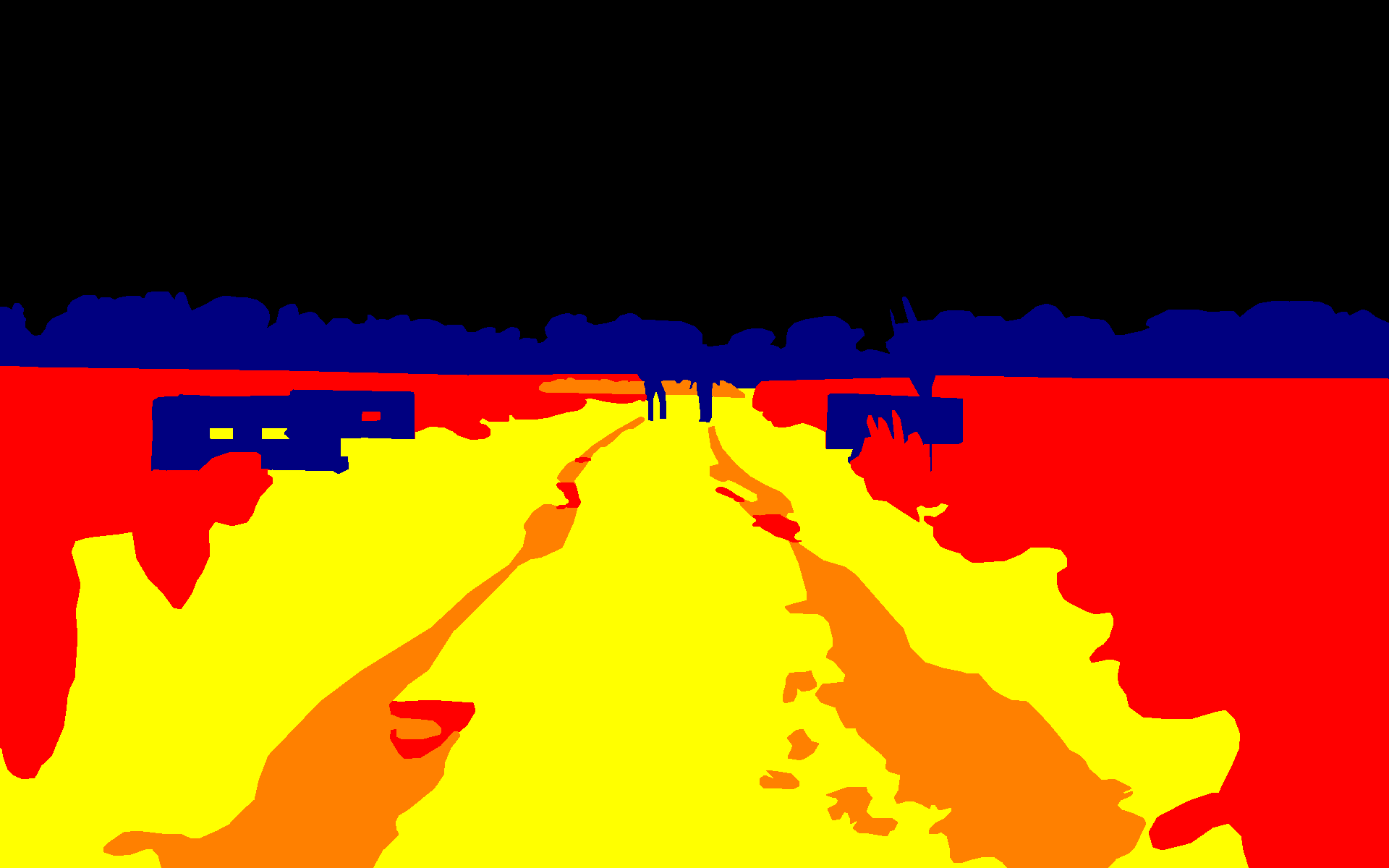}
    \caption{GT}
    \end{subfigure}
    \begin{subfigure}[b]{0.15\textwidth}
    \includegraphics[width = \textwidth]{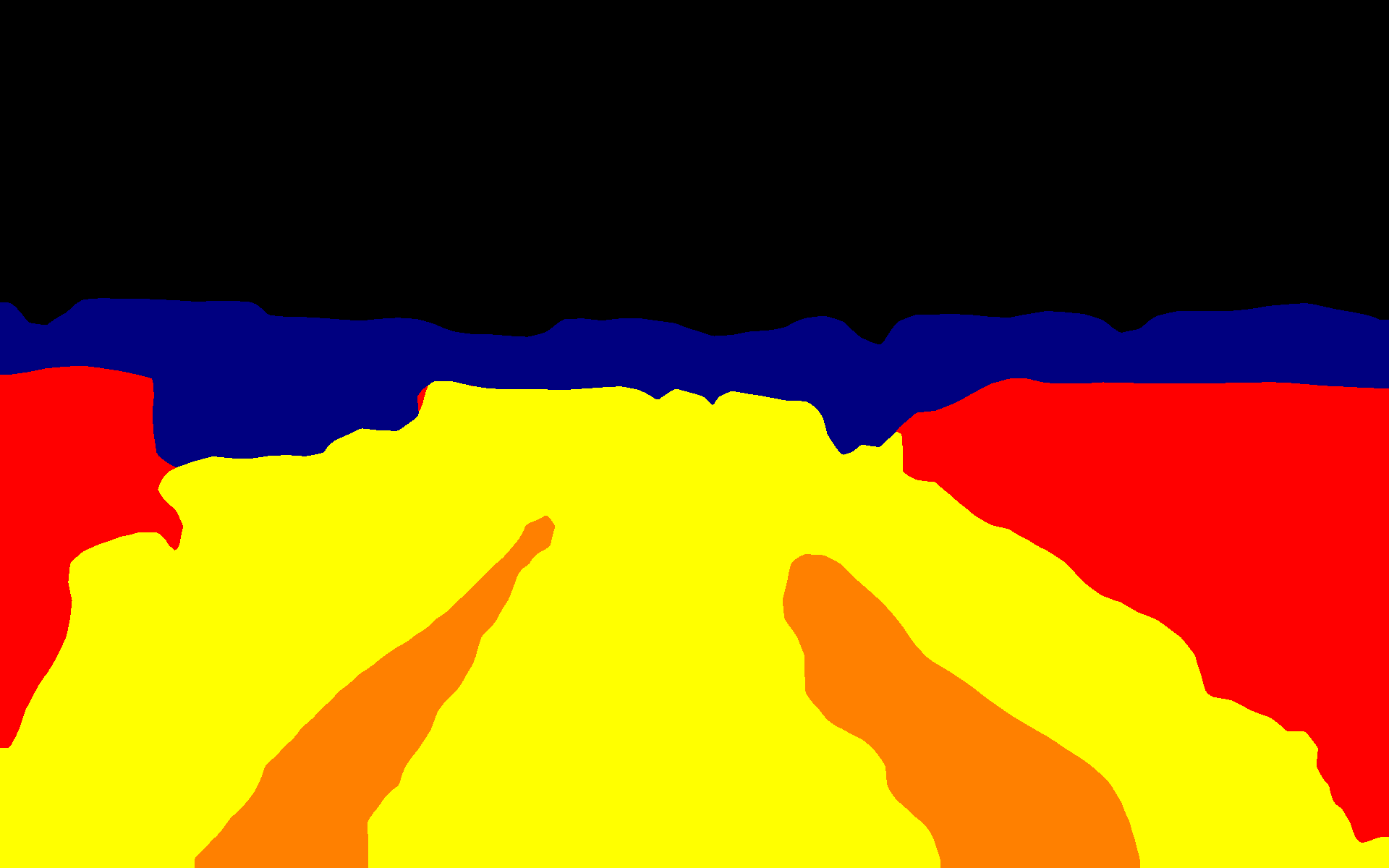}
    \caption{Ours}
    \end{subfigure}
    \begin{subfigure}[b]{0.15\textwidth}
    \includegraphics[width = \textwidth]{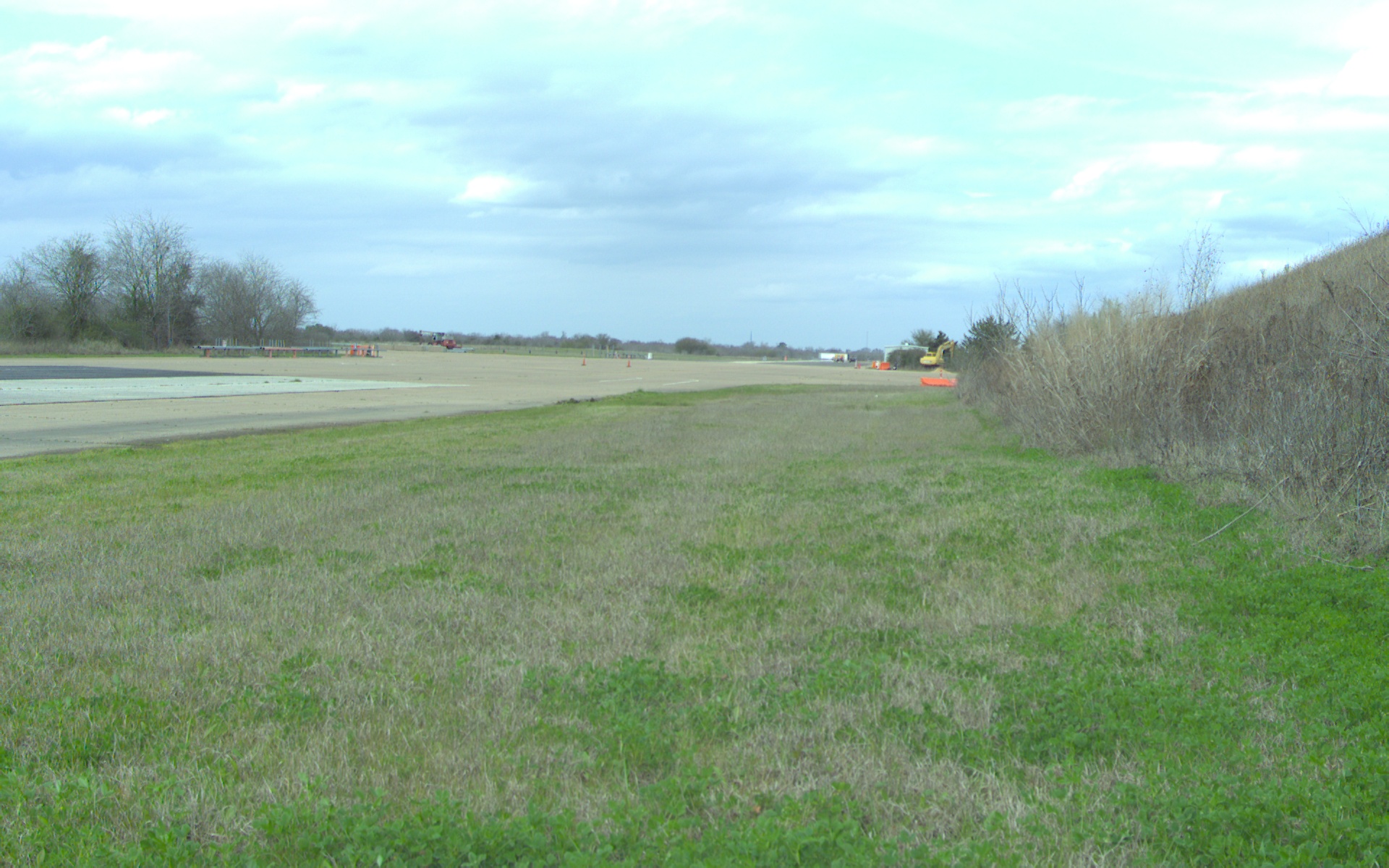}
    \caption{trail-4}
    \end{subfigure}
    \begin{subfigure}[b]{0.15\textwidth}
    \includegraphics[width = \textwidth]{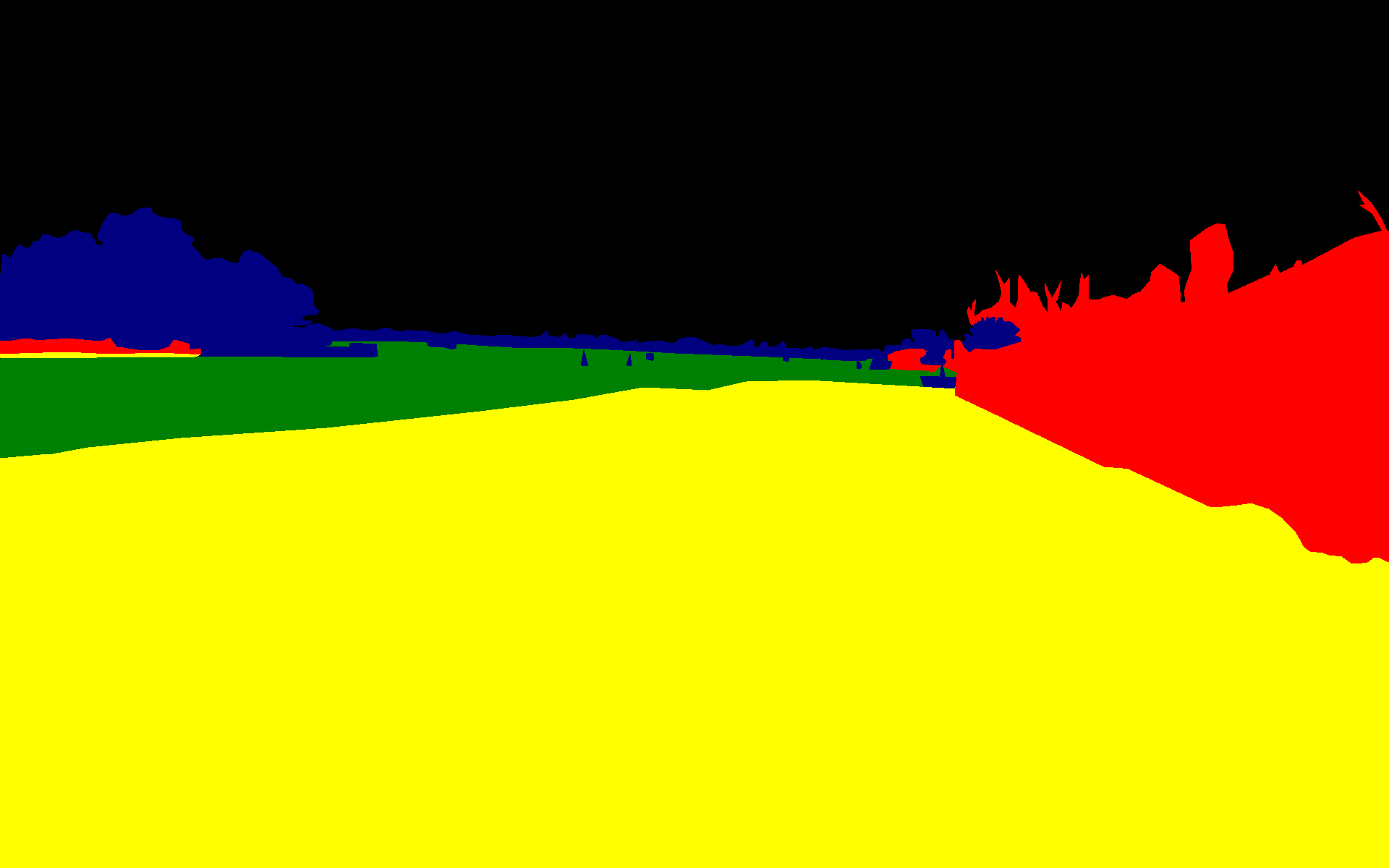}
    \caption{GT}
    \end{subfigure}
    \begin{subfigure}[b]{0.15\textwidth}
    \includegraphics[width = \textwidth]{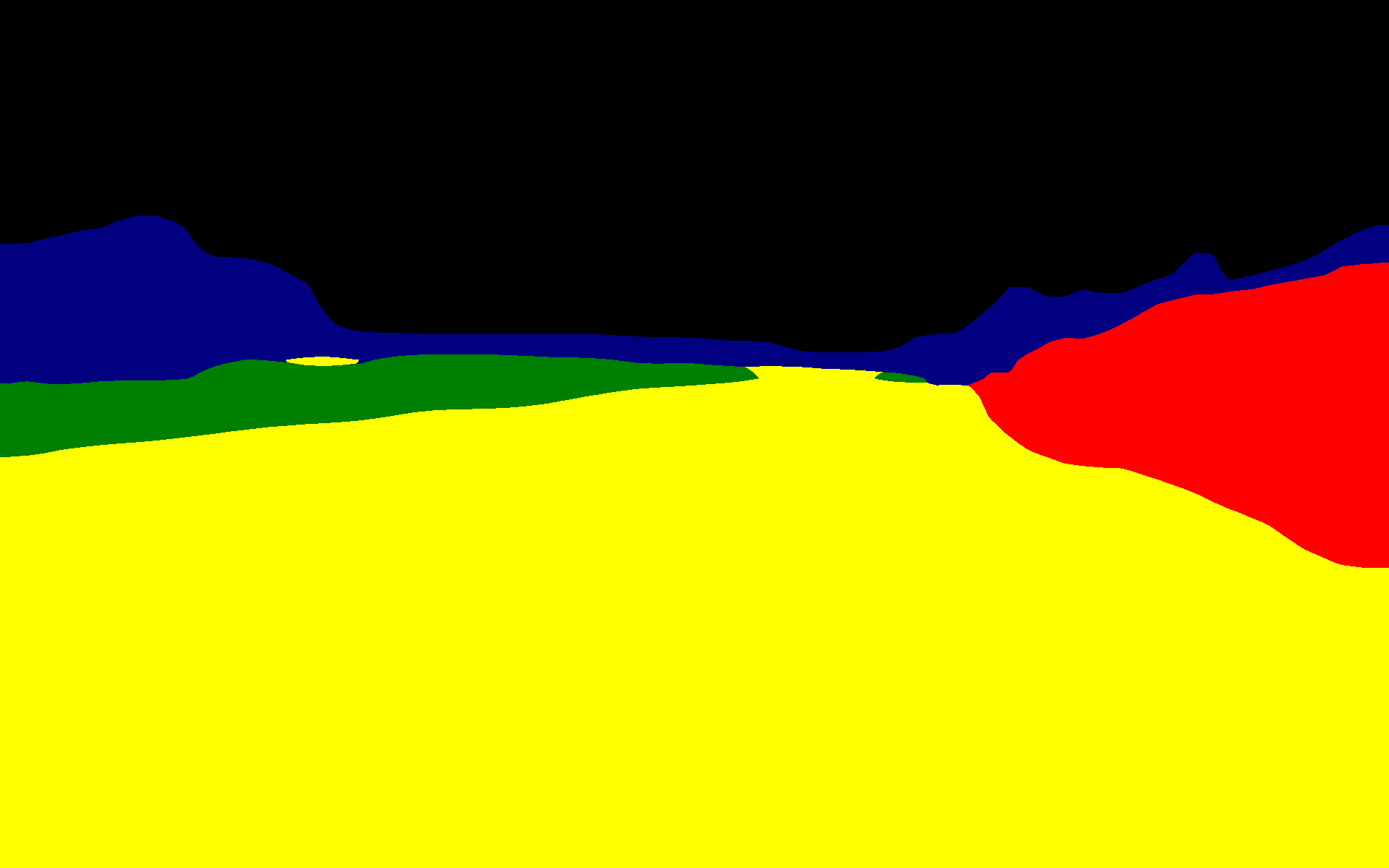}
    \caption{Ours}
    \end{subfigure}      
     \\
    \begin{subfigure}[b]{0.15\textwidth}
    \includegraphics[width = \textwidth]{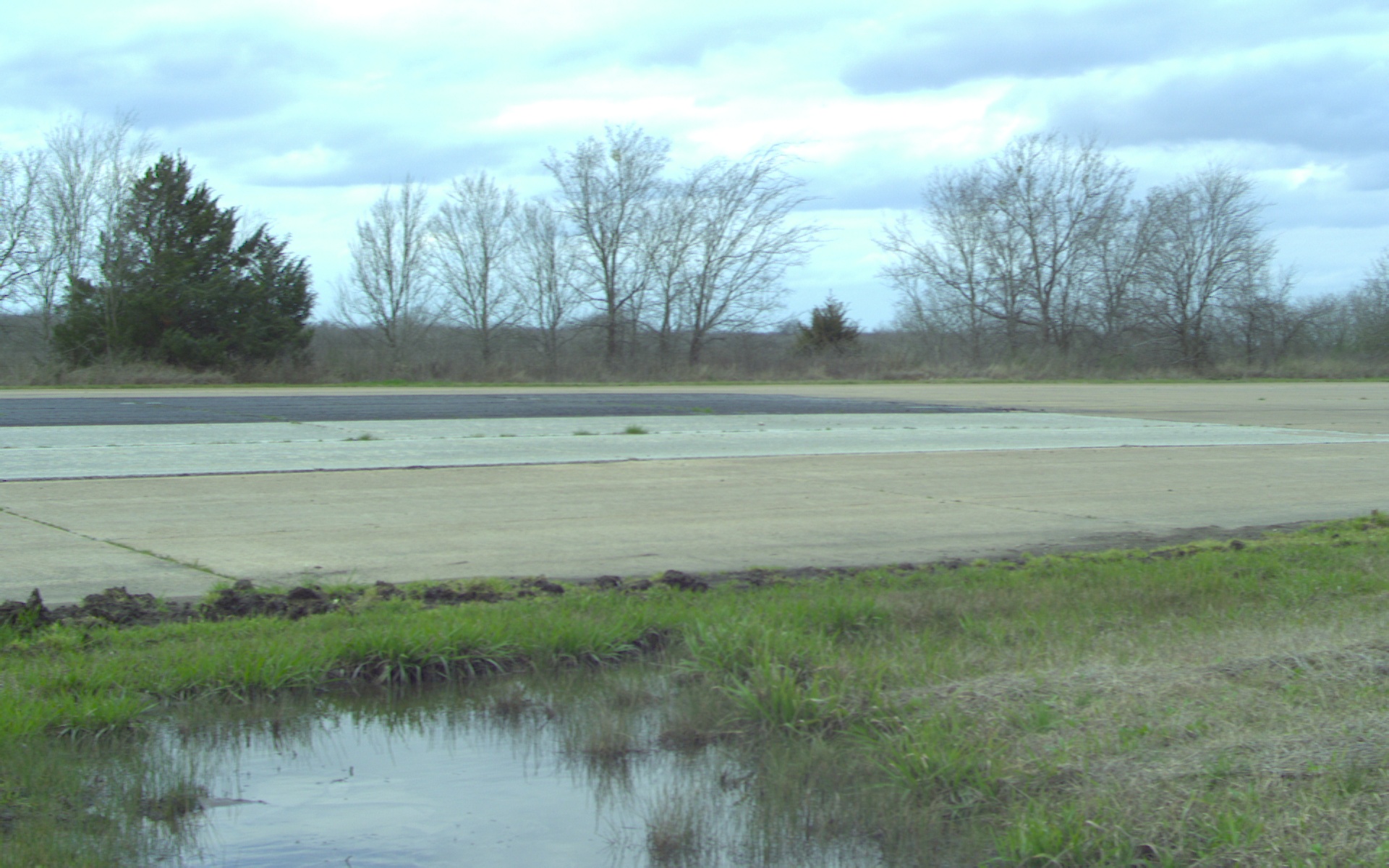}
    \caption{trail-5}
    \end{subfigure}
    \begin{subfigure}[b]{0.15\textwidth}
    \includegraphics[width = \textwidth]{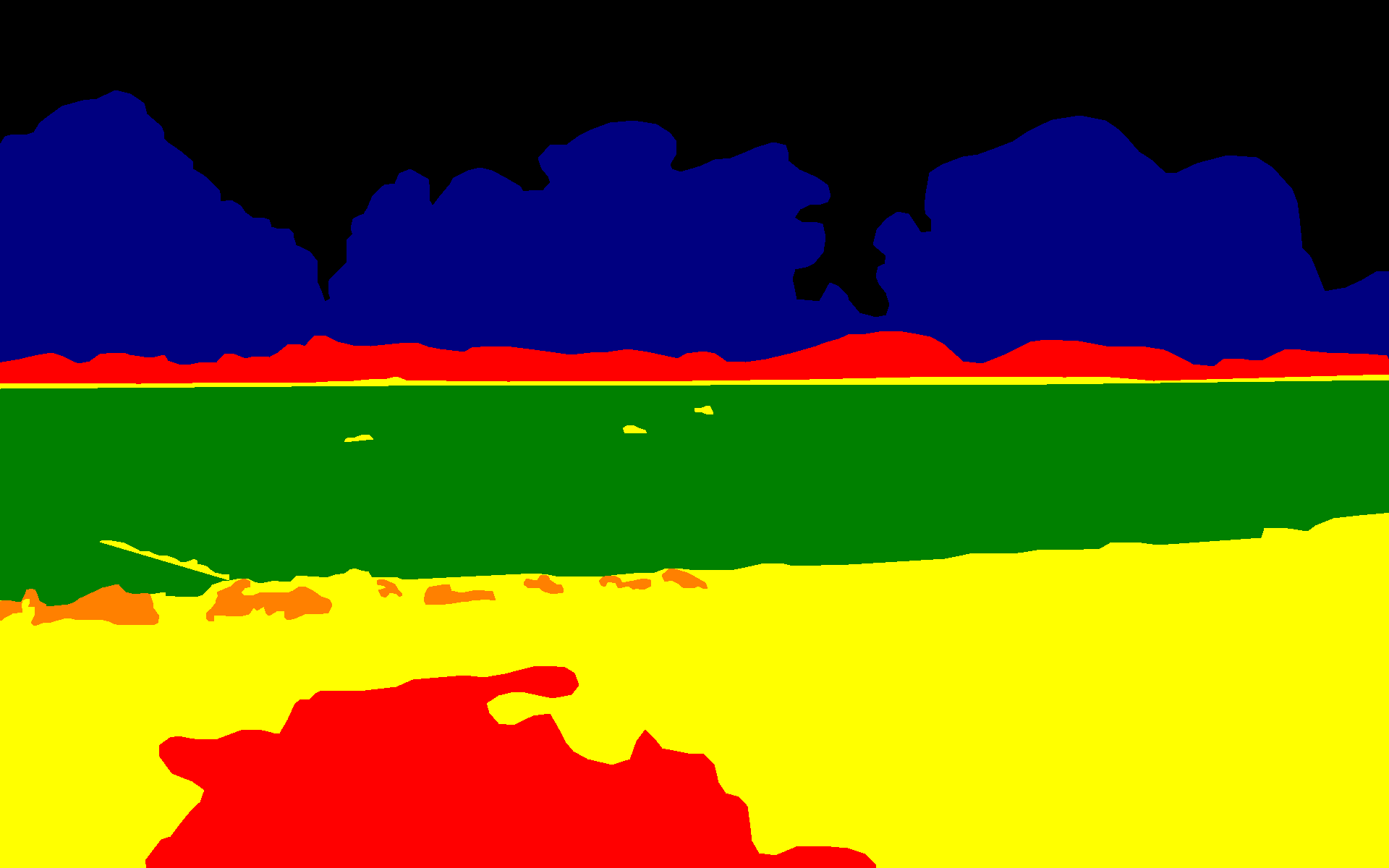}
    \caption{GT}
    \end{subfigure}
    \begin{subfigure}[b]{0.15\textwidth}
    \includegraphics[width = \textwidth]{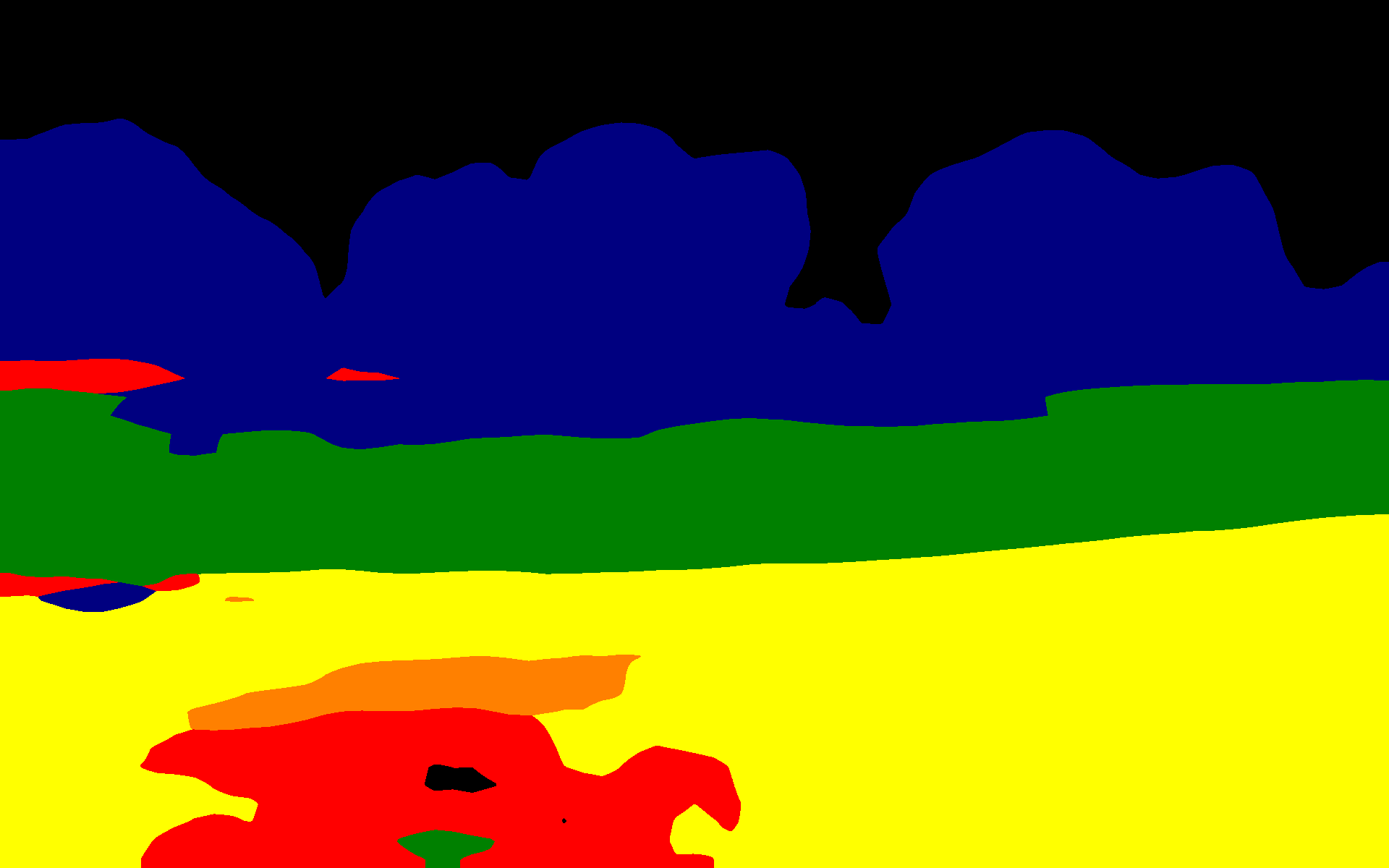}
    \caption{Ours}
    \end{subfigure}
    \begin{subfigure}[b]{0.15\textwidth}
    \includegraphics[width = \textwidth]{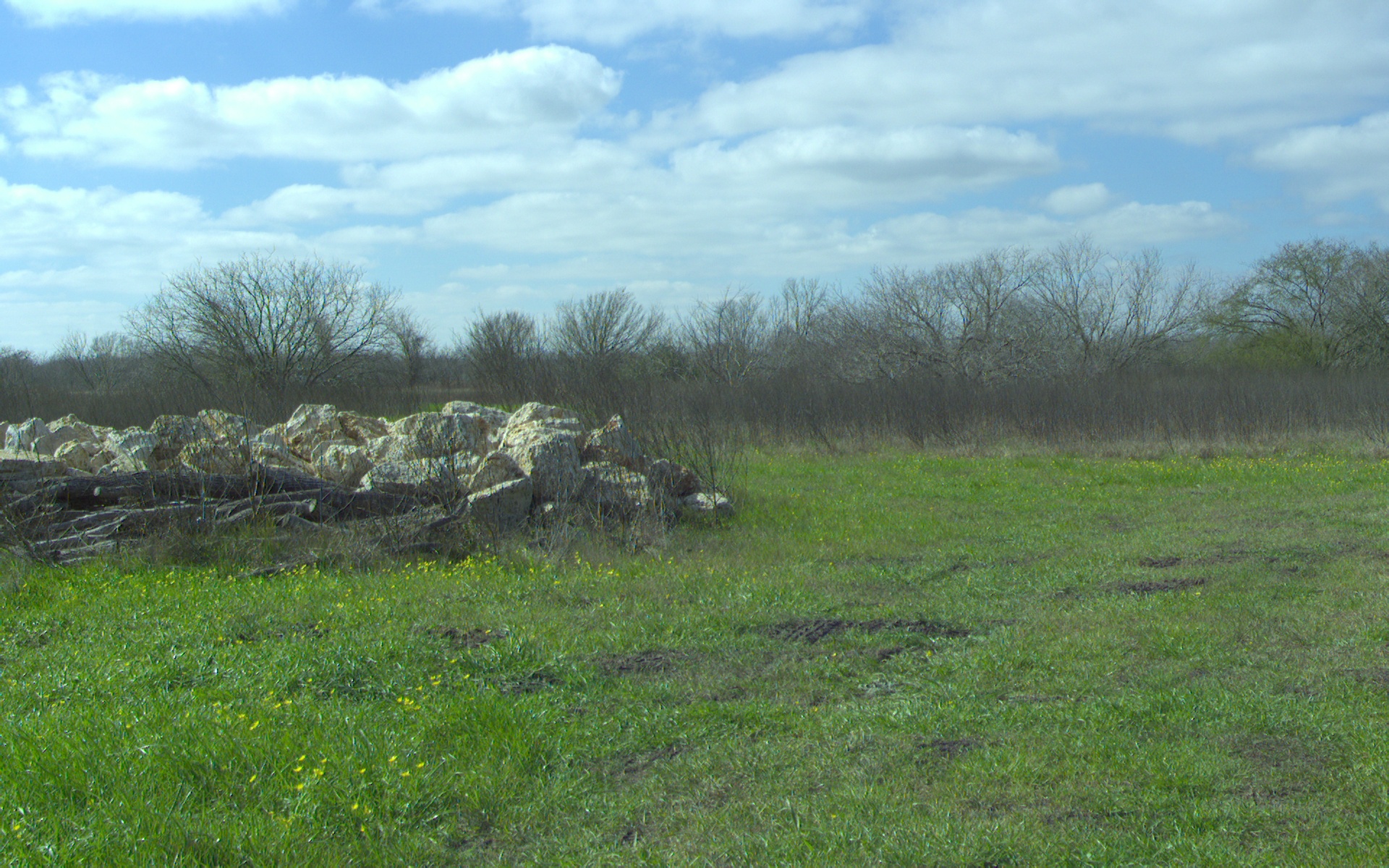}
    \caption{trail-6}
    \end{subfigure}
    \begin{subfigure}[b]{0.15\textwidth}
    \includegraphics[width = \textwidth]{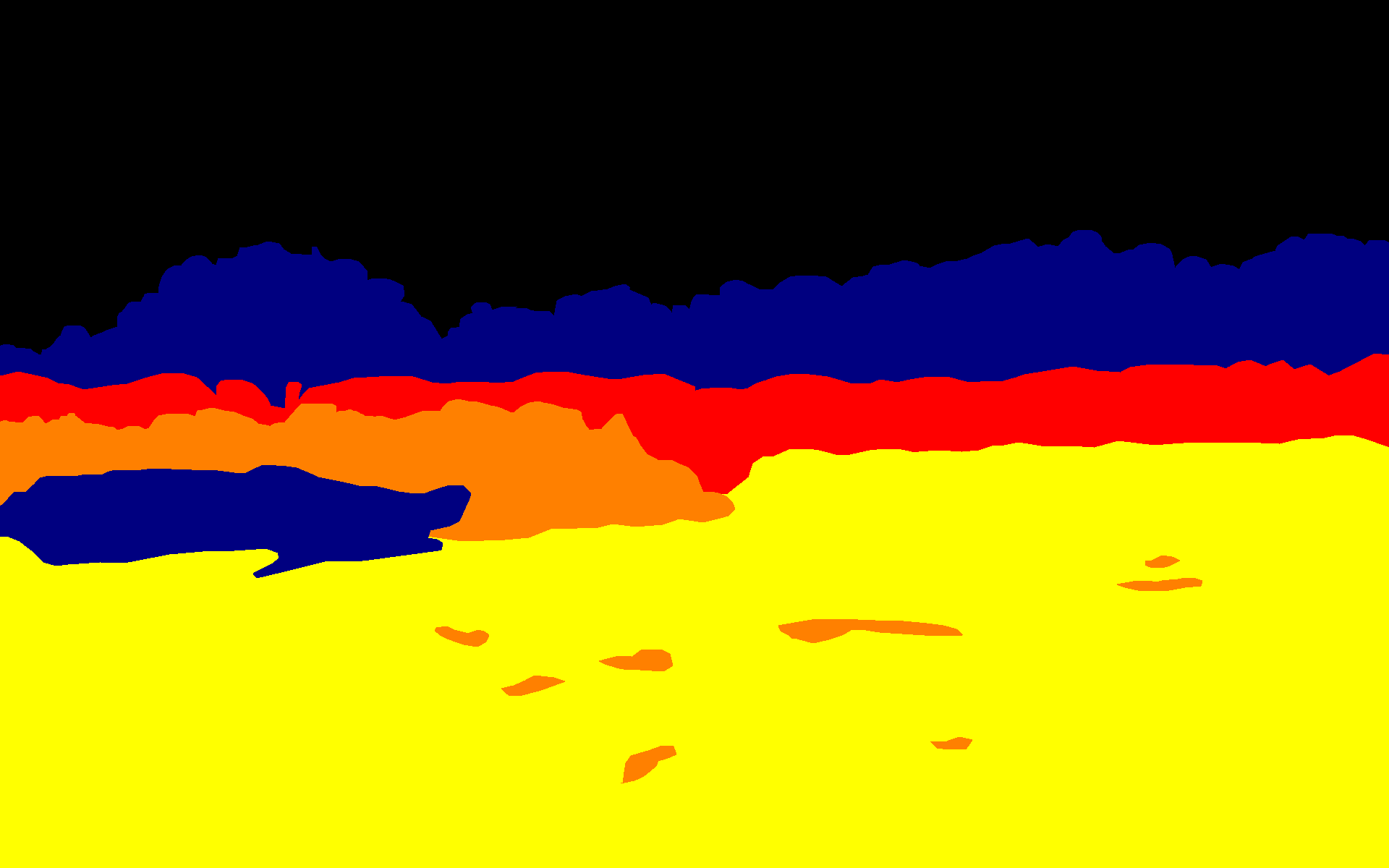}
    \caption{GT}
    \end{subfigure}
    \begin{subfigure}[b]{0.15\textwidth}
    \includegraphics[width = \textwidth]{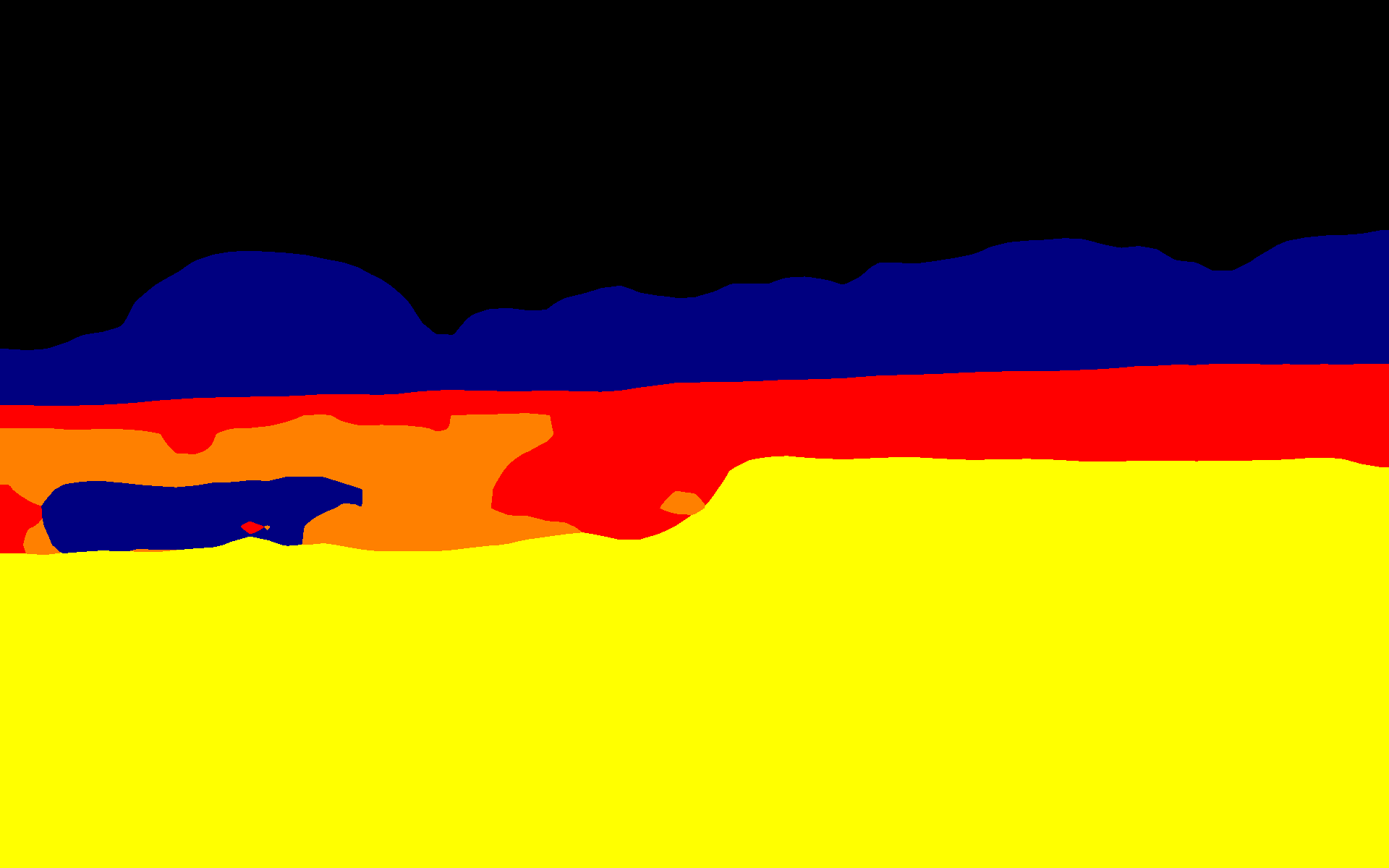}
    \caption{Ours}
    \end{subfigure}

    \caption{\textbf{More Qualitative Results on RELLIS-3D}}
    \label{fig:visualisations3}
    
\end{figure*}